\documentclass{article}

     \usepackage[nonatbib,preprint]{neurips_2019}

\usepackage[utf8]{inputenc} 
\usepackage[T1]{fontenc}    
\usepackage{hyperref}       
\usepackage{url}            
\usepackage{booktabs}       
\usepackage{amsfonts}       
\usepackage{amsmath}
\usepackage{amssymb}
\usepackage{bbm}
\usepackage{algorithm}
\usepackage{algpseudocode}
\usepackage{graphicx}
\usepackage{subcaption}
\usepackage{nicefrac}       
\usepackage{microtype}      

\newcommand{\btheta}{\boldsymbol{\theta}}
\renewcommand{\b}{\mathbf{b}}
\renewcommand{\H}{\mathbf{H}}
\renewcommand{\P}{\mathbb{P}}
\newcommand{\E}{\mathbb{E}}
\newcommand{\I}{\mathbf{I}}
\renewcommand{\i}{\mathbf{i}}
\newcommand{\J}{\mathbf{J}}
\newcommand{\X}{\mathbf{X}}
\newcommand{\Y}{\mathbf{Y}}
\newcommand{\W}{\mathbf{W}}
\newcommand{\R}{\mathbf{R}}
\newcommand{\G}{\mathbf{G}}
\newcommand{\Hb}{\mathbf{H}}
\newcommand{\rb}{\mathbf{r}}
\newcommand{\gb}{\mathbf{g}}
\newcommand{\ub}{\mathbf{u}}
\newcommand{\wb}{\mathbf{w}}
\newcommand{\p}{\mathbf{p}}
\newcommand{\q}{\mathbf{q}}
\newcommand{\s}{\mathbf{s}}
\DeclareMathOperator*{\argmax}{arg\,max}

\newcommand{\Info}{\mathcal{I}}
\newcommand{\Ent}{\mathcal{H}}

\title{Simple Algorithms for Dueling Bandits}

\author{%
	Tyler Lekang \\
	University of Minnesota, Twin Cities\\
	Minneapolis, MN \\
	\texttt{lekang@umn.edu} \\
	\And
	Andrew Lamperski \\
	University of Minnesota, Twin Cities\\
	Minneapolis, MN \\
	\texttt{alampers@umn.edu} \\
}

\begin{document}

\maketitle

\begin{abstract}
	In this paper, we present simple algorithms for Dueling
	Bandits. We prove that the algorithms have regret bounds for
	time horizon $T$ of order $O(T^\rho)$ with
	$1/2\leq\rho\leq3/4$, which importantly do not depend on any
	preference gap between actions $\Delta$. Dueling Bandits is an
	important extension of the Multi-Armed Bandit problem, in
	which the algorithm must select two actions at a time and only
	receives binary feedback for the duel outcome. This is
	analogous to comparisons in which the rater can only provide
	yes/no or better/worse type responses. We compare our simple
	algorithms to the current state-of-the-art for Dueling
	Bandits, ISS and DTS, discussing complexity and regret upper
	bounds, and conducting experiments on synthetic data that
	demonstrate their regret performance, which in some cases
	exceeds state-of-the-art. 
\end{abstract}

\section{Introduction}

Dueling Bandits, first proposed in \cite{yue2009interactively}, is an
important variation on the Multi-Armed Bandit (MAB), a well-known
online machine learning problem that has been studied extensively by
many previous works, such as \cite{auerfinite2002},
\cite{cesa2006prediction}, and \cite{bubeck2012regret}. Dueling
Bandits is different from MAB in that it provides binary feedback at
each time, the win/lose outcome of a duel between two actions. This
corresponds well to comparisons between two system states that receive
better/worse type responses from users, patients, raters,
and so on. Previous work on this topic has proposed various algorithms
that generally allow for regret bounds of the order $\log T/\Delta$ to
be proven, where $\Delta$ represents the preference gap between two
different states (or actions). See \cite{sui2018advancements} for a
reference. Such algorithms include, Beat the Mean \cite{yue2011beat},
Interleaved Filter \cite{yue2012k}, SAVAGE \cite{urvoy2013generic},
RUCB \cite{zoghi2013relative} and RCS \cite{zoghi2014relative},
MultiSBM and Sparring \cite{ailon2014reducing}, Sparse Borda \cite{jamieson2015sparse},
RMED \cite{komiyama2015regret}, CCB \cite{zoghi2015copeland}, and
(E)CW-RMED \cite{komiyama2016copeland}. Thompson Sampling, first
proposed in \cite{thompson1933likelihood}, is a powerful method of
learning true parameters values $\theta$, by sampling from
a posterior distribution using Bayes Theorem. See
\cite{russo2014learning} and \cite{russo2018tutorial} for
reference. It has been implemented in algorithms for multi-armed bandits,
such as in \cite{chapelle2011empirical}, \cite{agrawal2012analysis},
\cite{kaufmann2012thompson}, \cite{agrawal2013further},
\cite{komiyama2015optimal}, and \cite{xia2015thompson}. The current
state-of-the-art algorithms for Dueling Bandits both utilize Thompson
Sampling methods, Independent Self-Sparring (ISS) \cite{sui2017multi}
and Double Thompson Sampling (DTS)
\cite{wu2016double}. The ISS method is relatively simple, has strong empirical performance,
and has been proven to converge asymptotically to a Condorcet winner,
if one exists. However, its non-asymptotic regret has not been
analyzed. The DTS algorithm is a relatively complex algorithm with a
highly complex proof. It achieves regret of order $\log T/\Delta$. However, the worst-case $\Delta$
values, lead to regret bounds that are actually of
order $\sqrt{T \log T}$. We address these issues in this paper, with our main
contributions: (1)we present four simple algorithms for Dueling
Bandits, each of which allows provable upper bounds on regret of order
$O(T^\rho)$ with $1/2\leq\rho\leq3/4$ that do not depend on any
preference gap $\Delta$ between actions, (2) we compare and contrast
the algorithm complexity and theoretical results of the presented
simple algorithms against the current state-of-the-art algorithms for
Dueling Bandits, and (3) we evaluate the algorithms on multiple
scenarios using synthetically generated data, demonstrating their
performance for multiple definitions of optimality, that in some cases
exceeds the state-of-the-art.

\section{Background}

\subsection{Dueling Bandits}
\begin{algorithm}[H]
	\floatname{algorithm}{Problem}
	\caption{\label{DBprob} Dueling Bandits}
	\begin{algorithmic}[0]
		\For{$t\leq T$}
		\State Environment randomly draws matrix of duel
		outcomes $\X_t$ 
		\State Select actions $\I_t$ and $\J_t$
		\State Observe outcome $\X_t(\I_t,\J_t)$
		\EndFor
	\end{algorithmic}
\end{algorithm}

The dueling bandits problem is described in
Problem~\ref{DBprob}. The random matrices $\X_1,\ldots,\X_T \in
\{0,1\}^{A\times A}$ are independent and identically distributed. Each
element is Bernoulli distributed such that 
$\P[\X_t(i,j)=1]=\P[i\succ j]$ denotes the probability of action $i$
winning a duel with action $j$.

For Thompson sampling algorithms, we
will assume that the win probabilities depend on an unobserved random
parameter, $\btheta$, so that $\P[\X_t(i,j)=1 | \btheta] =
\X(i,j)$. The parameter can be used to encode correlations between the
actions and other structural assumptions.

For algorithms based on Exp3.P and partial monitoring, we assumes that
$\P[\X_t(i,j)=1] = X(i,j)$, where $X$ is a fixed but unknown
matrix of win probabilities.

We assume that $\X_t(i,j) = 1-\X_t(j,i)$ when $i\ne j$  and that
$\X(i,i) = 1/2$ or $X(i,i) = 1/2$, depending on the problem setup. 

Random variables $I_t$ and
$J_t$ represent the actions selected to duel at each time, and we
denote $\H_t =
\{\I_\tau,\J_\tau,\X_\tau(\I_\tau,\J_\tau)\}_{\tau=1}^{t-1}$ as the
available \emph{history} to help guide the selections. Note that the
assumptions about $\X_t$ imply that if $\X_t(\I_t,\J_t)$ is observed,
then $\X_t(\J_t,\I_t)$ is also known.

\subsection{Optimal Actions}

It is assumed that there is a sub-set of optimal actions within
$\{1,\dots,A\}$, and that we wish to find an optimal action as
efficiently as possible. There are several optimality notions used for dueling bandits. We discuss some of these below, and note that section 4.1 of \cite{sui2018advancements} provides additional definitions.

\subsubsection{Copeland and Condorcet Winners}
The standard definition of optimal actions in dueling bandits literature are Copeland and Condorcet winners. These rely on counting the number of other actions that a particular action is likely to beat in a duel (in the sense of $\P[i\succ j]=X(i,j)>0.5$). Copeland winners $i^*_C$ are defined as,
\begin{align*}
i^*_C \in \argmax_i \sum_j \mathbbm{1}\big[X(i,j)>0.5\big]
\end{align*}
If there is a single action that is likely to beat all other actions, this is known as a Condorcet winner. Copeland winners always exist, even if a Condorcet winner does not exist.

\subsubsection{Maximin and Borda Winners}
In this paper, we focus on two alternatives to Copeland and Condorcet winners for defining optimal actions: Maximin winners and Borda winners. Both rely on simpler measures of $X$ to determine the optimal actions. Maximin winners use row minimum values of $X$, and Borda winners use row average values of $X$. Let us define Maximin winners $i^*_M$ and Borda winners $i^*_B$ as,
\begin{align*}
i^*_M \in \argmax_i \min_j X(i,j) \qquad\qquad i^*_B \in \argmax_i \frac{1}{A}\sum_j X(i,j)
\end{align*}
Maximin and Borda winners both always exist, even if a Condorcet winner does not exist. Also, Copeland winners are not guaranteed to align with either Maximin or Borda winners. Condorcet winners are guaranteed to align with Maximin winners, but not with Borda winners. For these reasons, we find these to be compelling alternative definitions for optimal actions.

\subsection{Regret}
To characterize the performance of the selected actions over time horizon $T$, we can compare them against ideal selections that could have been made over that time period. This is known as regret. While it may be intuitive that an ideal sequence of $\I_t$ selections would be any $\I_1,\dots,\I_T$ which maximizes $\sum_{t=1}^T \X_t(\I_t,\J_t)$, for a given sequence of $\J_1,\dots,\J_T$ selections (and vice versa, minimizes it for ideal $\J_t$ selections), this is unreasonable and not possible. Selections are unknown prior to a duel, and adaptations to selection strategies are made after a duel, meaning the original given selection sequence would no longer be valid. Instead, a reasonable ideal sequence of selections that could have been made is for both $\I_t$ and $\J_t$ to have been optimal actions, at all times. Therefore, if the regret incurred over time horizon $T$ is minimized, then the selected actions have converged to optimal actions as efficiently as possible in that time period.

\section{Algorithms}

\subsection{Thompson Sampling for Dueling Bandits}

We describe Thompson Sampling in generality, in order to highlight its
flexibility. It learns true parameter values $\btheta$, which can
represent $\X$ directly or some other latent values for each action,
by sampling the posterior distribution conditioned on the history $\H_t$. The
samples of $\btheta$ become more accurate as the information in $\H_t$
increases, and are used to form an estimate of $\X$, which can be used with any optimal action
definition. We present algorithms for both Maximin winners
(Alg. \ref{algTSmax}) and Borda winners (Alg. \ref{algTSborda}). 

An appropriate prior distribution over $\btheta$ must be chosen so
that the posterior distribution can either be determined analytically
or sampled from by using computational means (such as Markov chain
Monte Carlo). The prior can be used to model correlations between
actions, for example by using a Gaussian Process. 
\begin{algorithm}[H]
	\renewcommand{\thealgorithm}{1}
	\caption{\label{algTSmax} Thompson Sampling for Dueling Bandits with Maximin Winners}
	\begin{algorithmic}[1]
		\Statex \textbf{Input:} prior distribution $p(\theta)$
		\Statex \textbf{Init:} $\H_1 = \emptyset$
		\For{$t\leq T$}
		\State Environment draws $\X_t$ according to unknown $X$
		\State Independently draw $\btheta_I,\btheta_J \sim p(\theta|\H_t)$
		\State Select actions $\I_t \in \argmax_i \min_j \E[\X(i,j)|\btheta_I],\J_t \in \argmax_j \min_i \E[\X(j,i)|\btheta_J]$
		\State Observe outcome $\X_t(\I_t,\J_t)$
		\State Append observation to history $\H_{t+1} = \H_t \cup \{(\I_t,\J_t,\X_t(\I_t,\J_t))\}$
		\EndFor
	\end{algorithmic}
\end{algorithm}
\begin{algorithm}[H]
	\renewcommand{\thealgorithm}{2}
	\caption{\label{algTSborda} Thompson Sampling for Dueling Bandits with Borda Winners}
	\begin{algorithmic}[1]
		\Statex \textbf{Input:} prior distribution $p(\theta)$, $0<\alpha<1$
		\Statex \textbf{Init:} $\H_1 = \emptyset$
		\For{$t\leq T$}
		\State Environment draws $\X_t$ according to unknown $X$
		\State Draw $\b\sim Bernoulli(\alpha)$
		\If{$\b = 1$}
		\State Select actions $\I_t$ and $\J_t$ uniformly at random
		\Else
		\State Independently draw $\btheta_I,\btheta_J \sim p(\theta|\H_t)$
		\State Select actions $\I_t \in \argmax_i \sum_j \E[\X(i,j)|\btheta_I], \J_t \in \argmax_j \sum_i \E[\X(j,i)|\btheta_J]$
		\EndIf
		\State Observe outcome $\X_t(\I_t,\J_t)$
		\State Append observation to history $\H_{t+1} = \H_t \cup \{(\I_t,\J_t,\X_t(\I_t,\J_t))\}$
		\EndFor
	\end{algorithmic}
\end{algorithm}

\subsection{SparringExp3.P for Dueling Bandits}

SparringExp3.P is implemented for dueling bandits in Algorithm \ref{algExp3P}, and is inspired by the methods in \cite{ailon2014reducing} and \cite{dudik2015contextual}. It learns from the previous duel outcomes and accordingly adjusts the strategies $\p_{t+1}$ and $\q_{t+1}$ using hyperparameters $\eta>0$ and $0<\gamma<1$. For all times $t\leq T$ and all actions $i,j\in\{1,\dots,A\}$, the update equations are,
\begin{align}
\label{eq:expWeights}
\wb_{p,t}(i) &= \, \wb_{p,t-1}(i)\, \exp(\eta\,
\widetilde{\X}_{p,t}(i,\J_t)) &\wb_{q,t}(j) &= \,
\wb_{q,t-1}(j)\,
\exp(\eta\,
\widetilde{\X}_{q,t}(j,\I_t))
\\
\label{eq:expProbs}
\p_{t+1}(i) &= (1-\gamma)\, \frac{\wb_{p,t}(i)}{\W_{p,t}}\ +\
\gamma\, \frac{1}{A}
& \q_{t+1}(j) &= (1-\gamma)\, \frac{\wb_{q,t}(j)}{\W_{q,t}}\ +\ \gamma\, \frac{1}{A}
\end{align} 
Since only outcome $\X_t(\I_t,\J_t)$ is revealed at each time $t$, the other outcomes in the corresponding rows of $\X_t$ must be estimated. These estimates are made using the observed outcome and hyperparameter $0\leq\beta\leq1$ as follows,
\begin{equation}
\label{eq:expEstimate}
\widetilde{\X}_{p,t}(i,\J_t) =
\frac{\X_t(i,\J_t)\,\mathbbm{1}(i=\I_t) \ + \ \beta}{\p_t(i)}
,\quad 	\widetilde{\X}_{q,t}(j,\I_t) = \frac{\X_t(j,\I_t)\,\mathbbm{1}(j=\J_t) \ + \ \beta}{\q_t(j)}
\end{equation}
for all $i,j \in \{1,\ldots,A\}$. 
These estimates satisfy 
$\E[\widetilde{\X}_{p,t}(i,\J_t)|\J_t,\H_t] = \X_t(i,\J_t)+\beta/\p_t(i)
$ and $\E[\widetilde{\X}_{q,t}(j,\I_t)|\I_t,\H_t] = \X_t(j,\I_t) + \beta/\q_t(j)$ for all $i,j$ and all times $t$.
\begin{algorithm}[H]
	\renewcommand{\thealgorithm}{3}
	\caption{\label{algExp3P} SparringExp3.P for Dueling Bandits}
	\begin{algorithmic}[1]
		\Statex \textbf{Input:} $\beta,\eta,\gamma$
		\Statex \textbf{Init:} $\wb_{p,0}(i) = \wb_{q,0}(j) = 1$ and $\p_1(i) = \q_1(j)= 1/A$, for all $i,j$
		\For{$t\leq T$}
		\State Environment draws $\X_t$ according to unknown $X$
		\State Independently draw actions $\I_t\sim \p_t$ and $\J_t\sim \q_t$
		\State Observe outcome $\X_t(\I_t,\J_t)$
		\For{$i,j\in\{1,\dots,A\}$}
		\State Calculate estimates
		$\widetilde{\X}_{p,t}(i,\J_t)$ and
		$\widetilde{\X}_{q,t}(j,\I_t)$ as in \eqref{eq:expEstimate}
		\State Update weights $\wb_{p,t}(i)$ and $\wb_{q,t}(j)$ as
		in \eqref{eq:expWeights}
		\EndFor
		\State Calculate weight sums $\W_{p,t}=\sum_i \wb_{p,t}(i)$ and $\W_{q,t}= \sum_j \wb_{q,t}(j)$
		\For{$i,j\in\{1,\dots,A\}$}
		\State Update strategies $\p_{t+1}(i)$ and $\q_{t+1}(j)$
		as in \eqref{eq:expProbs}
		\EndFor
		\EndFor
	\end{algorithmic}
\end{algorithm}

\subsection{Partial Monitoring Forecaster for Dueling Bandits}

The Partial Monitoring forecaster \cite{cesa2006prediction} is implemented for dueling bandits in Algorithm \ref{algPM}. The forecaster learns from the previous duel outcomes and accordingly adjusts the strategy $\p_{t+1}$ using hyperparameters $\eta>0$ and $0<\gamma<1$. For all times $t\leq T$ and all actions $i\in\{1,\dots,A\}$, the update equations are,
\begin{align}
\wb_t(i) &= \, \wb_{t-1}(i)\, \exp(\eta\, \widetilde{\gb}_t(i)) \label{eq:PMWeights}\\
\p_{t+1}(i) &= (1-\gamma)\, \frac{\wb_t(i)}{\W_t}\ +\ \gamma\, \frac{1}{A} \label{eq:PMProb}
\end{align} 
Since only outcome $\X_t(\I_t,\J_t)$ is revealed at each time $t$, the Borda score for $\X_t$, must be estimated using the observed outcome as follows,
\begin{align}
\gb_t(i) = \frac{1}{A}\sum_{k=1}^A\, \X_t(i,k) \qquad	\widetilde{\gb}_t(i) = \frac{1}{A}\frac{\X_t(i,\J_t)\,\mathbbm{1}(i=\I_t)}{\p_t(\I_t)\,\p_t(\J_t)} \label{eq:PMEstimate}
\end{align}
for all $i\in\{1,\dots,A\}$. These estimates satisfy $\E[\widetilde{\gb}_t(i)|\H_t] = \gb_t(i)$ for all $i$ and all times $t$.
\begin{algorithm}[H]
	\renewcommand{\thealgorithm}{4}
	\caption{\label{algPM} Partial Monitoring Forecaster for Dueling Bandits}
	\begin{algorithmic}[1]
		\Statex \textbf{Input:} $\eta,\gamma$
		\Statex \textbf{Init:} $\wb_{p,0}(i) = 1$ and $\p_1(i) = 1/A$, for all $i$
		\For{$t\leq T$}
		\State Environment draws $\X_t$ according to unknown $X$
		\State Independently draw actions $\I_t$ and $\J_t$ $\ \sim \p_t$
		\State Observe reward $\X_t(\I_t,\J_t)=1-\X_t(\J_t,\I_t)$
		\For{$i\in\{1,\dots,A\}$}
		\State Calculate estimate $\widetilde{\gb}_t(i)$ as in \eqref{eq:PMEstimate}
		\State Update weight $\wb_t(i)$ as in \eqref{eq:PMWeights}
		\EndFor
		\State Calculate weight sum $\W_t = \sum_i \wb_t(i)$
		\For{$i\in\{1,\dots,A\}$}
		\State Update strategy $\p_{t+1}(i)$ as in \eqref{eq:PMProb}
		\EndFor
		\EndFor
	\end{algorithmic}
\end{algorithm}

\subsection{Comparison to State-of-the-Art}
Both state-of-the-art dueling bandits algorithms ISS \cite{sui2017multi} and DTS \cite{wu2016double} use variations of specific Thompson Sampling implementations. They both use $Beta(1,1)$ as prior distributions $p(\theta_n)$, for each independent, true $\theta_n$ value they attempt to learn. Since Beta distributions are conjugate pairs with Bernoulli likelihoods, the independent posterior distributions $p(\theta_n|\H_t)$ are able to be determined analytically and are themselves Beta distributions.

While the ISS algorithm is very simple, it does not learn an estimate for $X$. Instead, it learns the more basic overall probability of each action winning a duel with a Concorcet winner. It therefore learns $A$ independent $\theta_n$ values, one for each action. Since it does not learn $X$, it cannot learn to track a Borda winner unless it is also the Condorcet winner.

The DTS algorithm does learn an estimate of $X$. It thus learns $A^2$ independent $\theta_n$ values, one for each $i,j$ pair in $X$. However, it is a complex and specialized algorithm that tracks the Copeland winner, so it cannot learn to track a Borda winner unless it is also the Copeland winner.

\section{Theoretical Results}

In this section, we will present theorems that upper bound the
regret for each of the algorithms described in the previous section,
and also compare the bounds to those for the current
state-of-the-art. Each of the regret upper bounds is of the order
$O(T^\rho)$ with $1/2\leq\rho\leq3/4$, and this bound holds regardless
of the size of any preference gaps between any two actions
$\Delta$. All definitions of regret are normalized, such that the
regret incurred at any time $t$ satisfies $\rb_t\leq1$, and therefore
$\R_T=\sum_{t=1}^T \rb_t \leq T$. Detailed proofs are provided in the appendix.

\paragraph{Theorem 4.1} \textit{Let us define regret over time horizon $T$ in the sense of Maximin winner $\i^*_M$,}
\begin{align*}
\R_T = \frac{1}{2}\sum_{t=1}^T \Big(\X_t(\i^*_M,\I_t) - \X_t(\i^*_M,\i^*_M) + \X_t(\i^*_M,\J_t) - \X_t(\i^*_M,\i^*_M)\Big)
\end{align*}
\textit{Then, if actions $\I_t,\J_t$ are selected at each time using Thompson Sampling for Dueling Bandits with Maximin winners (Alg. \ref{algTSmax}), the expected regret is upper bounded as,}
\begin{equation*}
\E\left[\R_T\right]\ \leq\  \frac{A}{\sqrt{2}}\sqrt{\log A}\,\sqrt{T}
\end{equation*}
The proof method is a variation on the worst case bound from
\cite{russo2016information}.

\paragraph{Theorem 4.2} \textit{Let us define regret over time horizon $T$ in the sense of Borda winner $\i^*_B$,}
\begin{align*}
\R_T = \frac{1}{2A}\sum_{t=1}^T\sum_{k=1}^A \Big(\X_t(\i^*_B,k) - \X_t(\I_t,k) + \X_t(\i^*_B,k) - \X_t(\J_t,k)\Big)
\end{align*}
\textit{Then, if actions $\I_t,\J_t$ are selected at each time using Thompson Sampling for Dueling Bandits with Borda winners (Alg. \ref{algTSborda}), using $\alpha = c\,T^{-1/3} < \frac{1}{2}$ for $c>0$, the expected regret is upper bounded as,}
\begin{equation*}
\E\left[\R_T\right]\ \leq\  \left(c + \sqrt{\frac{A}{c}\log A}\right)\,T^{2/3}
\end{equation*}
The proof method uses the same concepts from
\cite{russo2016information} as the proof of Theorem 4.1. 

\paragraph{Theorem 4.3} \textit{Let us define regret over time horizon $T$ in the sense of Maximin winner $i^*_M$,}
\begin{align*}
\R_T = \frac{1}{2}\sum_{t=1}^T \Big(\X_t(i^*_M,\J_t) - \X_t(\I_t,\J_t) + \X_t(i^*_M,\I_t) - \X_t(\I_t,\J_t)\Big)
\end{align*}
\textit{Then, if actions $\I_t,\J_t$ are selected at each time using SparringExp3.P for Dueling Bandits (Alg. \ref{algExp3P}), with hyperparameter values of,}
\begin{align*}
\beta = \sqrt{\frac{\log A}{AT}} \qquad \eta = 0.95\,\sqrt{\frac{\log A}{AT}} \qquad \gamma = 1.05\,\sqrt{\frac{A\,\log A}{T}}
\end{align*}
\textit{and $T$ satisfying,}
\begin{align*}
T\ \geq\ \max\left[4.41\,A\log A\ ,\ \frac{0.95^2\log A}{0.1^2\,A}\right]
\end{align*}
\textit{the expected regret is upper bounded as,}
\begin{align*}
\E[\R_T] \, \leq \,  \bigg(\sqrt{A\,(\log A)^{\text{-}1}} + 4.2\sqrt{A\log A} \bigg)\sqrt{T}
\end{align*}
The proof method follows those used for lemma 3.1 and theorems 3.2 and 3.3 in \cite{bubeck2012regret}.

\paragraph{Theorem 4.4} \textit{Let us define regret over time horizon $T$ in the sense of Borda winner $i^*_B$,}
\begin{align*}
\R_T = \frac{1}{2A}\sum_{t=1}^T\sum_{k=1}^A \Big(\X_t(i^*_B,k) - \X_t(\I_t,k) + \X_t(i^*_B,k) - \X_t(\J_t,k)\Big)
\end{align*}
\textit{Then, if actions $\I_t,\J_t$ are selected at each time using the Partial Monitoring Forecaster for Dueling Bandits (Alg. \ref{algPM}), with hyperparameter values of,}
\begin{align*}
\eta = (e-2)^{-1/4}\,\bigg(\frac{\log A}{A^{2/3}\,T}\bigg)^{3/4} \qquad \gamma = (e-2)^{1/4}\,\bigg(\frac{A^2\log A}{T}\bigg)^{1/4}
\end{align*}
\textit{and $T$ satisfying,}
\begin{align*}
T\ >\ (e-2)\,A^2\,\log A
\end{align*}
\textit{the expected regret is upper bounded as,}
\begin{equation*}
\E\left[\R_T\right]\ \leq\  \ 2\, (e-2)^{1/4}\,\sqrt{A\,(\log A)^{1/2}}\ \, T^{3/4}
\end{equation*}
The proof method follows those used for theorem 6.5 in
\cite{cesa2006prediction}.

\subsection{Comparison to State-of-the-Art}
Many works on dueling bandits assume that a Condorcet winner, $i^*_C$, exists. In this case, $X(i^*_C,j) > 1/2$ for all $j\ne i_C$, and let $\Delta = \min_{j\ne i_C} X(i_C,j) - 1/2$ be the \emph{preference gap} between the Condorcet winner and the next best action. This commonly allows regret bounds of $O\left(\frac{\log T}{\Delta}\right)$ to be proven. These bounds appear to be superior to the $O(\sqrt{T})$ bounds derived in this paper. However, as discussed in \cite{bubeck2012regret} (and others), when $\Delta$ is small, the 
$(\log T) / \Delta$ bound becomes smaller than the regret for selecting the sub-optimal action each time, which is $\Delta T$. Therefore, taking a worst-case value over $\Delta$ leads to an actual regret bound of $O(\sqrt{T\log T})$, which is not superior to the $O(\sqrt{T})$ bounds we show. 

This is the case for both state-of-the-art methods ISS \cite{sui2017multi} and DTS \cite{wu2016double}. Furthermore, we note that the proof for ISS demonstrates only asymptotic convergence to a Condorect winner, while the proof for DTS is highly complex (owing the relatively complex nature of the algorithm). In comparison, the proofs available in appendix A are relatively simple (though presented in a detailed manner).

\section{Experimental Results}

\subsection{Methods}
We simulate each of the proposed algorithms, along with the two state-of-the-art algorithms ISS \cite{sui2017multi} and DTS \cite{wu2016double}, on two different scenarios using synthetic data. For the Thompson Sampling methods, we use $A^2 - A$ independent $Beta(1,1)$ priors for the $X$ values we attempt to learn. We set $\E[\X(i,i)|\btheta]=0.5$ directly, for all $i$. In the Condorcet scenario, an $X$ matrix is synthetically generated by linking a latent value for each action (called ``utility") to the duel winning probability $\P[i\succ j]=X(i,j)$ for each pair of actions $i,j\in\{1,\dots,A\}$. The utility of each action, $u(i)$, is uniformly distributed between $0$ and $c>0$. We chose $c=3$ to give a larger spread of probabilities over the actions. One action has a maximum utility, that is significantly better than all other actions, and so it is the lone Borda winner and Condorcet winner, and thus also the lone Maximin winner. Linking the utility of each pair of actions to the corresponding duel winning probability is accomplished by using the logistic function on the gap between utilities of the actions,
\begin{align*}
\P[i\succ j]=X(i,j) = \frac{1}{1 + \exp(u(j) - u(i))}
\end{align*}
In the Borda scenario, we modify the previous $X$ matrix such that the action with the second largest utility $i_2$ becomes the lone Borda winner, even though the same Condorcet and Maximin winner still exists. This is done by setting $X(i_2,j)=0.95$ for all $j\ne i_2$ other than the Condorcet winner. This aptly represents why the Borda winner is a reasonable definition for optimality. Even though it isn't likely to beat every action, it is the most likely to beat an action drawn at random. Each algorithm runs with a time horizon of $T=40,000$ iterations, for $100$ separate runs, on each scenario.

\subsection{Results}
\begin{figure}[H]
	\centering
	\begin{subfigure}[b]{0.35\textwidth}
		\centering
		\includegraphics[width=\textwidth]{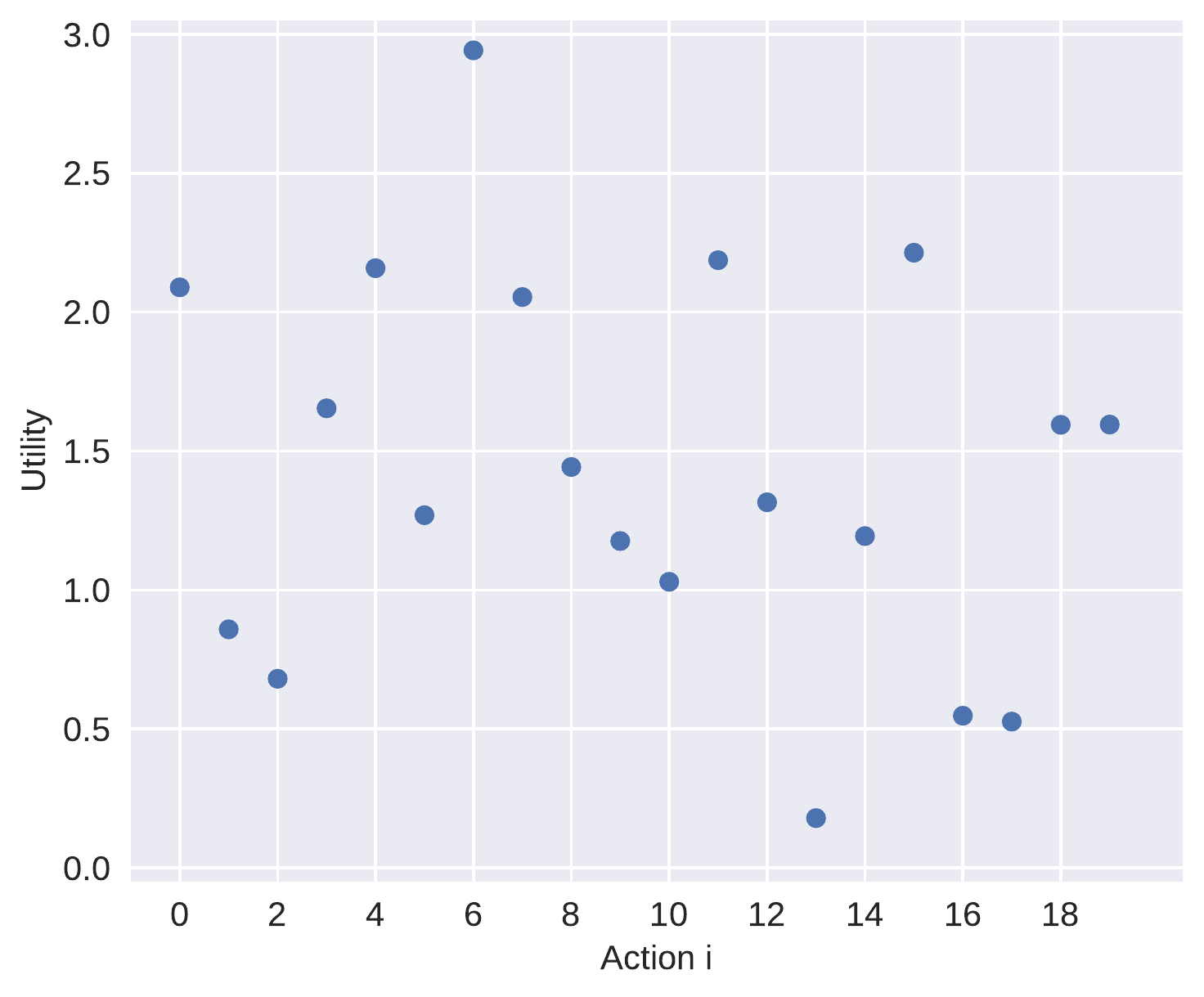}
		\subcaption{}
	\end{subfigure}
	\begin{subfigure}[b]{0.35\textwidth}
		\centering
		\includegraphics[width=\textwidth]{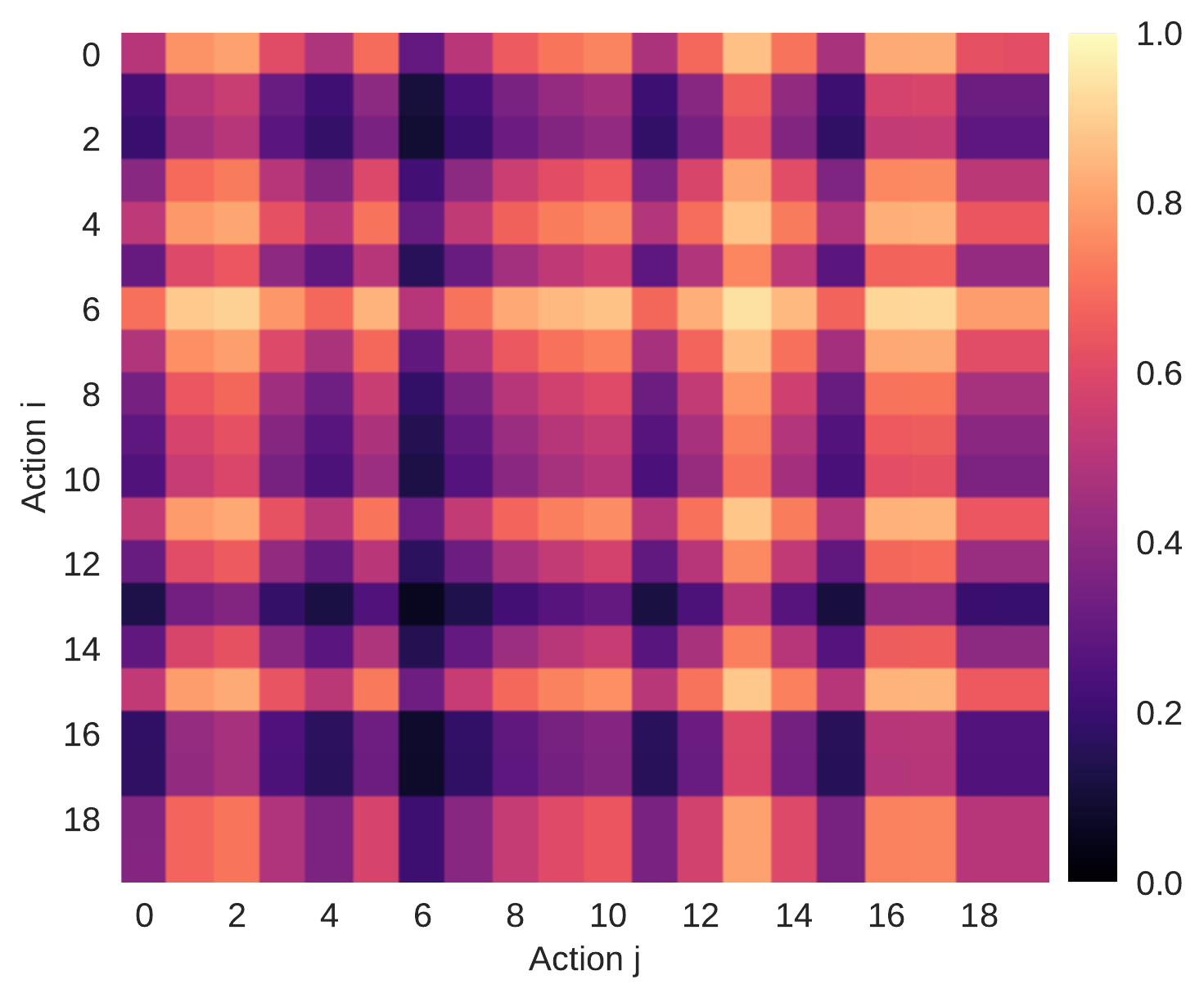}
		\subcaption{}
	\end{subfigure}\\
	\centering
	\begin{subfigure}[b]{0.4\textwidth}
		\centering
		\includegraphics[width=\textwidth]{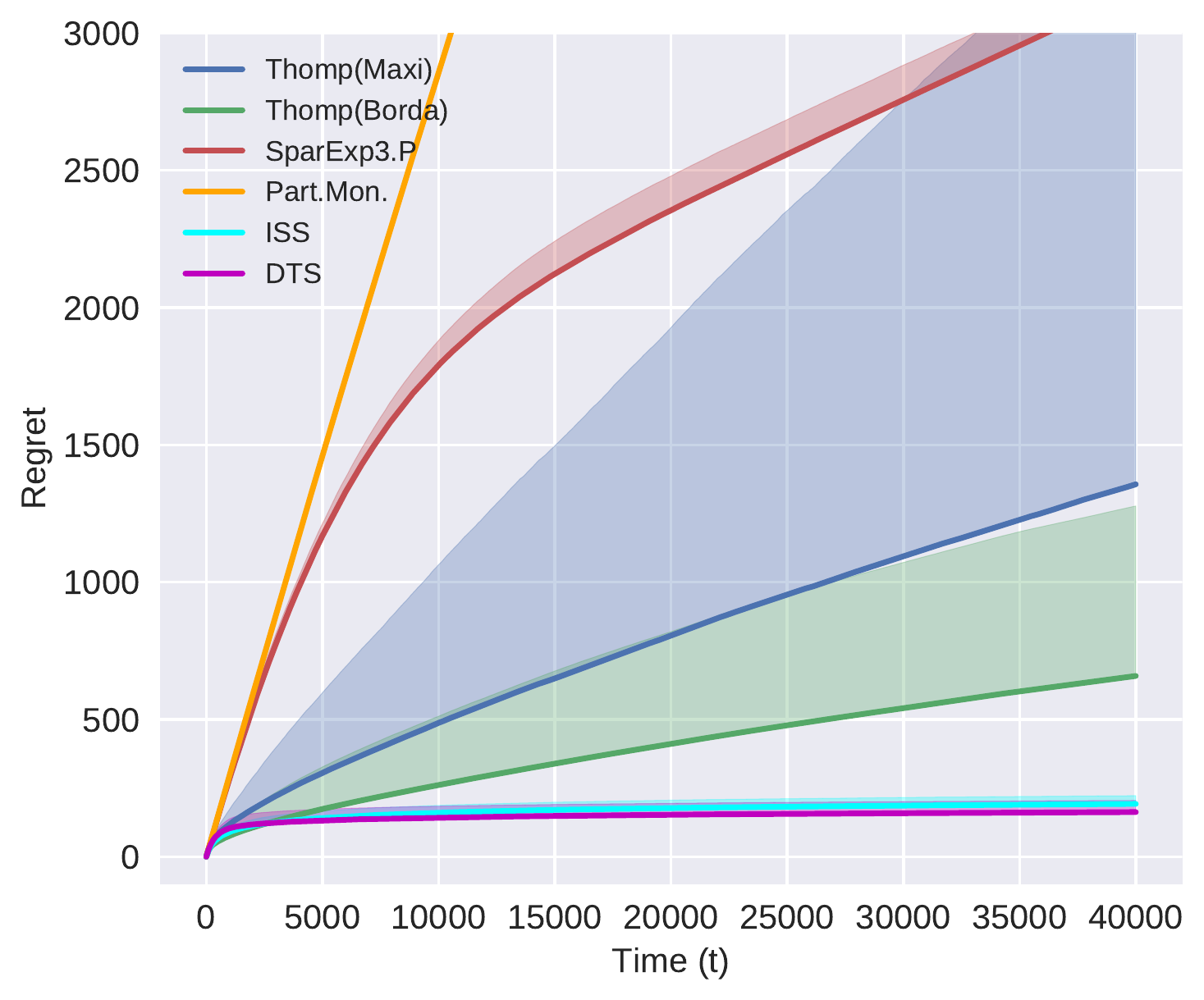}
		\subcaption{}
	\end{subfigure}
	\begin{subfigure}[b]{0.4\textwidth}
		\centering
		\includegraphics[width=\textwidth]{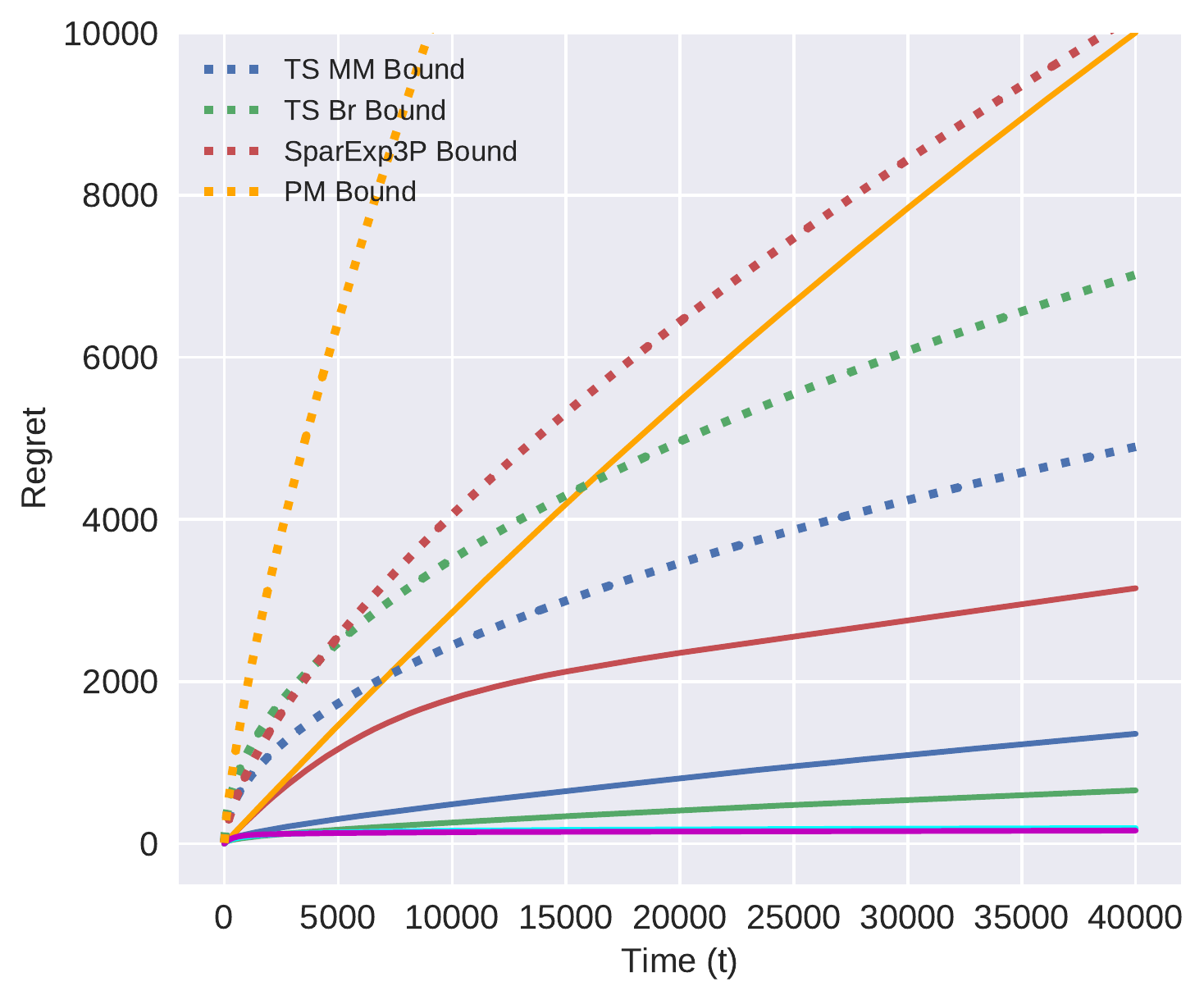}
		\subcaption{}
	\end{subfigure}
	\caption{\label{figScen1} {\bf Condorcet Scenario} (a) Latent "utility" values for all $i$, (b) linked $X(i,j)$ values for all $i,j$ , (c) algorithm mean regrets over scenario runs, and (d) with bounds included. }
\end{figure}

\begin{figure}[H]
	\centering
	\begin{subfigure}[b]{0.35\textwidth}
		\centering
		\includegraphics[width=\textwidth]{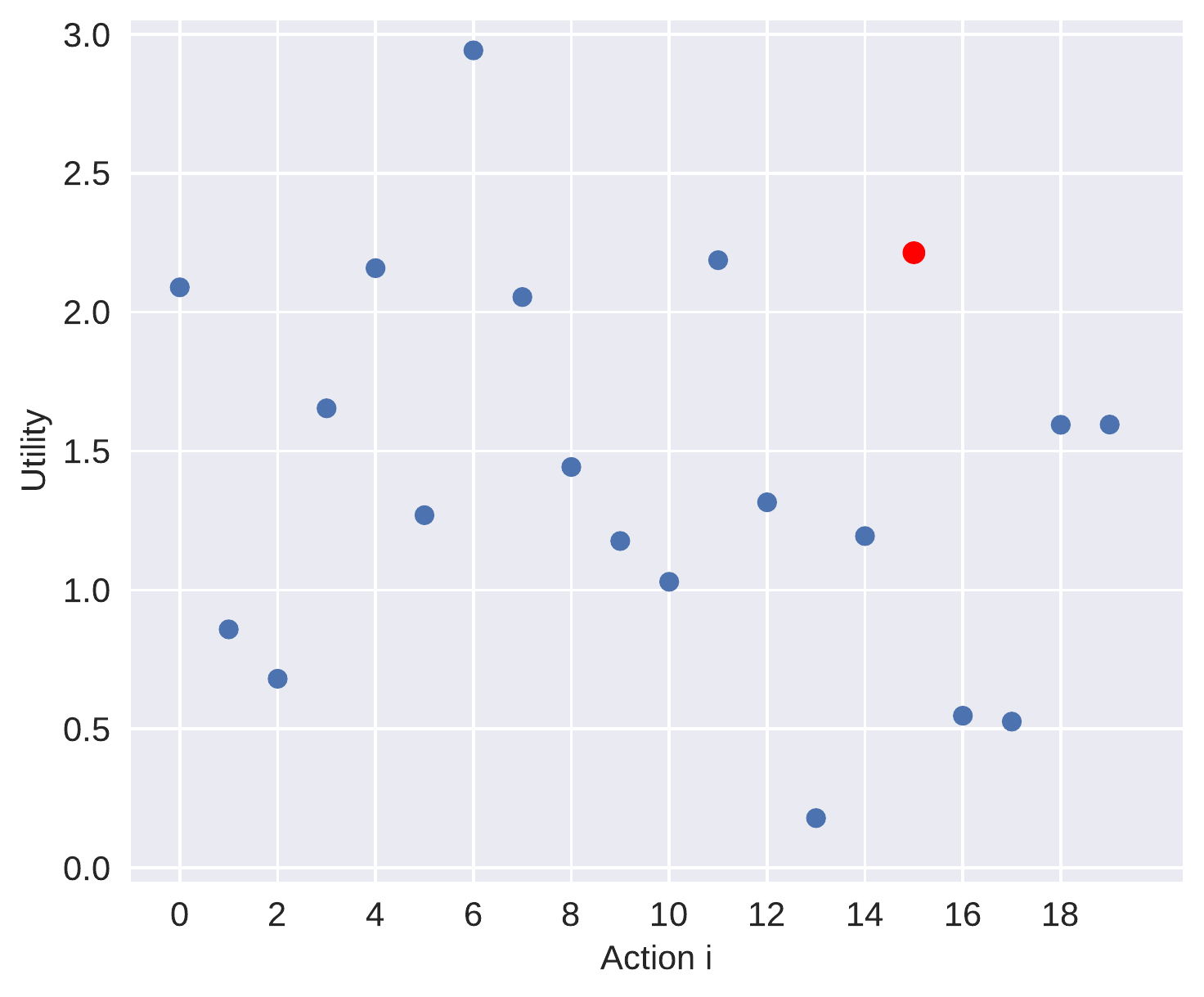}
		\subcaption{}
	\end{subfigure}
	\begin{subfigure}[b]{0.35\textwidth}
		\centering
		\includegraphics[width=\textwidth]{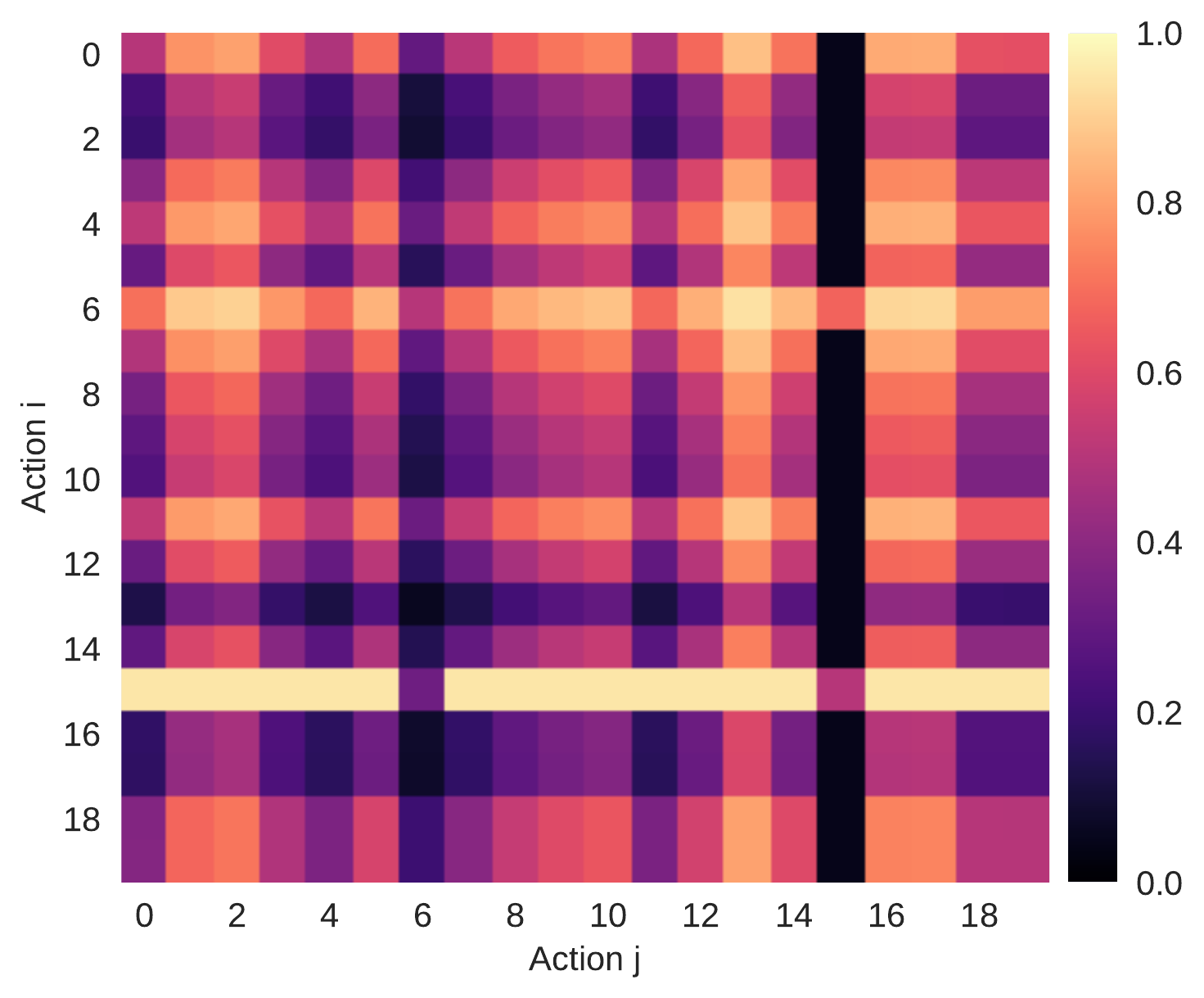}
		\subcaption{}
	\end{subfigure}\\
	\centering
	\begin{subfigure}[b]{0.4\textwidth}
		\centering
		\includegraphics[width=\textwidth]{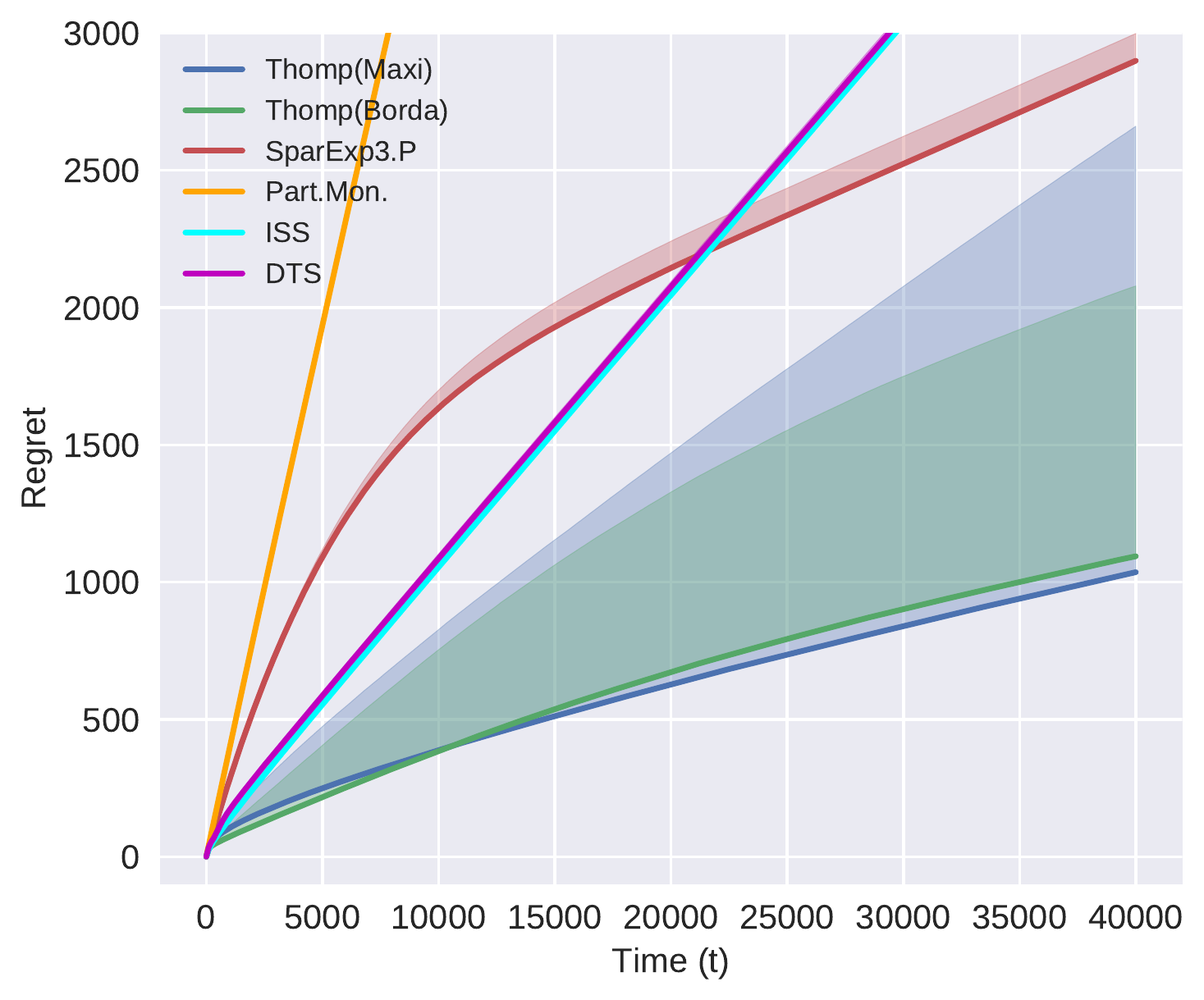}
		\subcaption{}
	\end{subfigure}
	\begin{subfigure}[b]{0.4\textwidth}
		\centering
		\includegraphics[width=\textwidth]{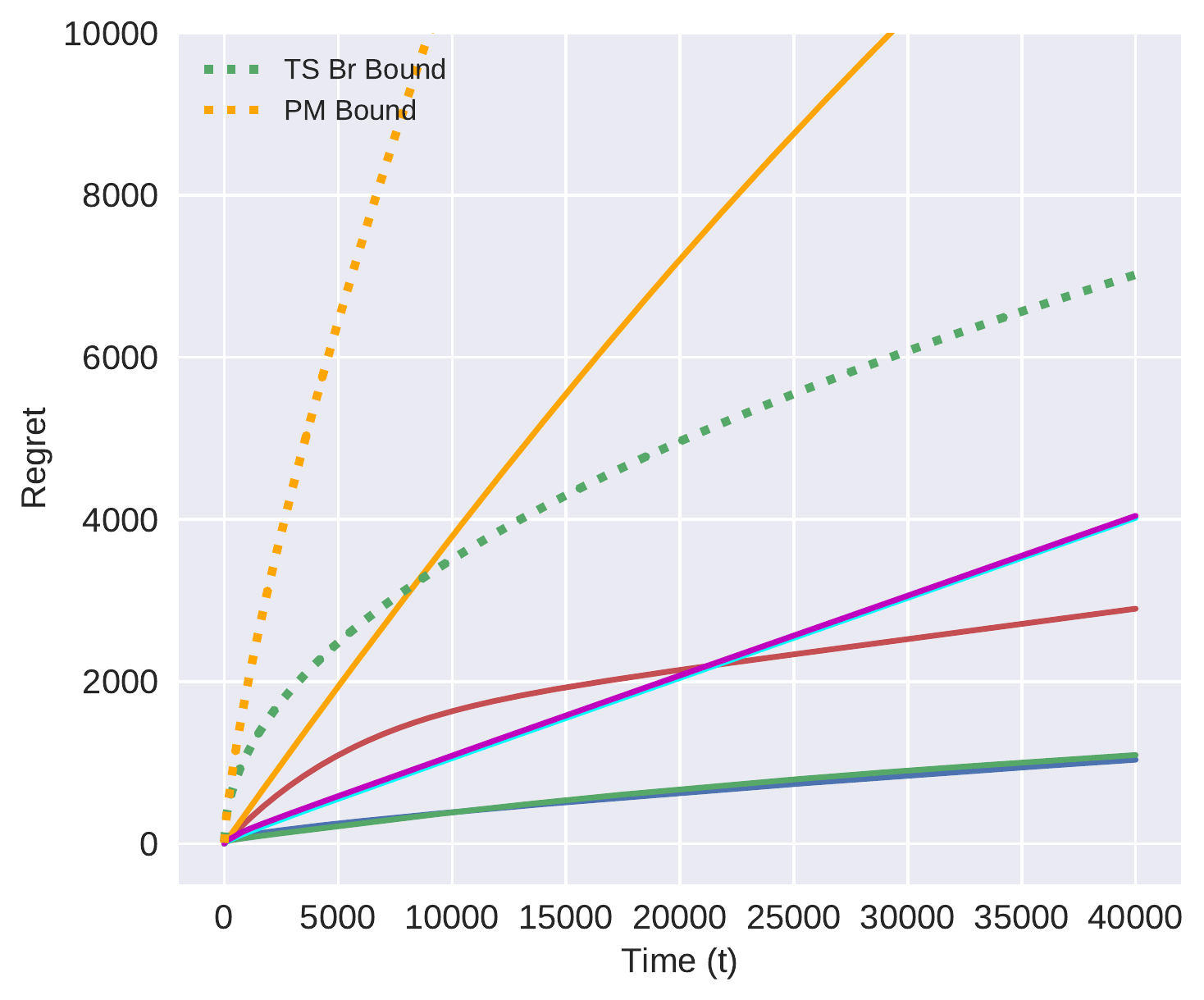}
		\subcaption{}
	\end{subfigure}
	\caption{\label{figScen2} {\bf Borda Scenario} (a) Latent "utility" values for all $i$, (b) linked $X(i,j)$ values for all $i,j$ ,  (c) algorithm mean regrets over scenario runs, and (d) with bounds included for Borda algorithms. }
\end{figure}

The results of the Condorcet scenario are shown in Figure \ref{figScen1}, and the results of the Borda scenario are shown in Figure \ref{figScen2}. In both subfigure (c), a shaded area, plotted above the mean, shows the standard deviation over the runs. Additional detailed plots of each algorithm, for each scenario, are available in appendix B. In the Condorcet scenario, the regret for each algorithm is as prescribed in the respective theorem, and the regret for ISS and DTS use the Maximin winner (theorem 4.1). All formulations for regret are comparable, due to the scenario having the same winning action in all cases. Both state-of-the-art methods show very strong regret performance. However, the Thompson Sampling with Borda winners method shows comparably strong performance, with other methods also performing well. All methods beat the regret upper bounds proposed in their respective theorems. In the Borda scenario, the regret for all algorithms (including ISS and DTS) uses the Borda winner. This is to highlight the fact that some of the methods are not capable of performing well in this type of scenario. Both state-of-the-art methods struggle with Borda winners, and so their Borda regret grows linearly. A similar behavior ultimately happens to SparringExp3.P (more details available in the appendix). Thompson Sampling shines in this case. Both methods that focus on Borda winners are able to beat their respective regret upper bounds.

\section{Conclusion}
In this paper, we have presented four simple algorithms for Dueling Bandits, each of which is able to efficiently find an optimal action within a finite set of available actions. We proved an upper bound on regret for each, over a variety of different optimal action types, such as the Borda Winner. The proven regret bounds were all of the order $O(T^\rho)$ with $1/2\leq\rho\leq3/4$, and did not depend on any preference gap between any two actions $\Delta_{ij}$. The algorithms were all evaluated and compared against the current state-of-the-art for Dueling Bandits, the ISS and DTS algorithms. While they did not meet or exceed the performance of ISS and DTS in certain scenarios, in others they demonstrated superior ability to find different types of optimal actions. Overall, their simplicity, regret bounds, and ability do merit inclusion with the current state-of-the-art.


\newpage

\newpage
\bibliography{../../CoOL-bib/cool-refs}
\bibliographystyle{plain}

\newpage

\appendix
\section{Theoretical Results}

In this section, we provide formal proofs for all theorems presented in the paper. All random variables and probability distributions use bold font.

\subsection{Proof of Theorem 4.1}
The proof method is a variation on the worst case bound from \cite{russo2016information}. \\\\
First, we make the following definitions: $\E_t$ is the expectation, $\P_t$ is the probability measure, $\p_t$ is the probability density, and $\Info_t(\cdot\ ;\ \cdot)$ is mutual information, all conditioned on the history $\Hb_t$, at time $t$. Furthermore, $D(\cdot\ ||\ \cdot)$ is the Kullback-Leibler divergence and $\Ent$ is entropy.\\\\
Then we note that Thompson Sampling selects both $\I_t$ and $\J_t$
using independent samples from the same posterior distribution
conditioned on $\Hb_t$. Therefore, $\I_t$ and $\J_t$ are independent
and identically distributed, and the terms $\X_t(\i^*_M,\I_t)$ and
$\X_t(\i^*_M,\J_t)$ are identically distributed.\\\\ 
Let $\rb_t$ be the instantaneous regret at time $t$, such that $\R_T = \sum_{t=1}^T \rb_t$.\\\\
We claim the following,
\begin{align}
\label{eq:regretExpand}
&\E_t\left[\rb_t\right] = \sum_{i^*,j} \P_t(\i^*_M =j)\P_t(\i^*_M=i^*) \left(\E_t[\X_t(i^*,j)|\i^*_M=i^*] - \E_t[\X_t(i^*,j)]\right) \\ \label{eq:infoExpand}
&\Info_t(\i^*_M; (\I_t,\J_t,\X_t(i,j))) \nonumber \\
&\quad = \sum_{i,j,i^*} \P_t(\i^*_M =i)\P_t(\i^*_M=i^*)\P_t(\i^*_M=j) D(\p_t(\X_t(i,j)|\i^*_M=i^*)|| \p_t(\X_t(i,j)))
\end{align}

To begin proving \eqref{eq:regretExpand}, we show,
\begin{align}
\nonumber \E_t[\X_t(\i^*_M,\J_t)] &= \sum_{i^*,j} \P(\i^*_M=i^*)\P(\i^*_M=j) \E_t[\X_t(i^*,j)|\i^*_M=i,\J_t=j]\\ \label{eq:mixedExpectation}
&= \sum_{i^*,j} \P(\i^*_M=i^*)\P(\i^*_M=j) \E_t[\X_t(i^*,j)|\i^*_M=i]
\end{align}
where the second equality follows because $\J_t$ is independent of $\X_t$, when conditioned on $\Hb_t$. \\\\
Furthermore,
\begin{align}
\nonumber \E_t[\X_t(\i^*_M,\i^*_M)] &= \frac{1}{2} \\ \label{eq:indepExpectation}
&= \sum_{i^*,j} \P(\i^*_M=i^*)\P(\i^*_M=j)\E_t[\X_t(i^*,j)]
\end{align}
where the second equality follows because of the assumption $\E_t[\X_t(i^*,j)] = 1-\E_t[\X_t(j,i^*)]$. Combining \eqref{eq:mixedExpectation} and \eqref{eq:indepExpectation}, gives \eqref{eq:regretExpand}.


Next we prove \eqref{eq:infoExpand}.
\begin{align*}
\Info_t(\i^*_M;&(\I_t,\J_t,\X_t(\I_t,\J_t))) \\
&= \Info_t(\i^*_M;(\I_t,\J_t)) + \Info_t(\i^*_M; \X_t(\I_t,\J_t)| \I_t,\J_t) \\
&= \Info_t(\i^*_M; \X_t(\I_t,\J_t)| \I_t,\J_t) \\
&= \sum_{i,j} \P_t(\i^*_M=i)\P_t(\i^*_M=j) \Info_t(\i^*_M; \X_t(i,j))  \\
&= \sum_{i,j,i^*} \P_t(\i^*_M=i)\P_t(\i^*_M=j) \P_t(\i^*_M=i^*)
D(\p_t(\X_t(i,j)|\i^*_M=i^*)|| \p_t(\X_t(i,j)))
\end{align*}
Here the first equality is the chain rule for mutual information, while the second follows from conditional independence of $\I_t$, $\J_t$, and $\i^*_M$, given $\Hb_t$. The third equality follows because of conditional independence of $(\X_t,\i^*_M)$ and $(\I_t,\J_t)$ given $\Hb_t$. The final equality is a standard identity for mutual information. Thus, \eqref{eq:infoExpand} holds.\\\\\\
Then we bound $\E_t[\rb_t]$ in terms of the mutual information.\\
\begin{align*}
\E_t[\rb_t] & \le \sum_{i^*,j} \P_t(\i^*_M =j)\P_t(\i^*_M=i^*)\sqrt{\frac{1}{2} D(\p_t(\X_t(i^*,j)|\i^*_M=i^*)|| \p_t(\X_t(i^*,j)))} \\
&\le \sqrt{\frac{A^2}{2} \sum_{i^*,j} \P_t(\i^*_M=j)^2\P_t(\i^*_M=i^*)^2 D(\p_t(\X_t(i^*,j)|\i^*_M=i^*)|| \p_t(\X_t(i^*,j)))}\\
& \le  \sqrt{\frac{A^2}{2}	\sum_{i,j,i^*} 	\P_t(\i^*_M =i)^2\P_t(\i^*_M=i^*)^2\P_t(\i^*_M=j) D(\p_t(\X_t(i,j)|\i^*_M=i^*)|| \p_t(\X_t(i,j)))} \\
& \le  \sqrt{\frac{A^2}{2}	\sum_{i,j,i^*} 	\P_t(\i^*_M =i)\P_t(\i^*_M=i^*)\P_t(\i^*_M=j) D(\p_t(\X_t(i,j)|\i^*_M=i^*)|| \p_t(\X_t(i,j)))}\\ 
&= \sqrt{\frac{A^2}{2}\Info_t(\i^*_M;(\I_t,\J_t,\X_t(i,j)))}
\end{align*}
The first inequality is from Pinsker's inequality. The second is from the Cauchy-Schwarz inequality. The third is because adding more non-negative terms cannot decrease the sum. The final inquality is because $\P_t(\i^*_M =i)^2 \le \P_t(\i^*_M=i)$.\\\\
Next we cite the following,
$\sum_{t=1}^T\,\Info_t(\i^*_M;(\I_t,\J_t,\X_t(i,j))) \leq \Ent(\i^*_M)$
(see section 5 of \cite{russo2016information}) and therefore
$\sum_{t=1}^T\,\sqrt{\Info_t(\i^*_M;(\I_t,\J_t,\X_t(i,j)))} \leq
\sqrt{T\,\sum_{t=1}^T\,\Info_t(\i^*_M;(\I_t,\J_t,\X_t(i,j)))} \leq
\sqrt{T\,\Ent(\i^*_M)}$ (Cauchy-Schwartz inequality), 
\begin{align*}
\E[\R_T]\ &\leq \ \sum_{t=1}^T\,\sqrt{\frac{A^2}{2}\Info_t(\i^*_M;(\I_t,\J_t,\X_t(i,j)))} \\
&\leq \ \sqrt{\frac{A^2}{2}\,T\,\Ent(\i^*_M)} \\
&=\  \frac{A}{\sqrt{2}}\sqrt{T\,\Ent(\i^*_M)}
\end{align*}\\
Finally, we have $\Ent(\i^*_M) \le \log A$ since there are $A$ actions, and so the desired bound is achieved.
$\hfill\blacksquare$


\subsection{Proof of Theorem 4.2}
The proof method uses the same concepts from \cite{russo2016information} as the proof of Theorem 4.1. \\\\
First, we make the following definitions: $\E_t$ is the expectation, $\P_t$ is the probability measure, $\p_t$ is the probability density, and $\Info_t(\cdot\ ;\ \cdot)$ is mutual information, all conditioned on the history $\Hb_t$, at time $t$. Furthermore, $D(\cdot\ ||\ \cdot)$ is the Kullback-Leibler divergence and $\Ent$ is entropy.\\\\
Then we note that Thompson Sampling selects both $\I_t$ and $\J_t$ using independent samples from the same posterior distribution conditioned on $\Hb_t$. Therefore, $\I_t$ and $\J_t$ are independent and identically distributed, and the terms $\X_t(\i^*_B,\I_t)$ and $\X_t(\i^*_B,\J_t)$ are identically distributed.\\\\
Let $\rb_t$ be the instantaneous regret at time $t$, such that $\R_T = \sum_{t=1}^T \rb_t$.\\\\
By construction,
\begin{equation}
\label{eq:fsProb} \P_t(\I_t = i) = (1-\alpha) \P_t(\i_B^*=i)+\frac{\alpha}{A}
\end{equation}
Now we bound $\E_t[\rb_t]$ in terms of mutual information.
\begin{align}
&\E_t[\rb_t] = \frac{1}{A}\sum_{i,j} \left(\P_t(\i_B^*=i)\E_t[\X_t(i,j)|\i_B^*=i] - \P_t(\I_t=i)\E_t[\X_t(i,j)]\right) \label{eq:fsExpectation} \\
&= \frac{1}{A}\sum_{i,j} \left(\P_t(\i_B^*=i)\left(\E_t[\X_t(i,j)|\i_B^*=i]-\E_t[\X_t(i,j)]\right) + \alpha \left(\P_t(\i_B^*=i)-\frac{\alpha}{A}\right) \E_t[\X_t(i,j)]\right) \\
&\le \alpha + \frac{1}{A}\sum_{i,j}\P_t(\i_B^*=i)\left(\E_t[\X_t(i,j)|\i_B^*=i]-\E_t[\X_t(i,j)]\right) \label{eq:simpleAlpha} \\
&\le \alpha + \frac{1}{A}\sum_{i,j} \P_t(\i_B^*=i)\sqrt{\frac{1}{2} D(\p_t(\X_t(i,j)|\i_B^*=i)|| \p_t(\X_t(i,j)))} \label{eq:pinskerBorda} \\
&\le \alpha + \frac{1}{A}\sum_j\sqrt{\sum_i	\frac{A}{2} \P_t(\i_B^*=i)^2	D(\p_t(\X_t(i,j)|\i_B^*=i)|| \p_t(\X_t(i,j)))} \label{eq:csBorda} \\ 
&\le \alpha + \sqrt{\sum_{i,j} \frac{1}{2} \P_t(\i_B^*=i)^2	D(\p_t(\X_t(i,j)|\i_B^*=i)|| \p_t(\X_t(i,j))} \label{eq:csConcavity}
\end{align}
\begin{align}
&\le \alpha + \sqrt{\frac{A}{\alpha(1-\alpha)}\frac{1}{2} \sum_{i,j} \P_t(\i_B^*=i)\P_t(\I_t=i)\P_t(\J_t=j) D(\p_t(\X_t(i,j)|\i_B^*=i)|| \p_t(\X_t(i,j))} \label{eq:fsInequality} \\
&\le \alpha + \sqrt{\frac{A}{\alpha(1-\alpha)}\frac{1}{2} \sum_{i^*,i,j} \P_t(\i_B^*=i^*)\P_t(\I_t=i)\P_t(\J_t=j) D(\p_t(\X_t(i,j)|\i_B^*=i^*)|| \p_t(\X_t(i,j))} \label{eq:bordaExtra} \\
&= \alpha + \frac{1}{\sqrt{\alpha(1-\alpha)}} \sqrt{\frac{A}{2}\Info_t(\i_B^*;(\I_t,\J_t,\X_t(\I_t,\J_t)))} \label{eq:Info}\\ 
&\le \alpha +\frac{1}{\sqrt{\alpha}}\sqrt{A\,\Info_t(\i_B^*;(\I_t,\J_t,\X_t(\I_t,\J_t)))} \label{eq:halfIneq}
\end{align}
Here \eqref{eq:fsExpectation} is derived analogously to \eqref{eq:regretExpand}, and the inequality \eqref{eq:simpleAlpha} follows because $\E_t[\X_t(i,j)]\leq 1$ and $\frac{1}{A}\sum_{i,j}\P_t(\i_B^*=i)=1$. Then the inequalities \eqref{eq:pinskerBorda}, \eqref{eq:csBorda}, and \eqref{eq:csConcavity} respectively follow from Pinsker's inequality, the Cauchy-Schwarz inequality, and concavity. The inequality \eqref{eq:fsInequality} follows because,
\begin{align*}
\P_t(\i_B^*=i)&\le \frac{1}{1-\alpha} \P_t(\I_t=i)
\end{align*}
and also from \eqref{eq:fsProb},
\begin{align*}
1 & \le \frac{A}{\alpha} \P_t(\I_t=i),
\end{align*}

The inequality \eqref{eq:bordaExtra} follows because adding extra non-negative terms cannot decrease the sum, and the result $\Info_t(\i_B^*;(\I_t,\J_t,\X_t(\I_t,\J_t)))$ in \eqref{eq:Info} is derived analogously to \eqref{eq:infoExpand}. The inequality \eqref{eq:halfIneq} follows because $\alpha < \frac{1}{2}$ implies that $\frac{1}{\sqrt{1-\alpha}} < \sqrt{2}$.

Next we cite the following,
$\sum_{t=1}^T\,\Info_t(\i^*_B;(\I_t,\J_t,\X_t(i,j))) \leq \Ent(i^*_B)$
(see section 5 of \cite{russo2016information}) and therefore
$\sum_{t=1}^T\,\sqrt{\Info_t(\i^*_B;(\I_t,\J_t,\X_t(i,j)))} \leq
\sqrt{T\,\sum_{t=1}^T\,\Info_t(\i^*_B;(\I_t,\J_t,\X_t(i,j)))} \leq
\sqrt{T\,\Ent(i^*_B)}$, from the Cauchy-Schwartz inequality. Thus, the
regret can be bounded as
\begin{align*}
\E[\R_T]\ &\leq \ \sum_{t=1}^T\,\bigg(\alpha +\frac{1}{\sqrt{\alpha}}\sqrt{A\,\Info_t(\i_B^*;(\I_t,\J_t,\X_t(\I_t,\J_t)))}\,\bigg) \\
&\leq \ \sum_{t=1}^T\,\alpha +\frac{1}{\sqrt{\alpha}}\sqrt{A\,T\,\Ent(\i^*_B)}\\
&= \ \alpha\,T +\frac{1}{\sqrt{\alpha}}\sqrt{A\,T\,\Ent(\i^*_B)}
\end{align*}\\
Finally, we have $\Ent(i^*_B) \le \log A$ since there are $A$ actions, and so the desired bound is achieved when substituting $\alpha = c\,T^{-1/3}$,
\begin{align*}
\E[\R_T]\ &\leq \ c\,T^{-1/3}\,T + \sqrt{\frac{T}{c\,T^{-1/3}}}\sqrt{A\,\log A}\\
&= \ c\,T^{2/3} + T^{2/3}\,\sqrt{\frac{A}{c}\,T\,\log A}
\end{align*}
$\hfill\blacksquare$

\subsection{Proof of Theorem 4.3}

The proof of Theorem 4.3 requires the following auxiliary lemma.

\textbf{Lemma.} \textit{If hyperparameter $\beta\leq1$, then the following holds for all $i,j$ and any $0<\delta<1$,}
\begin{align}
\P\left[\sum_{t=1}^T \X_t(i,\J_t) - \sum_{t=1}^T \widetilde{\X}_{p,t}(i,\J_t) \ \leq \ \frac{\log \delta^{\text{-}1}}{\beta} \right] \ \geq \ 1-\delta \label{lemmaI} \\
\P\left[\sum_{t=1}^T \X_t(j,\I_t) - \sum_{t=1}^T \widetilde{\X}_{q,t}(j,\I_t) \ \leq \ \frac{\log \delta^{\text{-}1}}{\beta} \right] \ \geq \ 1-\delta \label{lemmaJ}
\end{align}\\\\
\textbf{Proof.} The proof method follows those used for lemma 3.1 in \cite{bubeck2012regret}.\\\\
Taking the expected value with respect to $\I_t$, for any $i$ and any $t\leq T$,
\begin{align*}
&\E_{\I_t}\!\left[\exp\left(\beta \X_t(i,\J_t) - \beta\widetilde{\X}_{p,t}(i,\J_t)\right)\right] \\
&= \E_{\I_t}\!\left[\exp\left(\beta \X_t(i,\J_t) - \beta \frac{\X_t(i,\J_t)\,\mathbbm{1}(i=\I_t)}{\p_t(i)}\right)\exp\left(\frac{\text{-}\beta^2}{\p_t(i)}\right)\right] \\
&\stackrel{\text{(a)}}{\leq} \E_{\I_t}\!\bigg[\bigg(1+\left(\beta \X_t(i,\J_t) - \beta \frac{\X_t(i,\J_t)\,\mathbbm{1}(i=\I_t)}{\p_t(i)}\right) \\
&\qquad\qquad +\left(\beta \X_t(i,\J_t) - \beta \frac{\X_t(i,\J_t)\,\mathbbm{1}(i=\I_t)}{\p_t(i)}\right)^2\bigg)\exp\bigg(\frac{\text{-}\beta^2}{\p_t(i)}\bigg)\bigg] \\
&= \E_{\I_t}\!\bigg[1 + \bigg(\beta \X_t(i,\J_t) - \beta \frac{\X_t(i,\J_t)\,\mathbbm{1}(i=\I_t)}{\p_t(i)}\bigg) \\
&\qquad\qquad +\left(\beta \X_t(i,\J_t) - \beta \frac{\X_t(i,\J_t)\,\mathbbm{1}(i=\I_t)}{\p_t(i)}\right)^2\bigg]\ \E_{\I_t}\!\left[\exp\left(\frac{\text{-}\beta^2}{\p_t(i)}\right)\right] \\
&= \bigg(1 + \E_{\I_t}\!\left[\beta \X_t(i,\J_t) - \beta \frac{\X_t(i,\J_t)\,\mathbbm{1}(i=\I_t)}{\p_t(i)}\right] \\
&\qquad\qquad+ \E_{\I_t}\!\left[\left(\beta \X_t(i,\J_t) - \beta \frac{\X_t(i,\J_t)\,\mathbbm{1}(i=\I_t)}{\p_t(i)}\right)^2\right]\bigg)\exp\left(\frac{\text{-}\beta^2}{\p_t(i)}\right) \\
&\stackrel{\text{(b)}}{=} \left(1 + \E_{\I_t}\!\left[\left(\beta \X_t(i,\J_t) - \beta \frac{\X_t(i,\J_t)\,\mathbbm{1}(i=\I_t)}{\p_t(i)}\right)^2\right]\right)\exp\left(\frac{\text{-}\beta^2}{\p_t(i)}\right) \\
&\stackrel{\text{(c)}}{\leq} \left(1 + \beta^2 \frac{\X_t(i,\J_t)^2}{\p_t(i)}\right)\exp\left(\frac{\text{-}\beta^2}{\p_t(i)}\right) \\
&\stackrel{\text{(d)}}{\leq} 1 \\
&\text{and similarly,}\quad \E_{\J_t}\!\left[\exp\left(\beta \X_t(j,\I_t) - \beta\widetilde{\X}_{q,t}(j,\I_t)\right)\right] \ \leq \ 1
\end{align*}
where (a) uses $\exp(x)\leq1+x+x^2$ for $x\leq1$, which is true because $\beta\leq1$, $|\X_t(i,\J_t)|\leq1$, and $|\X_t(j,\I_t)|\leq1$ for all $i,j$ and $t\leq T$,\\\\
(b) uses,
\begin{align*}
&\E_{\I_t}\!\left[\beta \X_t(i,\J_t) - \beta \frac{\X_t(i,\J_t)\,\mathbbm{1}(i=\I_t)}{\p_t(i)}\right] = \E_{\I_t}\!\Big[\beta \X_t(i,\J_t)\Big] - \E_{\I_t}\!\left[\beta \frac{\X_t(i,\J_t)\,\mathbbm{1}(i=\I_t)}{\p_t(i)}\right] \\
&= \beta \X_t(i,\J_t) - \sum_{k=1}^A \p_t(k) \beta\frac{\X_t(i,\J_t)\,\mathbbm{1}(i=k)}{\p_t(i)} = \beta \X_t(i,\J_t) - \beta \X_t(i,\J_t) = 0 \\
&\text{and similarly,}\quad \E_{\J_t}\!\left[\beta \X_t(j,\I_t) - \beta \frac{\X_t(j,\I_t)\,\mathbbm{1}(j=\J_t)}{\q_t(i)}\right] = 0
\end{align*}
(c) uses,
\begin{align*}
&\E_{\I_t}\!\left[\left(\beta \X_t(i,\J_t) - \beta \frac{\X_t(i,\J_t)\,\mathbbm{1}(i=\I_t)}{\p_t(i)}\right)^2\right] \\
&= \E_{\I_t}\!\Big[\beta^2 \X_t(i,\J_t)^2\Big] - \E_{\I_t}\!\left[2\beta^2 \frac{\X_t(i,\J_t)^2\,\mathbbm{1}(i=\I_t)}{\p_t(i)}\right] + \E_{\I_t}\!\left[\beta^2 \frac{\X_t(i,\J_t)^2\,\mathbbm{1}(i=\I_t)}{\p_t(i)^2}\right] \\
&= \beta^2 \X_t(i,\J_t)^2 - \sum_{k=1}^A \p_t(k)\, 2\beta^2 \frac{\X_t(i,\J_t)^2\,\mathbbm{1}(i=k)}{\p_t(i)} + \sum_{k=1}^A \p_t(k)\, \beta^2 \frac{\X_t(i,\J_t)^2\,\mathbbm{1}(i=k)}{\p_t(i)^2}\\
&= \text{-}\beta^2 \X_t(i,\J_t)^2 + \beta^2 \frac{\X_t(i,\J_t)^2}{\p_t(i)} \ \leq \ \beta^2 \frac{\X_t(i,\J_t)^2}{\p_t(i)} \\
&\text{and similarly,}\quad \E_{\J_t}\!\left[\left(\beta \X_t(j,\I_t) - \beta \frac{\X_t(j,\I_t)\,\mathbbm{1}(j=\J_t)}{\q_t(i)}\right)^2\right] \ \leq \ \beta^2 \frac{\X_t(j,\I_t)^2}{\q_t(i)}
\end{align*}
and (d) uses $(1+x)\exp(\text{-}x)\leq1$ for all $x$. \\\\
Then the following holds for any $i,j$, since all $\X_t$ are independent,
\begin{align*}
&\E_{\I_t}\!\left[\exp\left(\beta \sum_{t=1}^T \X_t(i,\J_t) - \beta \sum_{t=1}^T\widetilde{\X}_{p,t}(i,\J_t)\right)\right] \\
&= \prod_{t=1}^T \E_{\I_t}\!\left[\exp\left(\beta \X_t(i,\J_t) - \beta \widetilde{\X}_{p,t}(i,\J_t)\right)\right] \\
&\leq \ 1^T = 1 \\
&\text{and similarly,}\quad \E_{\J_t}\!\left[\exp\left(\beta \sum_{t=1}^T \X_t(j,\I_t) - \beta \sum_{t=1}^T\widetilde{\X}_{q,t}(j,\I_t)\right)\right] \ \leq \ 1
\end{align*}\\
Finally, since Markov's inequality implies $\P\big[\log\exp(\Y) \leq \log\,\delta^{\text{-}1}\big] \geq 1 - \delta\,\E\big[\exp(\Y)\big]$, by then setting, 
\begin{align*}
\Y = \beta \sum_{t=1}^T \X_t(i,\J_t) - \beta \sum_{t=1}^T\widetilde{\X}_{p,t}(i,\J_t)
\end{align*}
we have that $\E\big[\exp(\Y)\big] \leq 1$, and therefore we achieve the desired results,
\begin{align*}
&\P\left[\beta \sum_{t=1}^T \X_t(i,\J_t) - \beta \sum_{t=1}^T\widetilde{\X}_{p,t}(i,\J_t) \leq \log\,\delta^{\text{-}1}\right] \geq 1 - \delta \\
&\text{and similarly,}\quad \P\left[\beta \sum_{t=1}^T \X_t(j,\I_t) - \beta \sum_{t=1}^T\widetilde{\X}_{q,t}(j,\I_t) \leq \log\,\delta^{\text{-}1}\right] \geq 1 - \delta
\end{align*}
$\hfill\blacksquare$\\\\

Now we turn to the proof of Theorem 4.3.
The proof method follows those used for Theorems 3.2 and 3.3 in \cite{bubeck2012regret}. \\\\
Recall that the regret has the form 
\begin{align}
\R_T = \frac{1}{2}\sum_{t=1}^T \Big(\X_t(i^*_M,\J_t) - \X_t(\I_t,\J_t) + \X_t(i^*_M,\I_t) - \X_t(\I_t,\J_t)\Big) \label{regret}
\end{align}

Taking the expected value with respect to $i\sim \p_t$ and $j\sim \q_t$, for any $t\leq T$,
\begin{align*}
&\E_{i\sim \p_t}\!\big[\widetilde{\X}_{p,t}(i,\J_t)\big] = \sum_{k=1}^A \,\p_t(k)\ \frac{\X_t(\I_t,\J_t)\,\mathbbm{1}(k=\I_t) \ + \ \beta}{\p_t(k)} \\
&= \X_t(\I_t,\J_t) + A\,\beta \\
&\text{and similarly,}\quad \E_{j\sim \q_t}\!\big[\widetilde{\X}_{q,t}(j,\I_t)\big] = \X_t(\J_t,\I_t) + A\,\beta
\end{align*} 
This means we have, for any $i,j$,
\begin{align}
&\sum_{t=1}^T \X_t(i,\J_t) - \sum_{t=1}^T \X_t(\I_t,\J_t) = A\,T\,\beta + \sum_{t=1}^T \X_t(i,\J_t) - \sum_{t=1}^T \E_{i\sim \p_t}\!\big[\widetilde{\X}_{p,t}(i,\J_t)\big] \label{mainI} \\
&\sum_{t=1}^T \X_t(j,\I_t) - \sum_{t=1}^T \X_t(\J_t,\I_t) = A\,T\,\beta + \sum_{t=1}^T \X_t(j,\I_t) - \sum_{t=1}^T \E_{j\sim \q_t}\!\big[\widetilde{\X}_{q,t}(j,\I_t)\big] \label{mainJ}
\end{align}\\
Now we will begin bounding the expectation terms, which are taken with respect to $i,j$ being distributed as $\p_t,\q_t$ respectively. But by the definitions of those distributions, we can split them up into the uniform portion $\ub$ and the softmax portions $\s_{p,t},\s_{q,t}$, such that $\p_t = (1-\gamma)\,\s_{p,t} + \gamma\,u$ and $\q_t = (1-\gamma)\,\s_{q,t} + \gamma\,u$. Therefore,
\begin{align}
-\E_{i\sim \p_t}\!\big[&\widetilde{\X}_{p,t}(i,\J_t)\big] =\ - (1-\gamma)\ \E_{i\sim \s_{p,t}}\big[\widetilde{\X}_{p,t}(i,\J_t)\big] \ - \ \gamma\,\E_{i\sim \ub}\big[\widetilde{\X}_{p,t}(i,\J_t)\big]\nonumber \\
&=\ (1-\gamma)\ \frac{1}{\eta}\log\exp\left(\text{-}\eta\,\E_{k\sim \s_{p,t}}\big[\widetilde{\X}_{p,t}(k,\J_t)\big]\right) \ - \ \gamma\,\E_{i\sim \ub}\big[\widetilde{\X}_{p,t}(i,\J_t)\big]\nonumber \\
&=\ (1-\gamma)\ \bigg(\frac{1}{\eta}\log\E_{i\sim \s_{p,t}}\Big[\exp\big(\eta\,\widetilde{\X}_{p,t}(i,\J_t)\big)\Big] \ +\ \frac{1}{\eta}\log\exp\left(\text{-}\eta\,\E_{k\sim \s_{p,t}}\big[\widetilde{\X}_{p,t}(k,\J_t)\big]\right)\nonumber \\
&\qquad-\ \frac{1}{\eta}\log\E_{i\sim \s_{p,t}}\Big[\exp\big(\eta\,\widetilde{\X}_{p,t}(i,\J_t)\big)\Big]\bigg) \ - \ \gamma\,\E_{i\sim \ub}\big[\widetilde{\X}_{p,t}(i,\J_t)\big]\nonumber \\
&=\ (1-\gamma)\ \bigg(\frac{1}{\eta}\log\E_{i\sim \s_{p,t}}\Big[\exp\Big(\eta\,\widetilde{\X}_{p,t}(i,\J_t) - \eta\,\E_{k\sim \s_{p,t}}\big[\widetilde{\X}_{p,t}(k,\J_t)\big]\Big)\Big]\nonumber \\
&\qquad -\ \frac{1}{\eta}\log\E_{i\sim \s_{p,t}}\Big[\exp\big(\eta\,\widetilde{\X}_{p,t}(i,\J_t)\big)\Big]\bigg) \ - \ \gamma\,\E_{i\sim \ub}\big[\widetilde{\X}_{p,t}(i,\J_t)\big] \label{expEqI} \\
&\text{and similarly,} \nonumber\\
-\E_{j\sim \q_t}\!\big[&\widetilde{\X}_{q,t}(j,\I_t)\big] =\ (1-\gamma)\ \bigg(\frac{1}{\eta}\log\E_{j\sim \s_{q,t}}\Big[\exp\Big(\eta\,\widetilde{\X}_{q,t}(j,\I_t) - \eta\,\E_{k\sim \s_{q,t}}\big[\widetilde{\X}_{q,t}(k,\I_t)\big]\Big)\Big]\nonumber \\
&\qquad\qquad\qquad  -\ \frac{1}{\eta}\log\E_{j\sim \s_{q,t}}\Big[\exp\big(\eta\,\widetilde{\X}_{q,t}(j,\I_t)\big)\Big]\bigg) \ - \ \gamma\,\E_{j\sim \ub}\big[\widetilde{\X}_{q,t}(j,\I_t)\big] \label{expEqJ}
\end{align}
Next we focus on the main softmax expectation terms in eqs. \ref{expEqI} and \ref{expEqJ},
\begin{align*}
\log\,\E_{i\sim \s_{p,t}}&\Big[\exp\Big(\eta\,\widetilde{\X}_{p,t}(i,\J_t) - \eta\,\E_{k\sim \s_{p,t}}\big[\widetilde{\X}_{p,t}(k,\J_t)\big]\Big)\Big] \\
&= \log\E_{i\sim \s_{p,t}}\Big[\exp\Big(\eta\,\widetilde{\X}_{p,t}(i,\J_t)\Big)\Big] - \E_{k\sim \s_{p,t}}\big[\eta\,\widetilde{\X}_{p,t}(k,\J_t)\big] \\
&\stackrel{\text{(a)}}{\leq} \E_{i\sim \s_{p,t}}\Big[\exp\Big(\eta\,\widetilde{\X}_{p,t}(i,\J_t)\Big)\Big] - 1 - \E_{k\sim \s_{p,t}}\big[\eta\,\widetilde{\X}_{p,t}(k,\J_t)\big] \\
&= \E_{i\sim \s_{p,t}}\Big[\exp\Big(\eta\,\widetilde{\X}_{p,t}(i,\J_t)\Big) - 1 - \eta\,\widetilde{\X}_{p,t}(i,\J_t)\Big] \\
&\stackrel{\text{(b)}}{\leq} \E_{i\sim \s_{p,t}}\Big[\Big(\eta\,\widetilde{\X}_{p,t}(i,\J_t)\Big)^2\,\Big] \\
&= \eta^2 \sum_{k=1}^A \s_{p,t}(k)\ \frac{\X_t(\I_t,\J_t)\,\mathbbm{1}(k=\I_t) + \beta}{\p_t(k)} \ \widetilde{\X}_{p,t}(k,\J_t) \\
&\stackrel{\text{(c)}}{\leq}\ \frac{1 + \beta}{1 - \gamma}\,\eta^2\sum_{k=1}^A\widetilde{\X}_{p,t}(k,\J_t) \\
\text{and similarly,}&\\
\log\,\E_{j\sim \s_{q,t}}&\Big[\exp\Big(\eta\,\widetilde{\X}_{q,t}(j,\I_t) - \eta\,\E_{k\sim \s_{q,t}}\big[\widetilde{\X}_{q,t}(k,\I_t)\big]\Big)\Big] \ \leq \ \frac{1 + \beta}{1 - \gamma}\,\eta^2\sum_{k=1}^A\widetilde{\X}_{q,t}(k,\I_t)
\end{align*}
where (a) uses $\log x\leq x-1$, (b) uses $\exp x\leq 1+x+x^2$, and (c) uses $\X_t(\I_t,\J_t)\,\mathbbm{1}(k=\I_t) \leq b$, $\X_t(\J_t,\I_t)\,\mathbbm{1}(k=\I_t) \leq b$, $\s_{p,t}(k)/\p_t(k) \ \leq \ 1/(1-\gamma)$, and $\s_{q,t}(k)/\q_t(k) \ \leq \ 1/(1-\gamma)$ for all $k$.\\\\
Note that (a) and (b) require $x\leq1$, meaning that we need $\eta\,\widetilde{\X}_{p,t}(i,\J_t) \leq 1$ and $\eta\,\widetilde{\X}_{q,t}(j,\I_t) \leq 1$ for all $i,j$ and $t\leq T$. \\\\
From their definitions,
\begin{align*}
\widetilde{\X}_{p,t}(i,\J_t) = &\frac{\X_t(\I_t,\J_t)\,\mathbbm{1}(i=\I_t) \ + \ \beta}{\p_t(i)}\qquad \forall i\in\{1,\dots,A\} \\
&\leq \frac{1 + \beta}{\gamma/A} \ =\ \frac{(1+\beta)\,A}{\gamma} \\
\text{and similarly,}&\quad \widetilde{\X}_{q,t}(j,\I_t) \leq \frac{(1+\beta)\,A}{\gamma}\qquad \forall j\in\{1,\dots,A\}
\end{align*}
and so this requirement is exactly met by the assumption $0\leq(1+\beta)A\,\eta\leq\gamma\leq1/2$. \\\\\\
Then we look at the uniform expectation terms in eqs. \ref{expEqI} and \ref{expEqJ},
\begin{align*}
\E_{i\sim \ub}\big[\widetilde{\X}_{p,t}(i,\J_t)\big] &=\ \sum_{k=1}^A\frac{1}{A}\frac{\X_t(\I_t,\J_t)\,\mathbbm{1}(k=\I_t) + \beta}{\p_t(k)} \\
&\geq\ \sum_{k=1}^A\frac{1}{A}\,(0+\beta)\,\frac{A}{\gamma} \ =\ \frac{\beta\,A}{\gamma} \\
&\geq\ 0 \\
\text{and similarly,}\quad& \E_{j\sim \ub}\big[\widetilde{\X}_{q,t}(j,\I_t)\big]\ \geq\ 0
\end{align*}
Making these substitutions into eqs. \ref{expEqI} and \ref{expEqJ}, and summing over time,
\begin{align*}
-\sum_{t=1}^T &\ \E_{i\sim \p_t}\!\big[\widetilde{\X}_{p,t}(i,\J_t)\big] \\
&\leq\ (1-\gamma)\sum_{t=1}^T\ \bigg(\frac{1}{\eta}\ \frac{1 + \beta}{1 - \gamma}\,\eta^2\sum_{k=1}^A\widetilde{\X}_{p,t}(k,\J_t) \\
&\qquad\qquad -\ \frac{1}{\eta}\log\E_{i\sim \s_{p,t}}\Big[\exp\big(\eta\,\widetilde{\X}_{p,t}(i,\J_t)\big)\Big]\bigg) - \gamma\,\sum_{t=1}^T 0 \\
&=\ (1 + \beta)\,\eta\sum_{t=1}^T\sum_{k=1}^A\widetilde{\X}_{p,t}(k,\J_t) -\ \frac{1-\gamma}{\eta}\sum_{t=1}^T\,\log\bigg(\sum_{k=1}^A\s_{p,t}(k)\exp\big(\eta\,\widetilde{\X}_{p,t}(k,\J_t)\big)\bigg) \\
&\stackrel{\text{(a)}}{=}\ (1 + \beta)\,\eta\sum_{t=1}^T\sum_{k=1}^A\widetilde{\X}_{p,t}(k,\J_t) -\ \frac{1-\gamma}{\eta}\sum_{t=1}^T\,\log\bigg(\frac{\sum_{k=1}^A\exp\big(\eta\,\sum_{\tau=1}^t\widetilde{\X}_{p,\tau}(k,\J_t)\big)}{\sum_{k=1}^A\exp\big(\eta\,\sum_{\tau=1}^{t-1}\widetilde{\X}_{p,\tau}(k,\J_t)\big)}\bigg)  \\
&=\ (1 + \beta)\,\eta\sum_{t=1}^T\sum_{k=1}^A\widetilde{\X}_{p,t}(k,\J_t) -\ \frac{1-\gamma}{\eta}\,\log\bigg(\prod_{t=1}^T\frac{\sum_{k=1}^A\exp\big(\eta\,\sum_{\tau=1}^t\widetilde{\X}_{p,\tau}(k,\J_t)\big)}{\sum_{k=1}^A\exp\big(\eta\,\sum_{\tau=1}^{t-1}\widetilde{\X}_{p,\tau}(k,\J_t)\big)}\bigg)  \\
&\stackrel{\text{(b)}}{=}\ (1 + \beta)\,\eta\sum_{t=1}^T\sum_{k=1}^A\widetilde{\X}_{p,t}(k,\J_t) -\ \frac{1-\gamma}{\eta}\,\log\bigg(\sum_{k=1}^A\exp\big(\eta\,\sum_{t=1}^T\widetilde{\X}_{p,t}(k,\J_t)\big)\bigg)  \\
&\leq\ (1 + \beta)A\,\eta\,\max_k\sum_{t=1}^T\widetilde{\X}_{p,t}(k,\J_t) -\ \frac{1-\gamma}{\eta}\,\log\bigg(A\,\max_k\,\exp\big(\eta\,\sum_{t=1}^T\widetilde{\X}_{p,t}(k,\J_t)\big)\bigg) \\
&\stackrel{\text{(c)}}{\leq}\ (1 + \beta)A\,\eta\,\max_k\sum_{t=1}^T\widetilde{\X}_{p,t}(k,\J_t) -\ (1-\gamma)\,\max_k\,\sum_{t=1}^T\widetilde{\X}_{p,t}(k,\J_t) + \frac{\log(A)}{\eta} \\
&= \ -\big(1-\gamma - (1 + \beta)A\,\eta\big)\,\max_k\sum_{t=1}^T\widetilde{\X}_{p,t}(k,\J_t) + \frac{\log(A)}{\eta}  \\
&\stackrel{\text{(d)}}{\leq}\ -\big(1-\gamma - (1 + \beta)A\,\eta\big)\,\max_k\sum_{t=1}^T \X_t(k,\J_t) + \frac{\log(\delta^{\text{-}1})}{\beta} + \frac{\log(A)}{\eta} \\
&\text{and similarly,} \\
-\sum_{t=1}^T &\ \E_{j\sim \q_t}\!\big[\widetilde{\X}_{q,t}(j,\I_t)\big] \ \leq \ -\big(1-\gamma - (1 + \beta)A\,\eta\big)\,\max_k\sum_{t=1}^T \X_t(k,\I_t) + \frac{\log(\delta^{\text{-}1})}{\beta} + \frac{\log(A)}{\eta}
\end{align*}
where (a) uses the definitions of $\s_{p,t}(k)$ and $\s_{q,t}(k)$ as,
\begin{align*}
\s_{p,t}(k) = \frac{\exp\big(\eta\,\sum_{\tau=1}^{t-1}\widetilde{\X}_{p,\tau}(k,\J_t)\big)}{\sum_{k=1}^A\exp\big(\eta\,\sum_{\tau=1}^{t-1}\widetilde{\X}_{p,\tau}(k,\J_t)\big)} \\
\s_{q,t}(k) = \frac{\exp\big(\eta\,\sum_{\tau=1}^{t-1}\widetilde{\X}_{q,\tau}(k,\I_t)\big)}{\sum_{k=1}^A\exp\big(\eta\,\sum_{\tau=1}^{t-1}\widetilde{\X}_{q,\tau}(k,\I_t)\big)}
\end{align*}
(b) uses the cancellation of numerators and denominators in successive terms of the product, and that $\widetilde{\X}_{p,0}(k,\J_t)=\widetilde{\X}_{q,0}(k,\I_t)=0$ for all $k$, (c) uses that $-(1-\gamma)\,\log(A)/\eta\ \leq\ \log(A)/\eta$, and (d) uses that $\big(1-\gamma - (1 + \beta)A\,\eta\big) \ \leq \ 1$, which comes from the assumption $0\leq(1+\beta)A\,\eta\leq\gamma\leq1/2$, together with the lemma eqs. \ref{lemmaI} and \ref{lemmaJ}. Note that the inclusion of $\delta$ from the lemma equations implies that these results hold with probability $1-\delta$ for any $0<\delta<1$.\\\\
Then substituting these into eqs. \ref{mainI} and \ref{mainJ},
\begin{align*}
&\sum_{t=1}^T \X_t(i,\J_t) - \sum_{t=1}^T \X_t(\I_t,\J_t) \\
&\leq A\,T\,\beta + \sum_{t=1}^T \X_t(i,\J_t) -\big(1-\gamma - (1 + \beta)A\,\eta\big)\,\max_k\sum_{t=1}^T \X_t(k,\J_t) + \frac{\log(\delta^{\text{-}1})}{\beta} + \frac{\log(A)}{\eta} \\
&\leq A\,T\,\beta + \gamma\,T + (1 + \beta)A\,\eta\,T + \frac{\log(\delta^{\text{-}1})}{\beta} + \frac{\log(A)}{\eta} \\
&\stackrel{\text{(a)}}{\leq} A\,T\,\beta + 2\gamma\,T + \frac{\log(\delta^{\text{-}1})}{\beta} + \frac{\log(A)}{\eta} \\
&\text{and similarly,}\quad \sum_{t=1}^T \X_t(j,\I_t) - \sum_{t=1}^T \X_t(\J_t,\I_t) \leq A\,T\,\beta + 2\gamma\,T + \frac{\log(\delta^{\text{-}1})}{\beta} + \frac{\log(A)}{\eta}
\end{align*}
with probability $1-\delta$ for any $0<\delta<1$, where (a) uses the assumption $0\leq(1+\beta)A\,\eta\leq\gamma\leq1/2$. \\\\\\
Since these results are valid for any $i,j$, we can use them directly
in eq. \ref{regret},
\begin{align*}
\R_T\ \leq\ A\,T\,\beta + 2\gamma\,T + \frac{\log(\delta^{\text{-}1})}{\beta} + \frac{\log(A)}{\eta}
\end{align*}
and applying the defined hyperparameter values,
\begin{align*}
\R_T\ &\leq\ A\,T\,\sqrt{\frac{\log A}{AT}} + 2\,\left(1.05\,\sqrt{\frac{A\,\log A}{T}}\right)\,T + \frac{\log(\delta^{\text{-}1})\,\sqrt{AT}}{\sqrt{\log A}} + \frac{\log(A)\,\sqrt{AT}}{0.95\,\sqrt{\log A}} \\
&= \sqrt{A\log A}\,\sqrt{T} + 2.1\,\sqrt{A\log A}\,\sqrt{T} + \log \delta^{\text{-}1}\sqrt{A(\log A)^{\text{-}1}}\,\sqrt{T} + \frac{1}{0.95}\,\sqrt{A\log A}\,\sqrt{T} \\
&\leq \bigg(\log\delta^{\text{-}1}\sqrt{A\,(\log A)^{\text{-}1}} + 4.2\sqrt{A\log A} \bigg)\sqrt{T}
\end{align*}
with probability $1-\delta$ for any $0<\delta<1$. \\\\\\
Now we will verify the requirements on $T$ for enforcing the assumption $0\leq(1+\beta)A\,\eta\leq\gamma\leq1/2$.\\\\
Since all of the hyperparameter values are non-negative, then the left-hand side inequality is trivially satisfied. \\
\begin{align*}
\gamma = 1.05\,\sqrt{\frac{A\,\log A}{T}}\ &\leq\ \frac{1}{2} \\
2.1\,\sqrt{A\,\log A}\ &\leq\ \sqrt{T} \\
T\ &\geq\ 4.41\,A\,\log A\\
\quad \\
(1+\beta)A\,\eta = \left(1 + \sqrt{\frac{\log A}{AT}}\,\right)A\,0.95\,\sqrt{\frac{\log A}{AT}} \ &\leq\ \gamma = 1.05\,\sqrt{\frac{A\,\log A}{T}} \\
\frac{0.95\,\log A}{\sqrt{T}} \ &\leq\ 1.05\,\sqrt{A\,\log A} - 0.95\,\sqrt{A\,\log A} \\
T\ &\geq\ \frac{0.95^2\,\log A}{0.1^2\,A}
\end{align*}\\\\
And so the requirement is $T\ \geq\ \max\big[4.41\,A\log A\ ,\ (0.95^2\log A)/(0.1^2\,A)\big]$, as desired.
\newpage
Finally, we demonstrate the following fact for random variable $\W$ with cumulative distribution function $F_\W$,
\begin{align*}
\E[\W] &= \int_0^\infty 1 - F_\W(x)\,dx \ -\  \int_{-\infty}^0 F_\W(x)\,dx \\
&\leq \int_0^\infty 1 - F_\W(x)\,dx \\
&= \int_0^\infty \P(\W > x)\,dx \\
&\text{change variable:}\quad x = \log\delta^{\text{-}1} \\
&\qquad\qquad\qquad\quad\ \   dx = -\,\delta^{\text{-}1}\,d\delta \\
&\qquad\qquad\qquad\quad\ \ x=\infty \rightarrow \delta=0 \\
&\qquad\qquad\qquad\quad\ \ x=0 \ \rightarrow \delta=1 \\
&= \int_0^1 \delta^{\text{-}1}\,\P(\W > \log \delta^{\text{-}1})\, d\delta
\end{align*}\\
Then recalling the regret high probability upper bound, for the required $T$ and any $0<\delta<1$,
\begin{align*}
\P\left[ \R_T \, \leq \, \bigg(\log\delta^{\text{-}1}\sqrt{A\,(\log A)^{\text{-}1}} + 4.2\sqrt{A\log A} \bigg)\sqrt{T}\ \right] \ &\geq \ 1-\delta \\
\P\left[ \frac{\R_T - 4.2\sqrt{A\log A}\,\sqrt{T}}{\sqrt{A\,(\log A)^{\text{-}1}}\,\sqrt{T}} \, \leq \, \log\delta^{\text{-}1}\ \right] \ &\geq \ 1-\delta \\
\P\left[ \frac{\R_T - 4.2\sqrt{A\log A}\,\sqrt{T}}{\sqrt{A\,(\log A)^{\text{-}1}}\,\sqrt{T}} \, > \, \log\delta^{\text{-}1}\ \right] \ &\geq \ 1 - (1-\delta) = \delta
\end{align*}\\\\
Now selecting $\ \W = \big(\R_T - 4.2\sqrt{A\log A}\,\sqrt{T}\big)\ /\ \big(\sqrt{A\,(\log A)^{\text{-}1}}\,\sqrt{T}\big)\ $ we have,
\begin{align*}
\E[\W] &\leq \int_0^1 \delta^{\text{-}1}\,\P(\W > \log \delta^{\text{-}1})\, d\delta \\
&= \int_0^1 \delta^{\text{-}1}\,\delta\, d\delta \\
&= 1
\end{align*}\\

Therefore, we achieve the desired result:
\begin{align*}
\E\left[ \frac{\R_T - 4.2\sqrt{A\log A}\,\sqrt{T}}{\sqrt{A\,(\log A)^{\text{-}1}}\,\sqrt{T}} \right] \ &\leq\ 1 \\
\E[\R_T] \, &\leq \,  \bigg(\sqrt{A\,(\log A)^{\text{-}1}} + 4.2\sqrt{A\log A} \bigg)\sqrt{T}
\end{align*}
$\hfill\blacksquare$

\subsection{Proof of Theorem 4.4}
The proof method follows those used for theorem 6.5 in \cite{cesa2006prediction}. \\\\
First, we recall our definition of the (estimated) Borda score for $\X_t$ as,
\begin{align*}
\gb_t(i) = \frac{1}{A}\sum_{k=1}^A\, \X_t(i,k) \qquad\qquad \widetilde{\gb}_t(i) = \frac{1}{A}\frac{\X_t(i,\J_t)\,\mathbbm{1}(i=\I_t)}{\p_t(\I_t)\,\p_t(\J_t)}
\end{align*}
and we define the sum of (estimated) Borda scores for action $i$ over $t\leq T$ as,
\begin{align*}
\G_T(i) = \sum_{t=1}^T\, \gb_t(i) \qquad\qquad \widetilde{\G}_T(i) = \sum_{t=1}^T\, \widetilde{\gb}_t(i)
\end{align*}
which means we can redefine eq. \ref{Reg} as,
\begin{align*}
\R_T = \frac{1}{2}\sum_{t=1}^T \Big(\gb_t(i^*_B) - \gb_t(\I_t) + \gb_t(i^*_B) - \gb_t(\J_t)\Big)
\end{align*}\\\\
Since $\I_t$ and $\J_t$ are independently drawn from the same probability distribution $\p_t$ at each time $t$, we can equivalently prove eq. \ref{expReg} using an expected regret equation strictly in terms of the Borda scores for $\I_t$ and $i^*_B$,
\begin{align*}
\E[\R_T] &= \ \sum_{t=1}^T\, \E[\gb_t(i^*_B) \ - \  \gb_t(\I_t)]
\end{align*}\\
Next we define a lower bound for the log of the ratio of weight sums at times $T$ and $0$, for any $j$,
\begin{align*}
\log\frac{\W_T}{\W_0} &= \log \W_T - \log \W_0 = \log\sum_{i=1}^A \exp\big(\eta\, \widetilde{\G}_T(i)\big) - \log A \\
&\geq\ \eta\, \widetilde{\G}_T(j) - \log A
\end{align*}
and an upper bound for the log of the ratio of weight sums at times $t$ and $t-1$,
\begin{align*}
\log\frac{\W_t}{\W_{t-1}} &= \log\sum_{i=1}^A \frac{\wb_{t-1}(i)}{\W_{t-1}}\exp\big(\eta\, \widetilde{\gb}_t(i)\big) \\
&\stackrel{\text{(a)}}{=} \log\sum_{i=1}^A \frac{\p_t(i)-\gamma/A}{1-\gamma}\exp\big(\eta\, \widetilde{\gb}_t(i)\big) \\
&\stackrel{\text{(b)}}{\leq} \log\sum_{i=1}^A \frac{\p_t(i)-\gamma/A}{1-\gamma}\big(1 + \eta\, \widetilde{\gb}_t(i) + (e-2)\,\eta^2\,\widetilde{\gb}_t(i)^2 \big) \\
&\stackrel{\text{(c)}}{\leq} \log\bigg(1 + \frac{\eta}{1-\gamma}\sum_{i=1}^A\,\widetilde{\gb}_t(i)\left(\p_t(i)-\frac{\gamma}{A}\right) + \frac{(e-2)\,\eta^2}{1-\gamma}\sum_{i=1}^A\,\widetilde{\gb}_t(i)^2\,\p_t(i)\bigg) \\
&\stackrel{\text{(d)}}{\leq} \frac{\eta}{1-\gamma}\sum_{i=1}^A\,\widetilde{\gb}_t(i)\left(\p_t(i)-\frac{\gamma}{A}\right) + \frac{(e-2)\,\eta^2}{1-\gamma}\sum_{i=1}^A\,\widetilde{\gb}_t(i)^2\,\p_t(i)
\end{align*}
where (a) is from the definition of $\p_t(i)$, (b) is because $e^x \leq 1+x+(e-2)x^2$ for $x\leq1$, and (d) is because $\log(1+x)\leq x$.\\\\For (c), first note that $\sum_{i=1}^A(\p_t(i)-\gamma/A)/(1-\gamma) = 1$, as sum of softmax components. So it would be equivalent, except that on the far right side it has only a $\p_t(i)$ term, and hence $\p_t(i)-\gamma/A\leq \p_t(i)$. This gives the inequality, since $\log(1+a)\leq\log(1+b)$ if $a\leq b$.\\\\
The (b) requirement $\eta\, \widetilde{\gb}_t(i) \leq 1$ holds if we have that $\eta\,A/\gamma^2\leq1$, because $\eta>0$ and $\widetilde{\gb}_t(i) \leq A/\gamma^2$, with $0<\gamma<1$. We confirm this at the end of the proof.\\\\
Now we sum the upper bound over $t\leq T$, to get the log of the ratio of weight sums at times $T$ and $0$,
\begin{align*}
\sum_{t=1}^T\ \log\frac{\W_t}{\W_{t-1}} &= \log\frac{\W_1}{\W_0}\frac{\W_2}{\W_1}\dots\frac{\W_{T-1}}{\W_{T-2}}\frac{\W_T}{\W_{T-1}} = \log\frac{\W_T}{\W_0} \\
&\leq \sum_{t=1}^T\bigg(\frac{\eta}{1-\gamma}\sum_{i=1}^A\,\widetilde{\gb}_t(i)\left(\p_t(i)-\frac{\gamma}{A}\right) + \frac{(e-2)\,\eta^2}{1-\gamma}\sum_{i=1}^A\,\widetilde{\gb}_t(i)^2\,\p_t(i)\bigg) \\
&= \frac{\eta}{1-\gamma}\sum_{t=1}^T\sum_{i=1}^A\,\widetilde{\gb}_t(i)\left(\p_t(i) -\frac{\gamma}{A}\right) + \frac{(e-2)\,\eta^2}{1-\gamma}\sum_{t=1}^T\sum_{i=1}^A\,\widetilde{\gb}_t(i)^2\,\p_t(i)
\end{align*} \\
Then we can compare the lower and upper bounds, to get a single inequality.
\begin{align*}
\eta\,\widetilde{\G}_T(j) - \log A\, \leq\, \frac{\eta}{1-\gamma}\sum_{t=1}^T\sum_{i=1}^A\,\widetilde{\gb}_t(i)\left(\p_t(i) -\frac{\gamma}{A}\right) + \frac{(e-2)\,\eta^2}{1-\gamma}\sum_{t=1}^T\sum_{i=1}^A\,\widetilde{\gb}_t(i)^2\,\p_t(i)
\end{align*} \\
Multiplying both sides by $(1-\gamma)/\eta$ gives,
\begin{align*}
(1-\gamma)\,\widetilde{\G}_T(j) - (1-\gamma)\,\frac{\log A}{\eta}\, \leq\, \sum_{t=1}^T\sum_{i=1}^A\,\widetilde{\gb}_t(i)\left(\p_t(i) -\frac{\gamma}{A}\right) + \eta\,(e-2)\sum_{t=1}^T\sum_{i=1}^A\,\widetilde{\gb}_t(i)^2\,\p_t(i)
\end{align*}
and by rearranging terms and noting that $(1-\gamma)<1$,
\begin{align*}
(1-\gamma)\,\widetilde{\G}_T(j) - \sum_{t=1}^T\sum_{i=1}^A\,\widetilde{\gb}_t(i)\,\p_t(i) \, &\leq\, \frac{\log A}{\eta} - \frac{\gamma}{A}\sum_{i=1}^A\,\widetilde{\G}_T(i) + \eta\,(e-2)\sum_{t=1}^T\sum_{i=1}^A\,\widetilde{\gb}_t(i)^2\,\p_t(i)
\end{align*}
By definition of $\widetilde{\gb}_t(i)$ we then have,
\begin{align*}
(1-\gamma)\,\widetilde{\G}_T(j) - \sum_{t=1}^T\,\widetilde{\gb}_t(\I_t)\p_t(\I_t) \, &\leq\,  \frac{\log A}{\eta} - \frac{\gamma}{A}\sum_{i=1}^A\,\widetilde{\G}_T(i) + \eta\,(e-2)\sum_{t=1}^T\,\widetilde{\gb}_t(\I_t)^2\,\p_t(\I_t)
\end{align*}
and by definition $\p_t(\I_t)<1$ for all $t\leq T$,
\begin{align*}
(1-\gamma)\,\widetilde{\G}_T(j) - \sum_{t=1}^T\,\widetilde{\gb}_t(\I_t) \, &\leq\,  \frac{\log A}{\eta} - \frac{\gamma}{A}\sum_{i=1}^A\,\widetilde{\G}_T(i) + \eta\,(e-2)\sum_{t=1}^T\,\widetilde{\gb}_t(\I_t)^2
\end{align*}
Since all terms are using the unbiased estimates of the Borda scores, we can take the expected value on both sides and replace the estimates with the actual scores,
\begin{align*}
(1-\gamma)\,\G_T(j) - \sum_{t=1}^T\,\gb_t(\I_t) \, &\leq\,  \frac{\log A}{\eta} - \frac{\gamma}{A}\sum_{i=1}^A\,\G_T(i) + \eta\,(e-2)\sum_{t=1}^T\,\E_{\I_t,\J_t}\![\widetilde{\gb}_t(\I_t)^2]
\end{align*}
Noting that $\G_T(i)\leq T$ for any $i$,
\begin{align*}
\G_T(j) - \gamma T - \sum_{t=1}^T\,\gb_t(\I_t) \, &\leq\,  \frac{\log A}{\eta} - \frac{\gamma}{A}\,AT + \eta\,(e-2)\sum_{t=1}^T\,\E_{\I_t,\J_t}\![\widetilde{\gb}_t(\I_t)^2]
\end{align*}
Next we bound the remaining expectation term,
\begin{align*}
\E_{\I_t,\J_t}\![\widetilde{\gb}_t(\I_t)^2] &= \E_{\I_t,\J_t}\!\left[\bigg(\frac{1}{A}\frac{\X_t(\I_t,\J_t)\,\mathbbm{1}(\I_t=\I_t)}{\p_t(\I_t)\,\p_t(\J_t)}\bigg)^2\right] \\
&= \sum_{a=1}^A \sum_{k=1}^A\, \p_t(a)\, \p_t(k)\ \frac{1}{A^2}\,\frac{\X_t(\I_t,\J_t)^2}{\p_t(a)^2\, \p_t(k)^2} \\
&= \frac{1}{A^2}\ \sum_{a=1}^A \sum_{k=1}^A\, \frac{\X_t(\I_t,\J_t)^2}{\p_t(a)\, \p_t(k)} \\
&\leq \frac{1}{A^2}\ \sum_{a=1}^A \sum_{k=1}^A\,\frac{1}{(\gamma/A)^2} \\
&= \frac{A^2}{\gamma^2}
\end{align*}
and because the $\G_T(j)$ term from the original lower bound is valid for any $j$, we can arbitrarily choose the Borda winner $i^*_B$. We thus have,
\begin{align*}
\G_T(i^*_B) - \gamma T - \sum_{t=1}^T\,\gb_t(\I_t) \, &\leq\,  \frac{\log A}{\eta} - \gamma T + \eta\,(e-2)\sum_{t=1}^T\,\frac{A^2}{\gamma^2}
\end{align*}
Then by canceling the $\gamma T$ terms and taking the expected value of both sides,
\begin{align*}
\sum_{t=1}^T\,\E[\gb_t(i^*_B) - \gb_t(\I_t)] \, &\leq\,  \frac{\log A}{\eta} + \eta\,(e-2)\,T\,\frac{A^2}{\gamma^2}
\end{align*}\\
Now we define the hyperparameters $\gamma$ and $\eta$ by using the positive terms $T,\,A,\,\text{and}\, (e-2)$,
\begin{align*}
\eta &= (e-2)^{-1/4}\,\bigg(\frac{\log A}{A^{2/3}\,T}\bigg)^{3/4} \\
\gamma &= (e-2)^{1/4}\,\bigg(\frac{A^2\log A}{T}\bigg)^{1/4}
\end{align*}
which guarantees $\eta>0$ and $\gamma>0$.\\\\
Then we substitute them into the terms on the right-hand side of the inequality,
\begin{align*}
\frac{\log A}{\eta} &= \log A\ (e-2)^{1/4}\,\bigg(\frac{\log A}{A^{2/3}\,T}\bigg)^{-3/4} \\
& = (e-2)^{1/4}\,A^{1/2}\,(\log A)^{1/4}\,T^{3/4} \\\\
\eta\,(e-2)\,T\, \frac{A^2}{\gamma^2} &= (e-2)^{-1/4}\,\bigg(\frac{\log A}{A^{2/3}\,T}\bigg)^{3/4}\,(e-2)\,T\, A^2\,(e-2)^{-1/2}\,\bigg(\frac{A^2\log A}{T}\bigg)^{-1/2} \\
&= (e-2)^{1/4}\,A^{1/2}\,(\log A)^{1/4}\,T^{3/4}
\end{align*}
Combining the terms achieves the desired result.\\\\\\
Finally, we determine the required $T$ such that $\gamma<1$ and $\eta\,A/\gamma^2\leq1$ hold,
\begin{align*}
(e-2)^{1/4}\,\bigg(\frac{A^2\log A}{T}\bigg)^{1/4} &< 1 \\
\frac{A^2\log A}{T} &< (e-2)^{-1} \\
T &> (e-2)\,A^2\log A \\\\\\
\frac{\eta\,A}{\gamma^2} &\leq 1 \\
(e-2)^{-1/4}\,\bigg(\frac{\log A}{A^{2/3}\,T}\bigg)^{3/4}\,&A\,(e-2)^{-1/2}\,\bigg(\frac{A^2\log A}{T}\bigg)^{-1/2} \leq 1 \\
(e-2)^{-3/4}\,(\log A)^{1/4}\,A^{-2/4}\,T^{-1/4} &\leq 1 \\
T &\geq (e-2)^{-3}\,A^{-2}\log A
\end{align*}\\
Since $(e-2)\,A^2 > (e-2)^{-3}\,A^{-2}$ for all $A\geq2$, this gives the required $T$.

$\hfill\blacksquare$

\newpage
\section{Experimental Results}

\begin{figure}[H]
	\centering
	\begin{subfigure}[b]{0.75\textwidth}
		\centering
		\includegraphics[width=\textwidth]{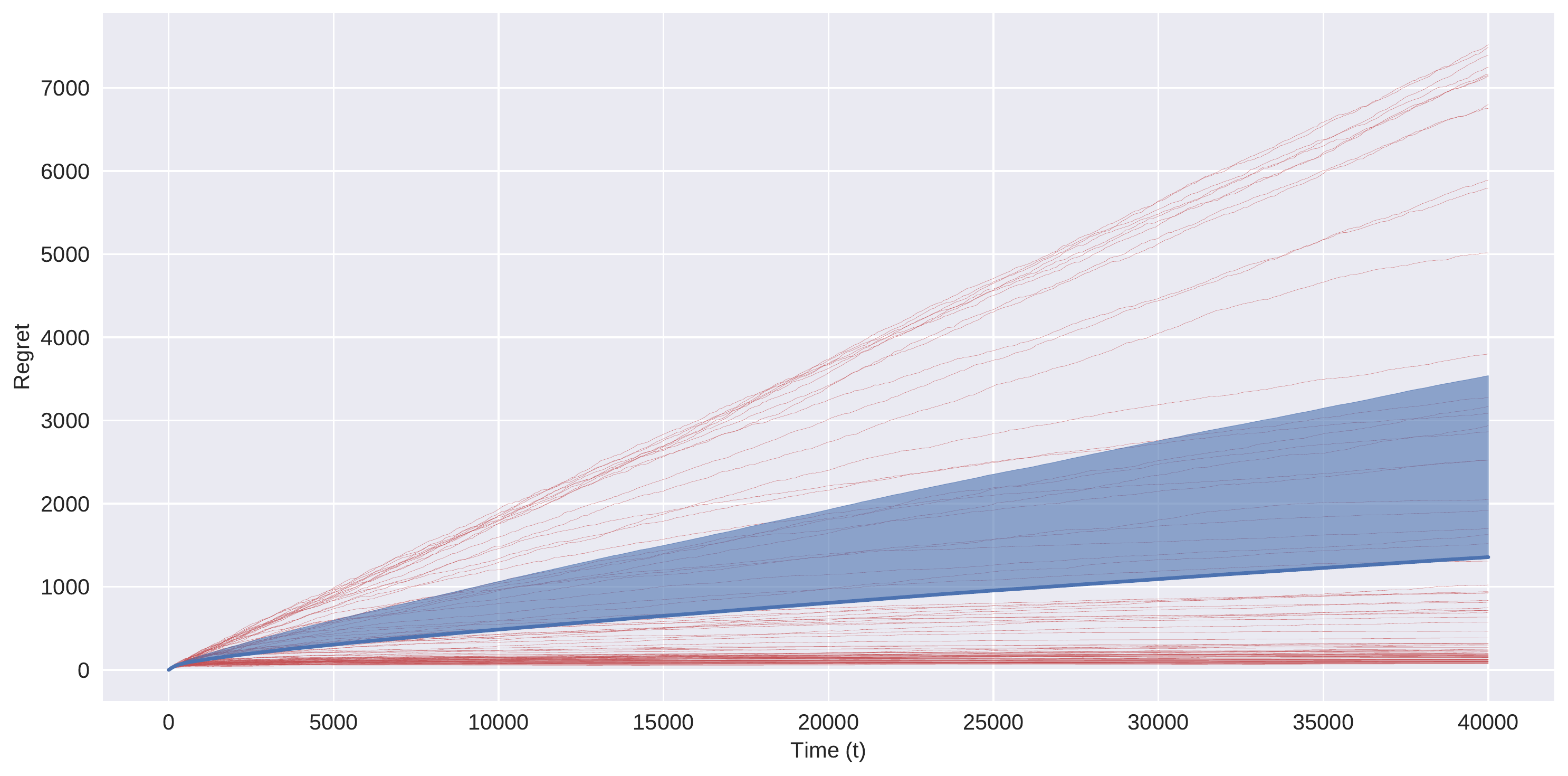}
		\subcaption{}
	\end{subfigure}\\
	\begin{subfigure}[b]{0.75\textwidth}
		\centering
		\includegraphics[width=\textwidth]{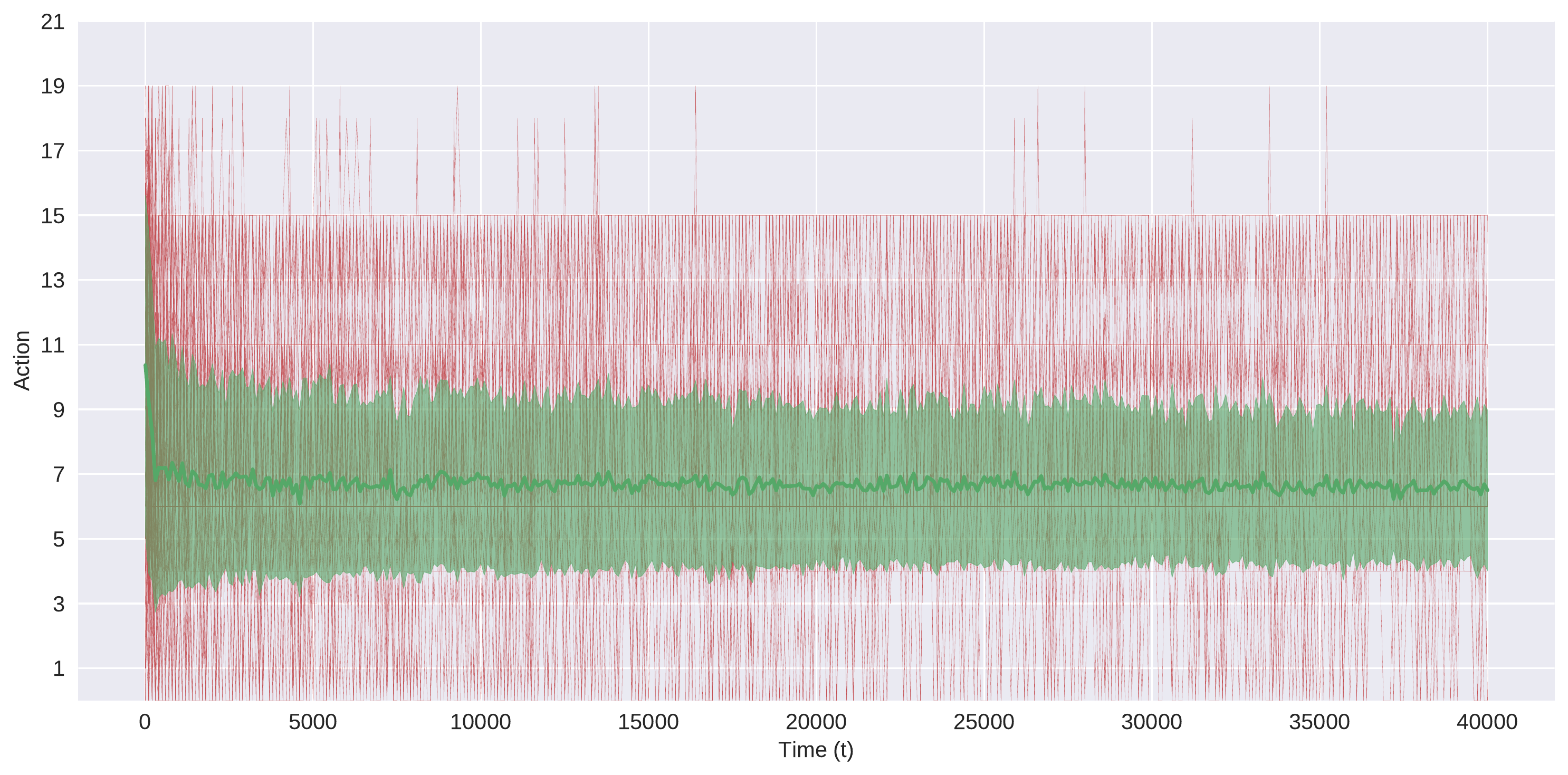}
		\subcaption{}
	\end{subfigure}\\
	\centering
	\begin{subfigure}[b]{0.75\textwidth}
		\centering
		\includegraphics[width=\textwidth]{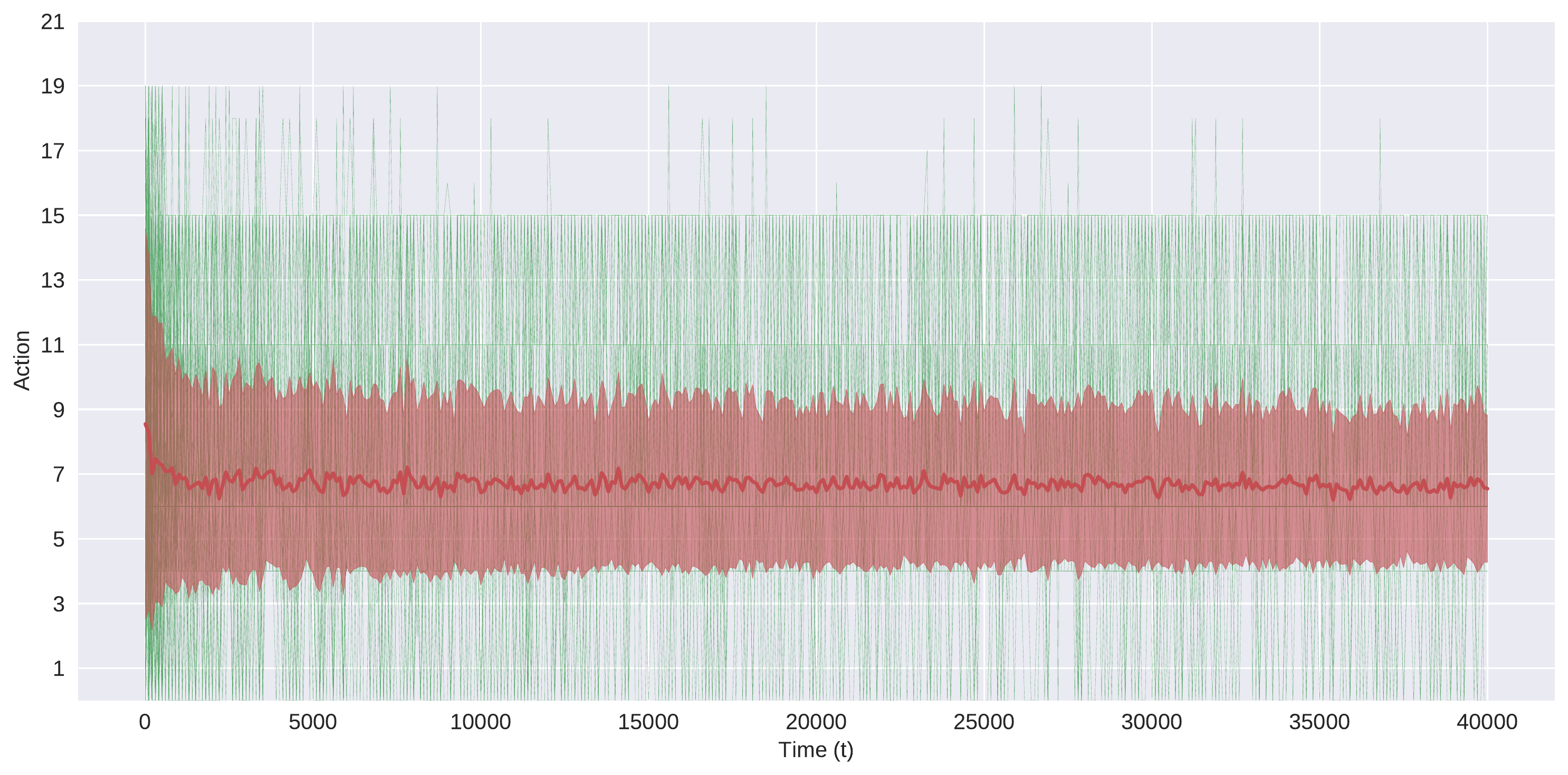}
		\subcaption{}
	\end{subfigure}
	\caption{\label{S1tsmm} {\bf Condorcet Scenario - Thompson Sampling (Maximin)} Off-color lines are individual runs, thick lines are mean over runs, shading is standard deviation. (a) algorithm specified regret, (b) $\I_t$ selections, and (c) $\J_t$ selections.  }
\end{figure}

\begin{figure}[H]
	\centering
	\begin{subfigure}[b]{0.75\textwidth}
		\centering
		\includegraphics[width=\textwidth]{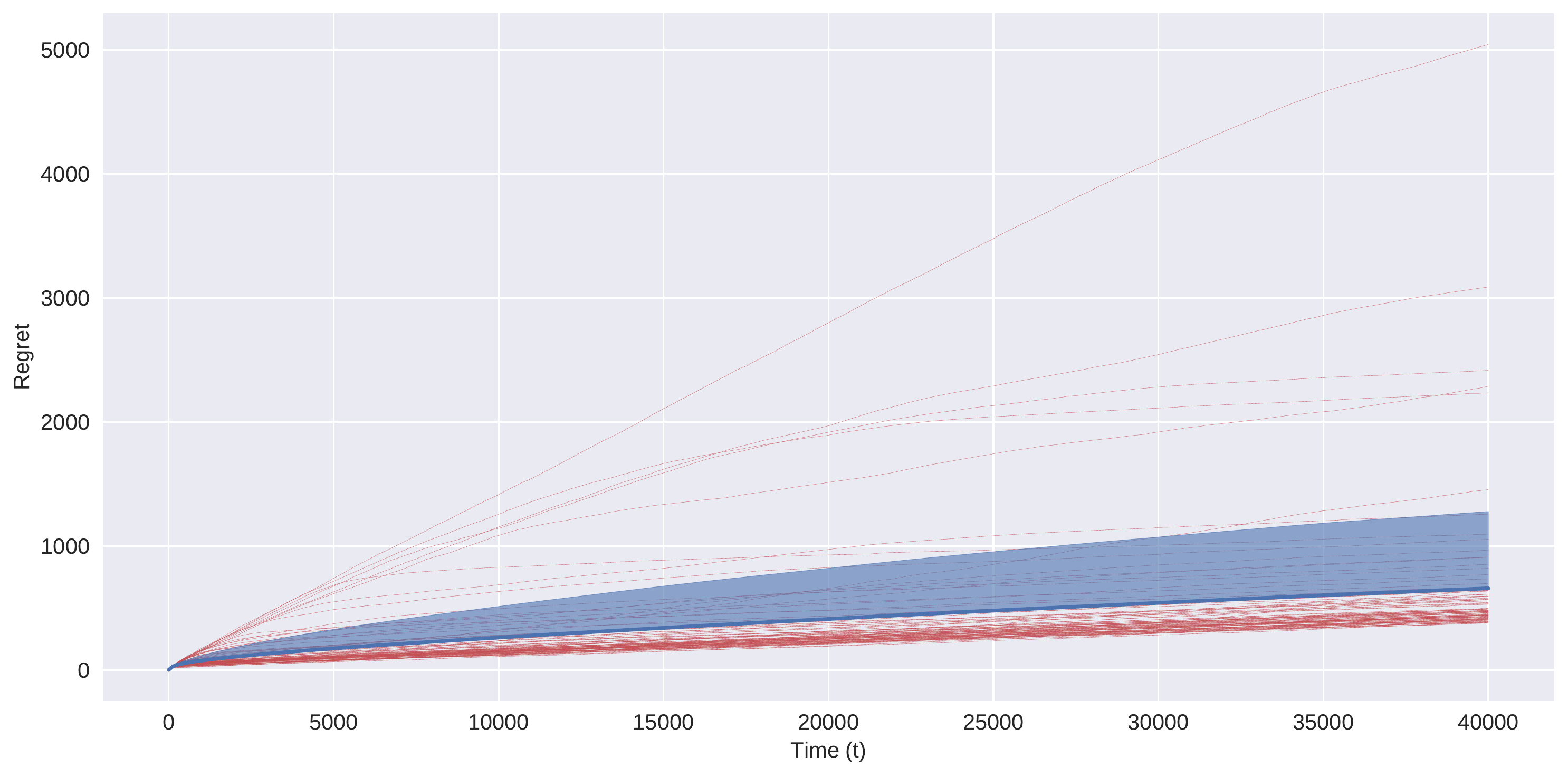}
		\subcaption{}
	\end{subfigure}\\
	\begin{subfigure}[b]{0.75\textwidth}
		\centering
		\includegraphics[width=\textwidth]{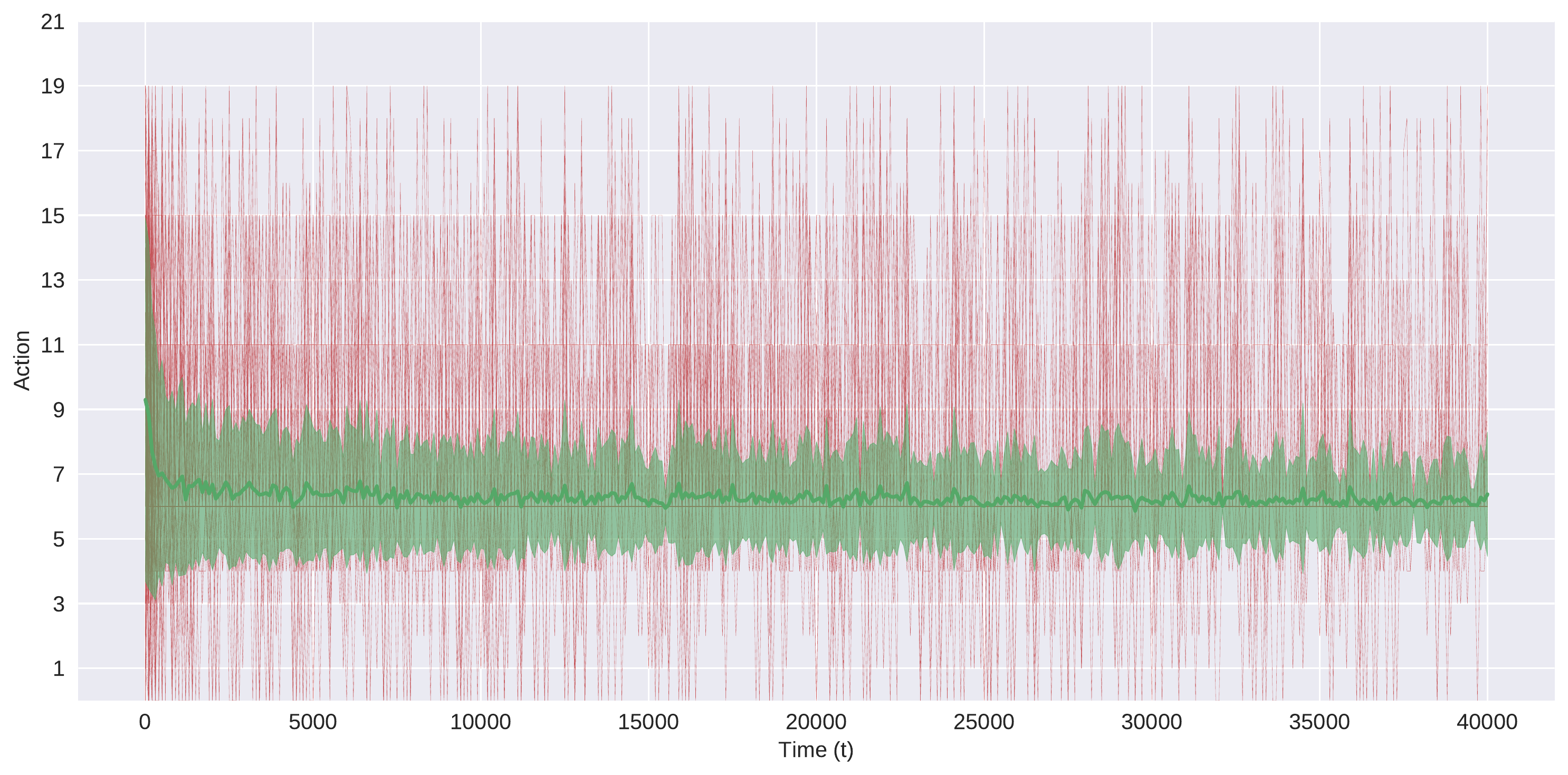}
		\subcaption{}
	\end{subfigure}\\
	\centering
	\begin{subfigure}[b]{0.75\textwidth}
		\centering
		\includegraphics[width=\textwidth]{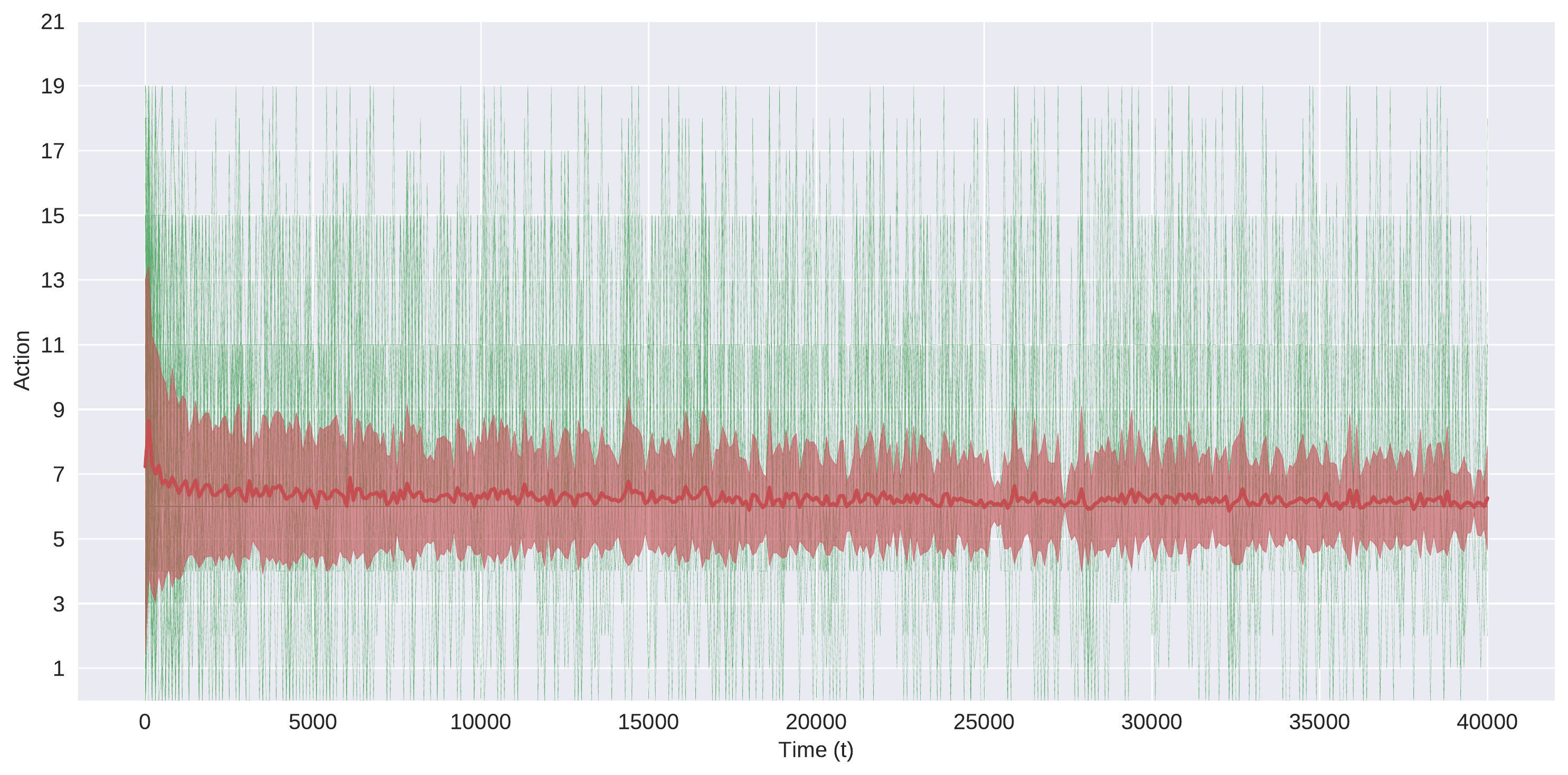}
		\subcaption{}
	\end{subfigure}
	\caption{\label{S1tsbr} {\bf Condorcet Scenario - Thompson Sampling (Borda)} Off-color lines are individual runs, thick lines are mean over runs, shading is standard deviation. (a) algorithm specified regret, (b) $\I_t$ selections, and (c) $\J_t$ selections.  }
\end{figure}

\begin{figure}[H]
	\centering
	\begin{subfigure}[b]{0.5\textwidth}
		\centering
		\includegraphics[width=\textwidth]{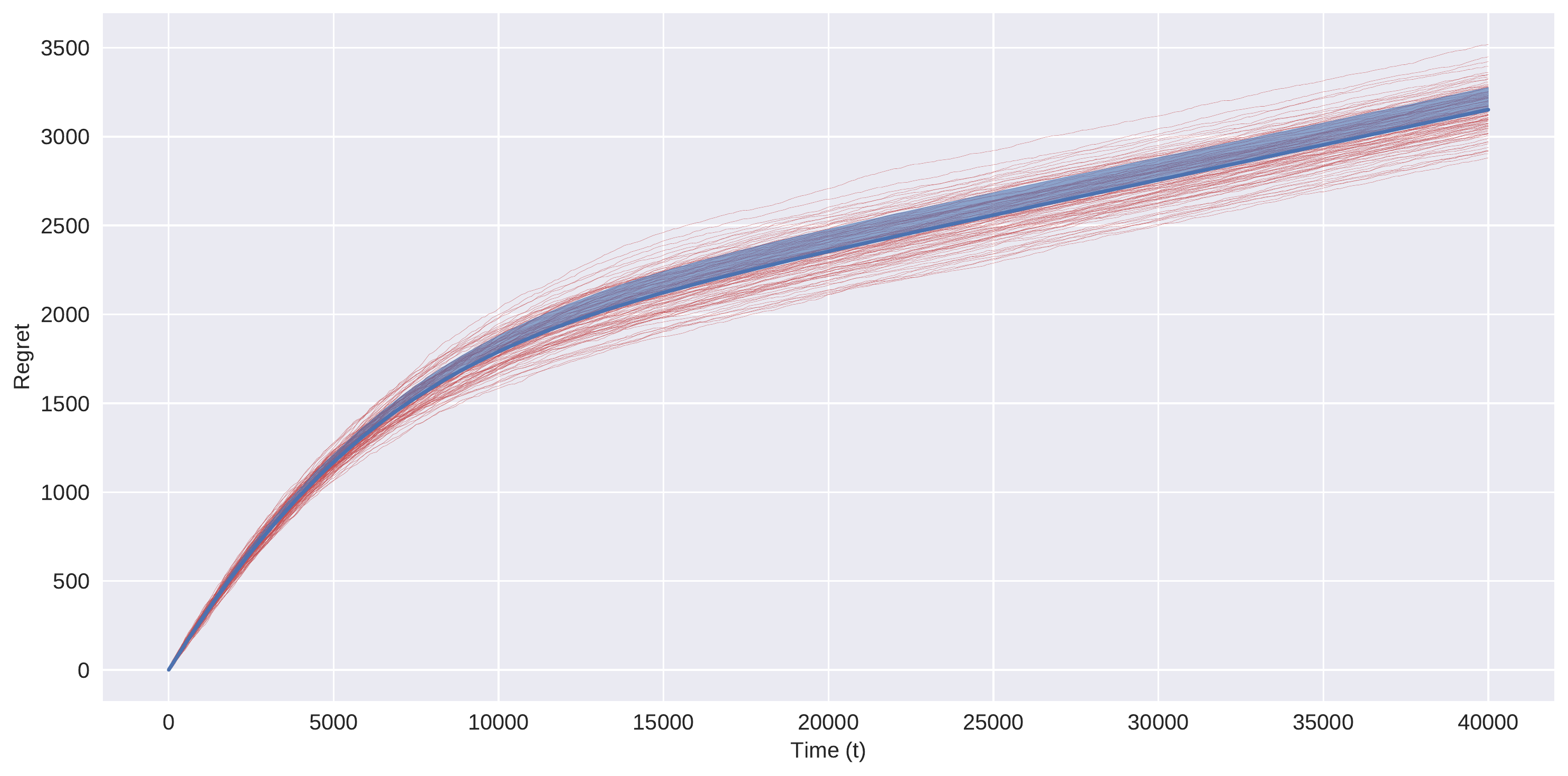}
		\subcaption{}
	\end{subfigure}\\
	\begin{subfigure}[b]{0.5\textwidth}
		\centering
		\includegraphics[width=\textwidth]{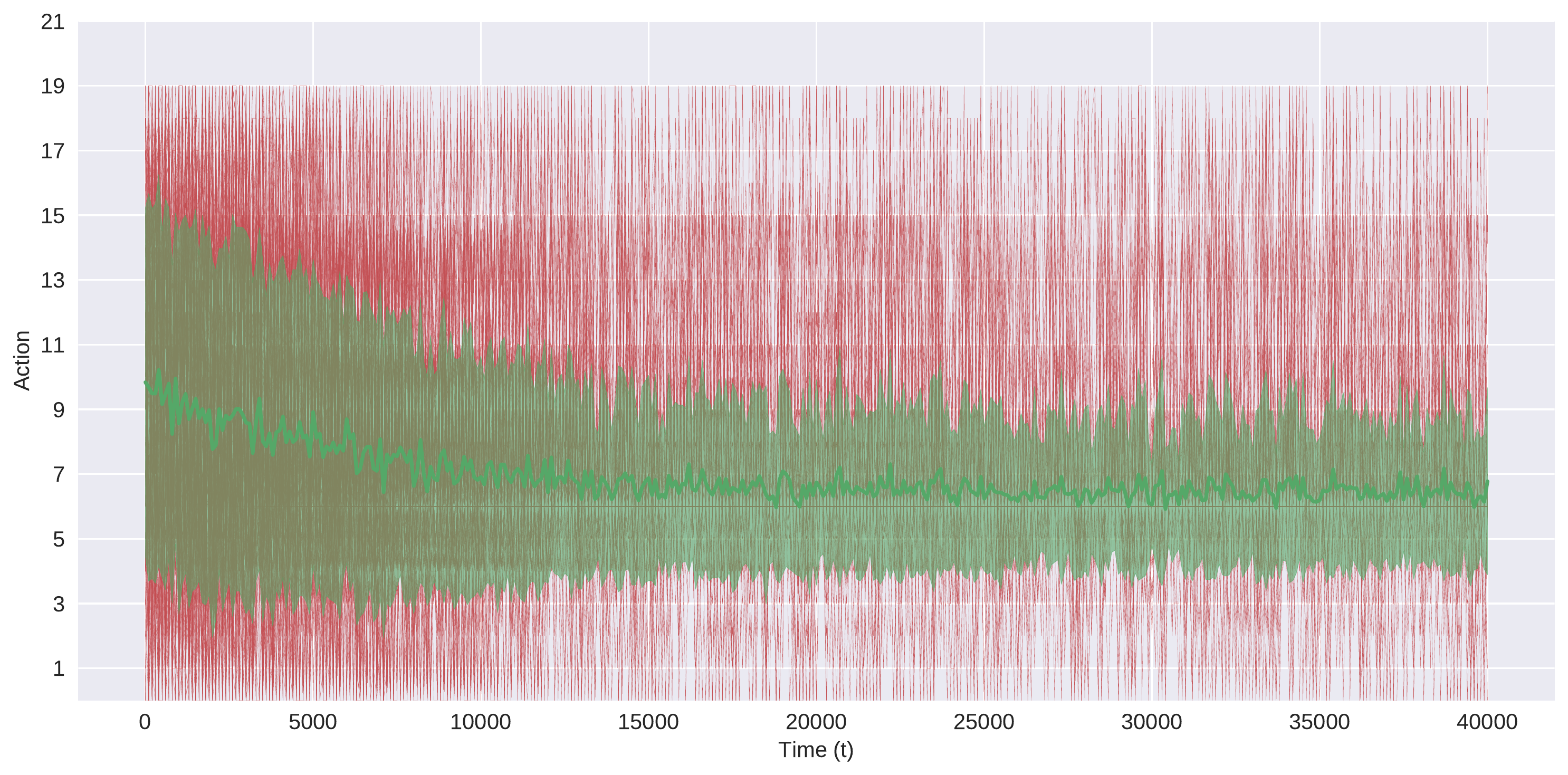}
		\subcaption{}
	\end{subfigure}\\
	\centering
	\begin{subfigure}[b]{0.5\textwidth}
		\centering
		\includegraphics[width=\textwidth]{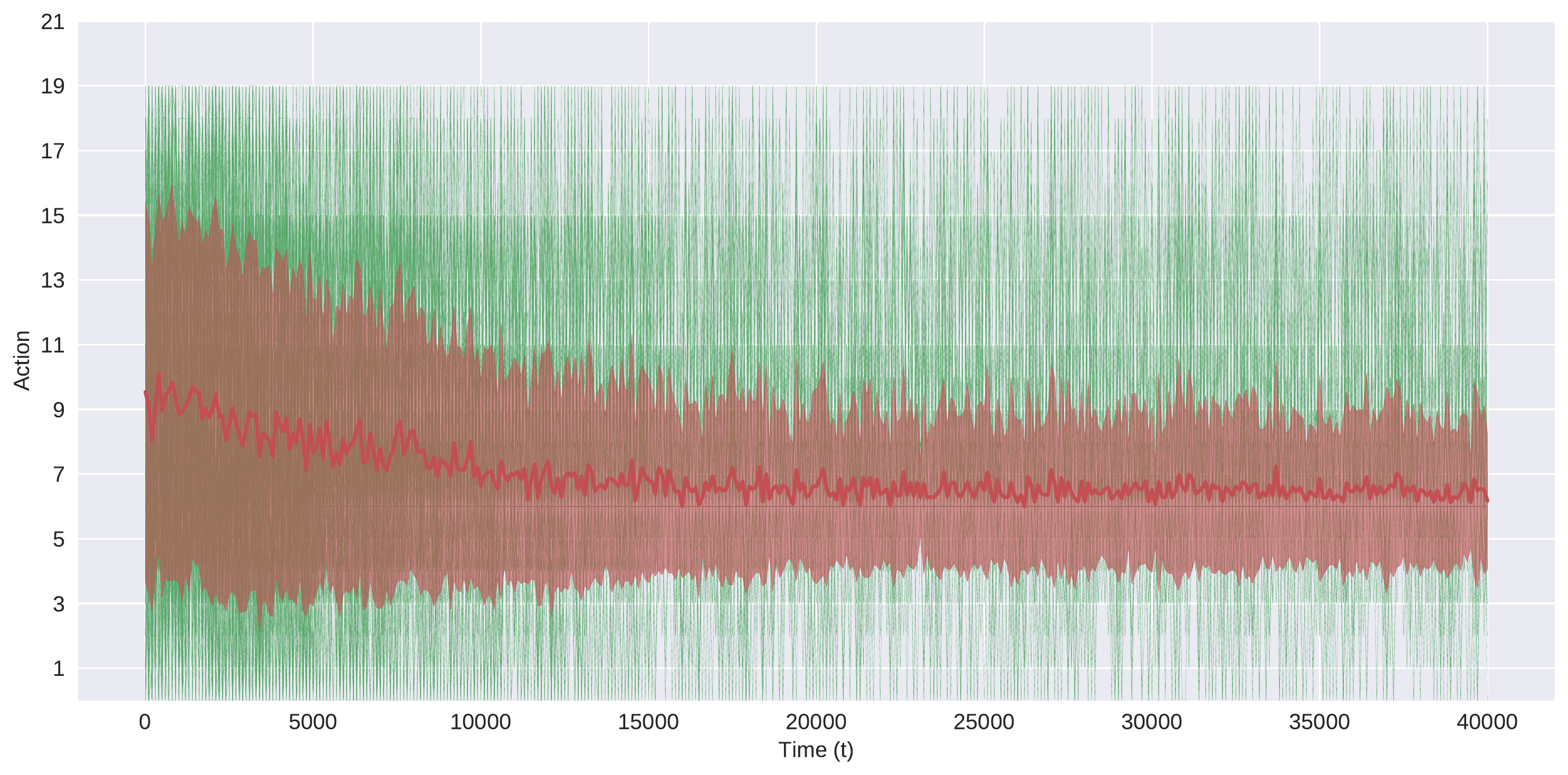}
		\subcaption{}
	\end{subfigure}\\
	\begin{subfigure}[b]{0.5\textwidth}
		\centering
		\includegraphics[width=\textwidth]{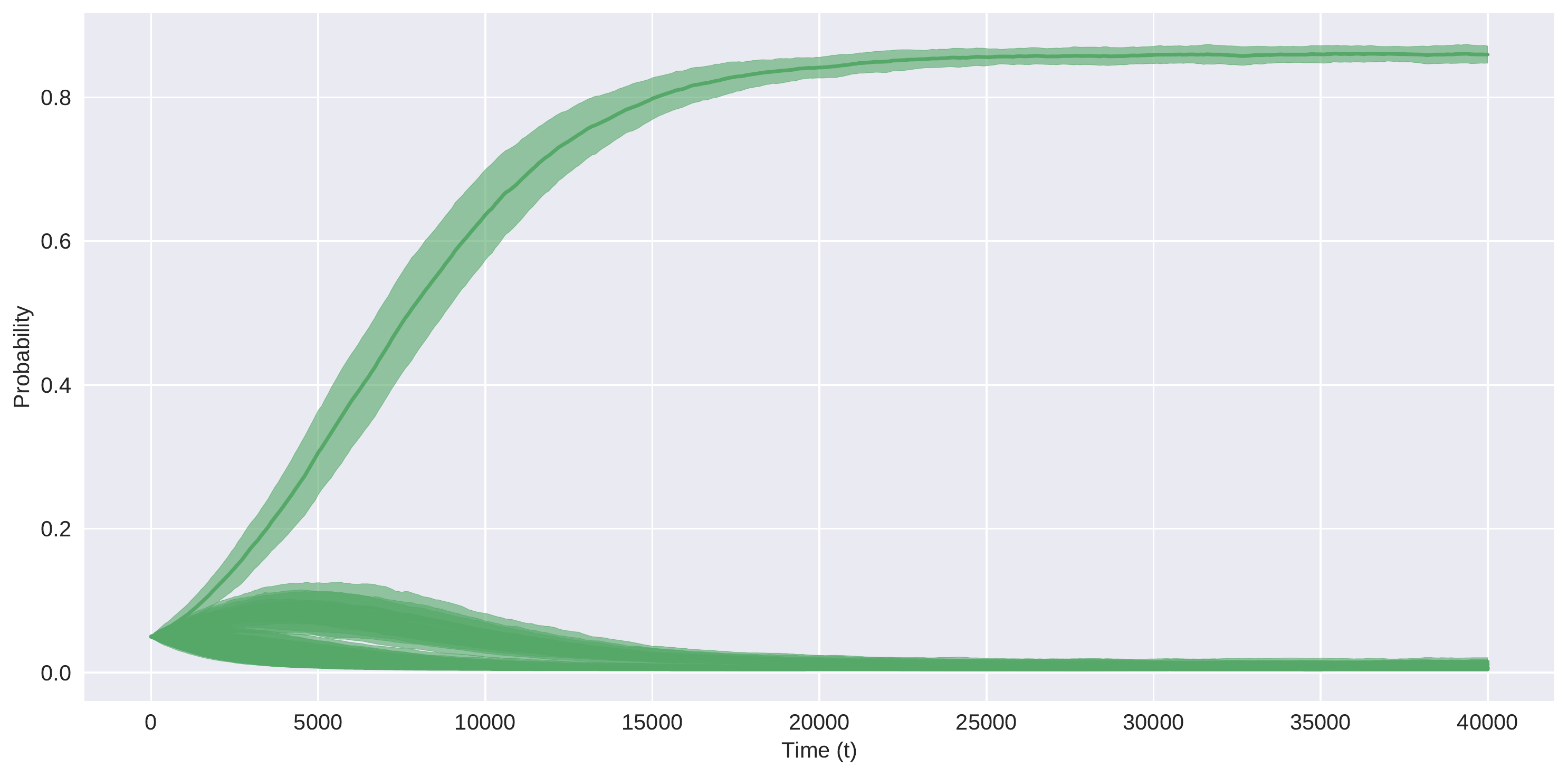}
		\subcaption{}
	\end{subfigure}\\
	\begin{subfigure}[b]{0.5\textwidth}
		\centering
		\includegraphics[width=\textwidth]{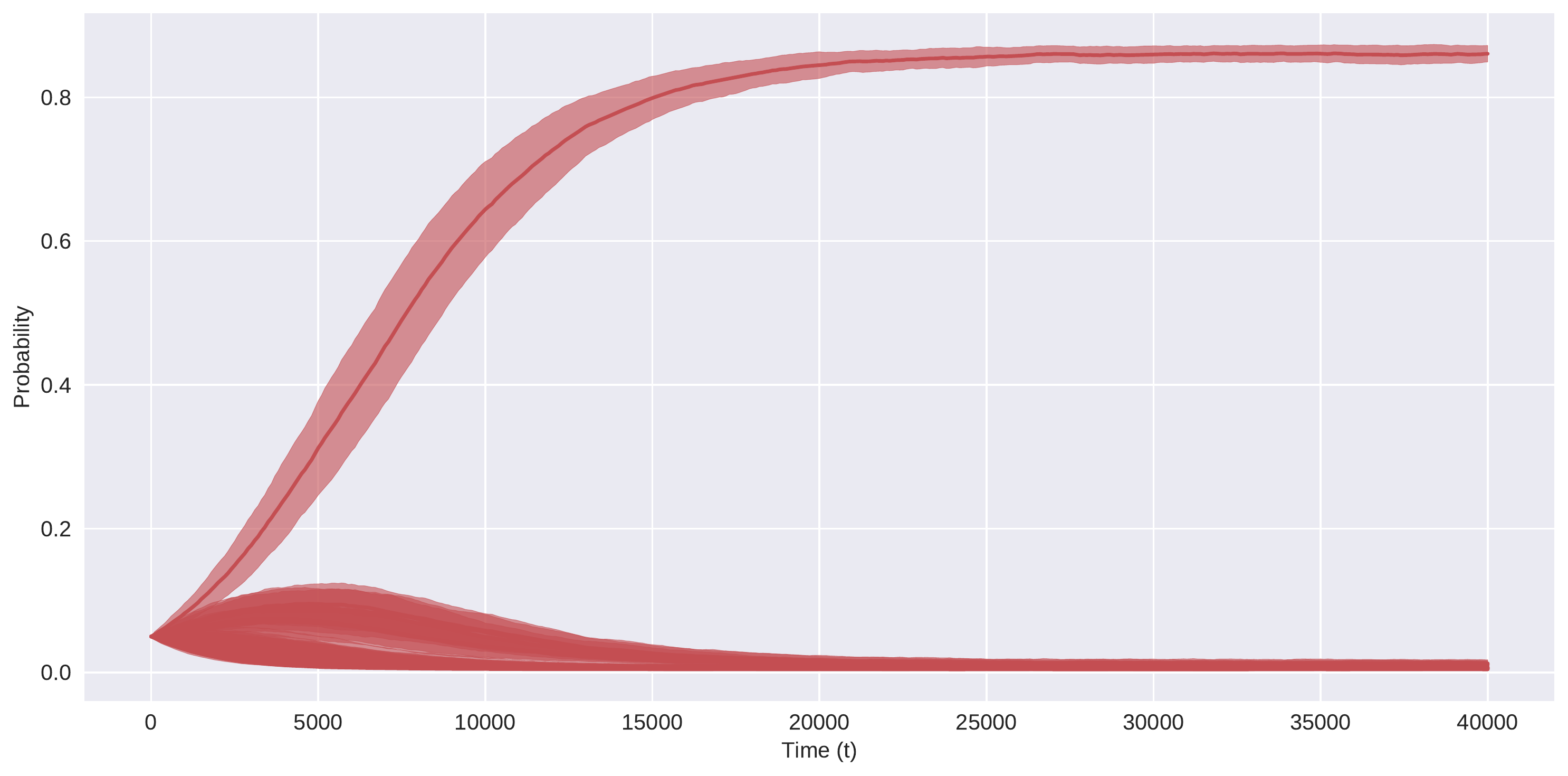}
		\subcaption{}
	\end{subfigure}
	\caption{\label{S1exp3p} {\bf Condorcet Scenario - SparringExp3.P} Off-color lines are individual runs, thick lines are mean over runs, shading is standard deviation. (a) algorithm specified regret, (b) $\I_t$ selections, (c) $\J_t$ selections, (d) $\p_t$ strategy, and (e) $\q_t$ strategy.  }
\end{figure}

\begin{figure}[H]
	\centering
	\begin{subfigure}[b]{0.65\textwidth}
		\centering
		\includegraphics[width=\textwidth]{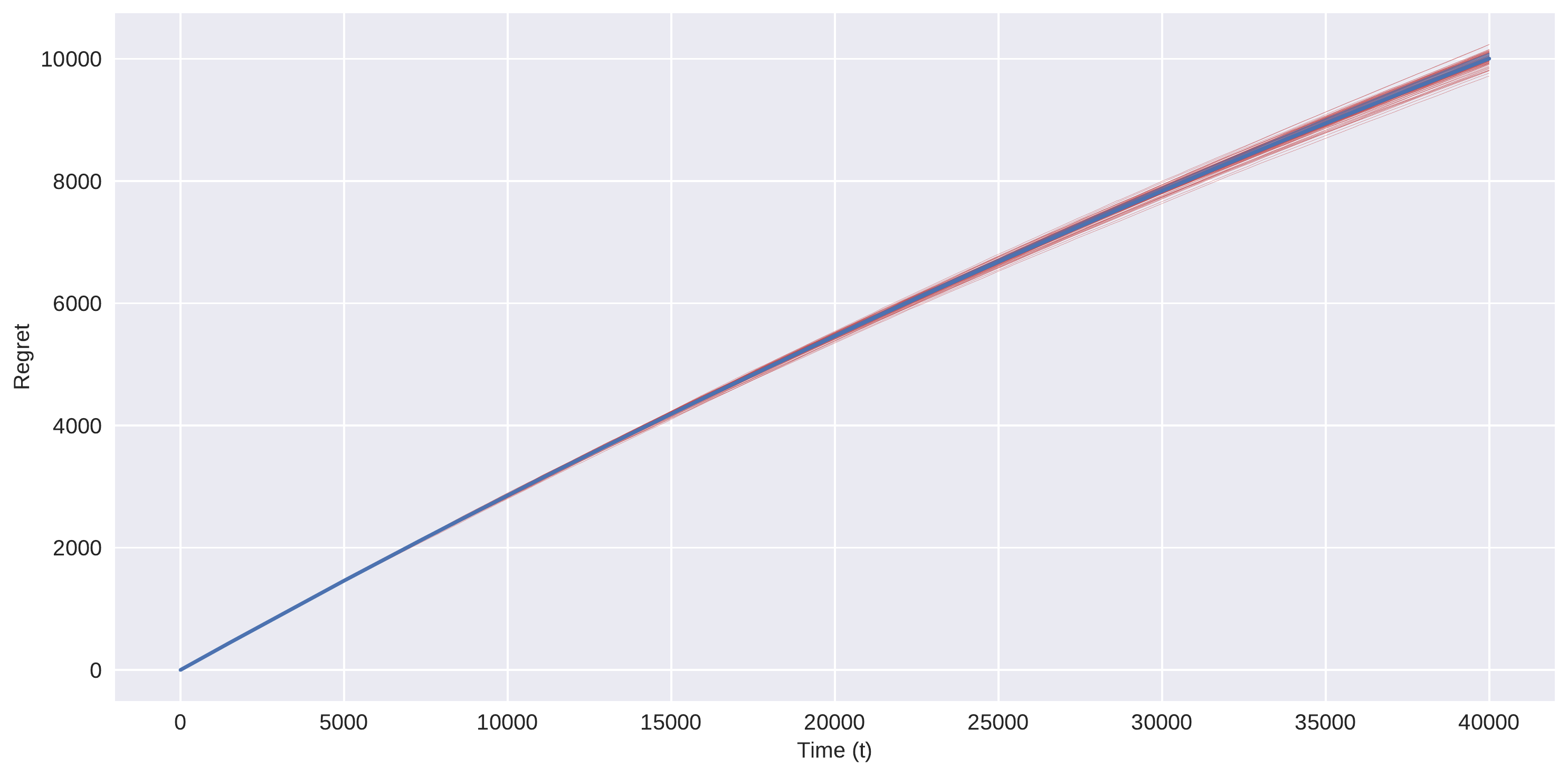}
		\subcaption{}
	\end{subfigure}\\
	\begin{subfigure}[b]{0.65\textwidth}
		\centering
		\includegraphics[width=\textwidth]{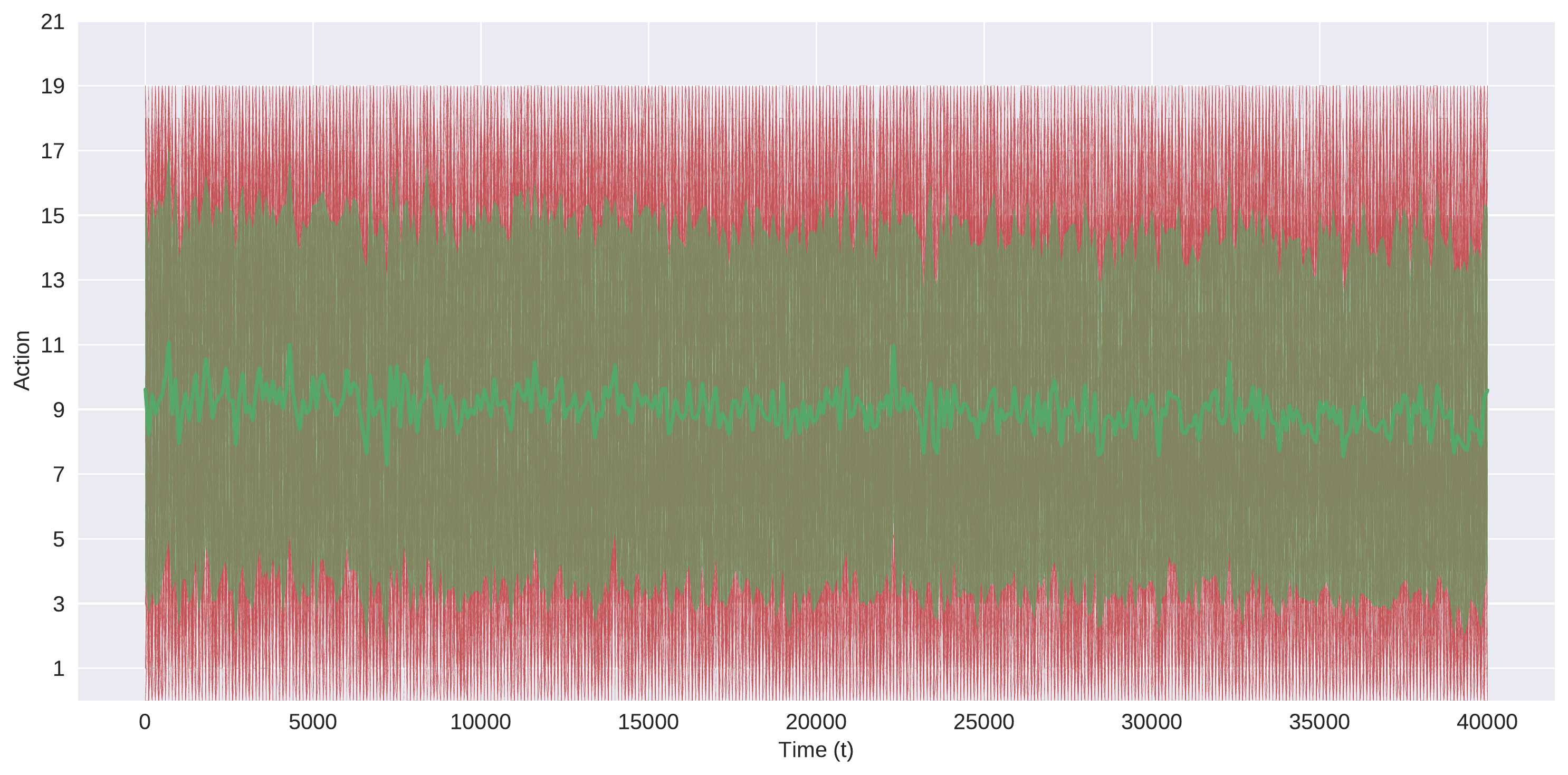}
		\subcaption{}
	\end{subfigure}\\
	\centering
	\begin{subfigure}[b]{0.65\textwidth}
		\centering
		\includegraphics[width=\textwidth]{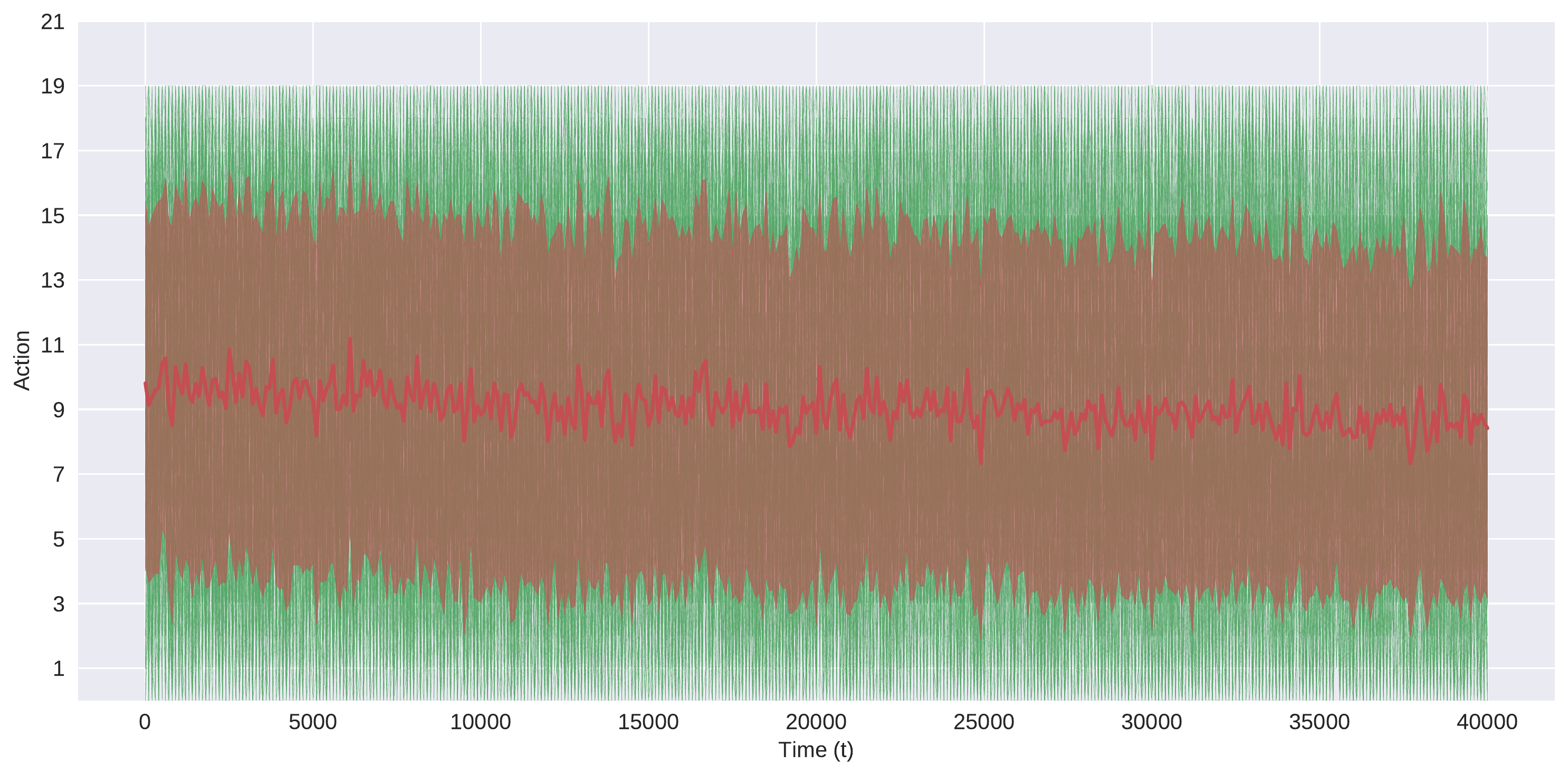}
		\subcaption{}
	\end{subfigure}\\
	\begin{subfigure}[b]{0.65\textwidth}
		\centering
		\includegraphics[width=\textwidth]{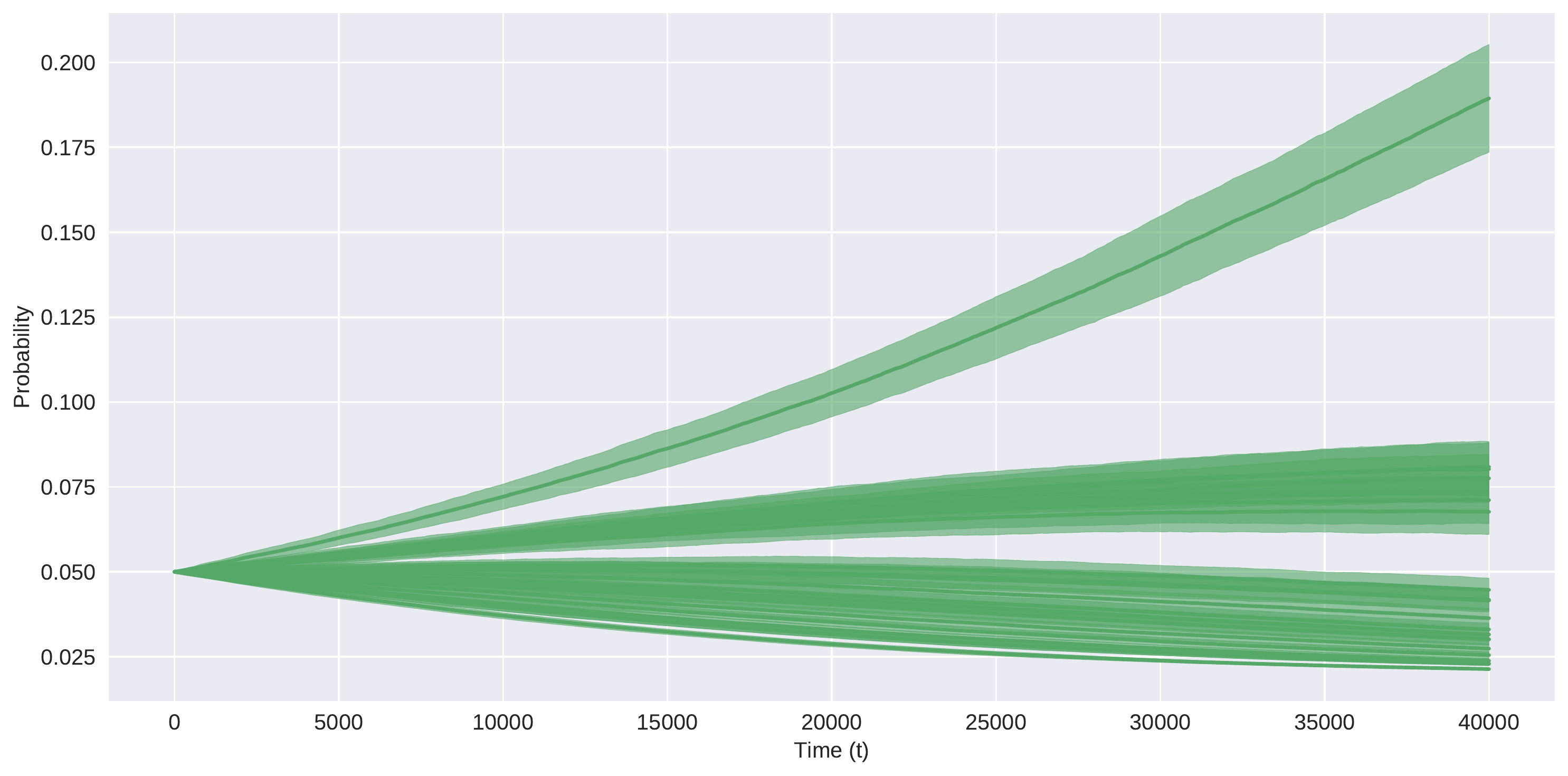}
		\subcaption{}
	\end{subfigure}
	\caption{\label{S1pm} {\bf Condorcet Scenario - Partial Monitoring} Off-color lines are individual runs, thick lines are mean over runs, shading is standard deviation. (a) algorithm specified regret, (b) $\I_t$ selections, (c) $\J_t$ selections, and (d) $\p_t$ strategy.  }
\end{figure}

\begin{figure}[H]
	\centering
	\begin{subfigure}[b]{0.75\textwidth}
		\centering
		\includegraphics[width=\textwidth]{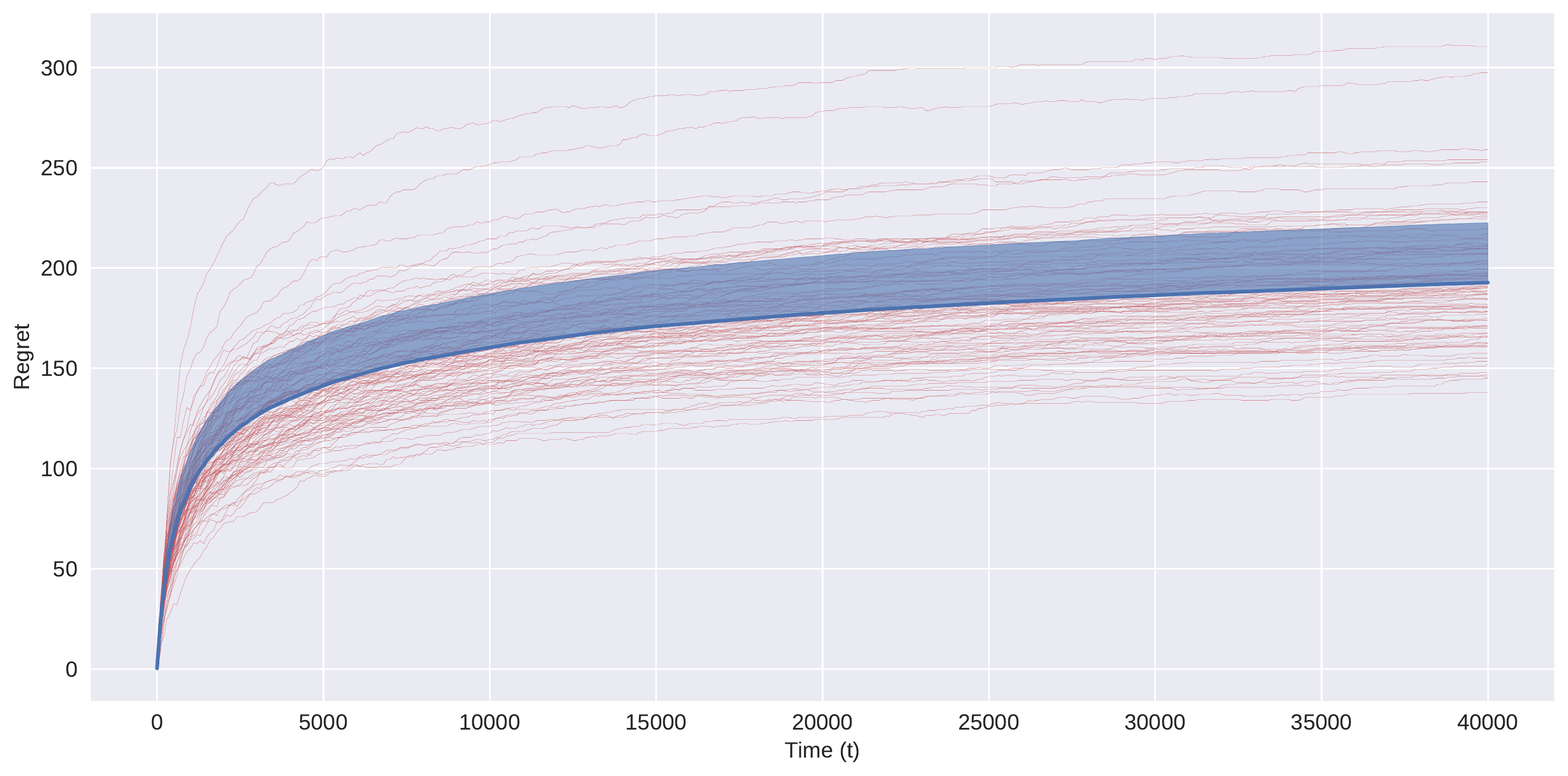}
		\subcaption{}
	\end{subfigure}\\
	\begin{subfigure}[b]{0.75\textwidth}
		\centering
		\includegraphics[width=\textwidth]{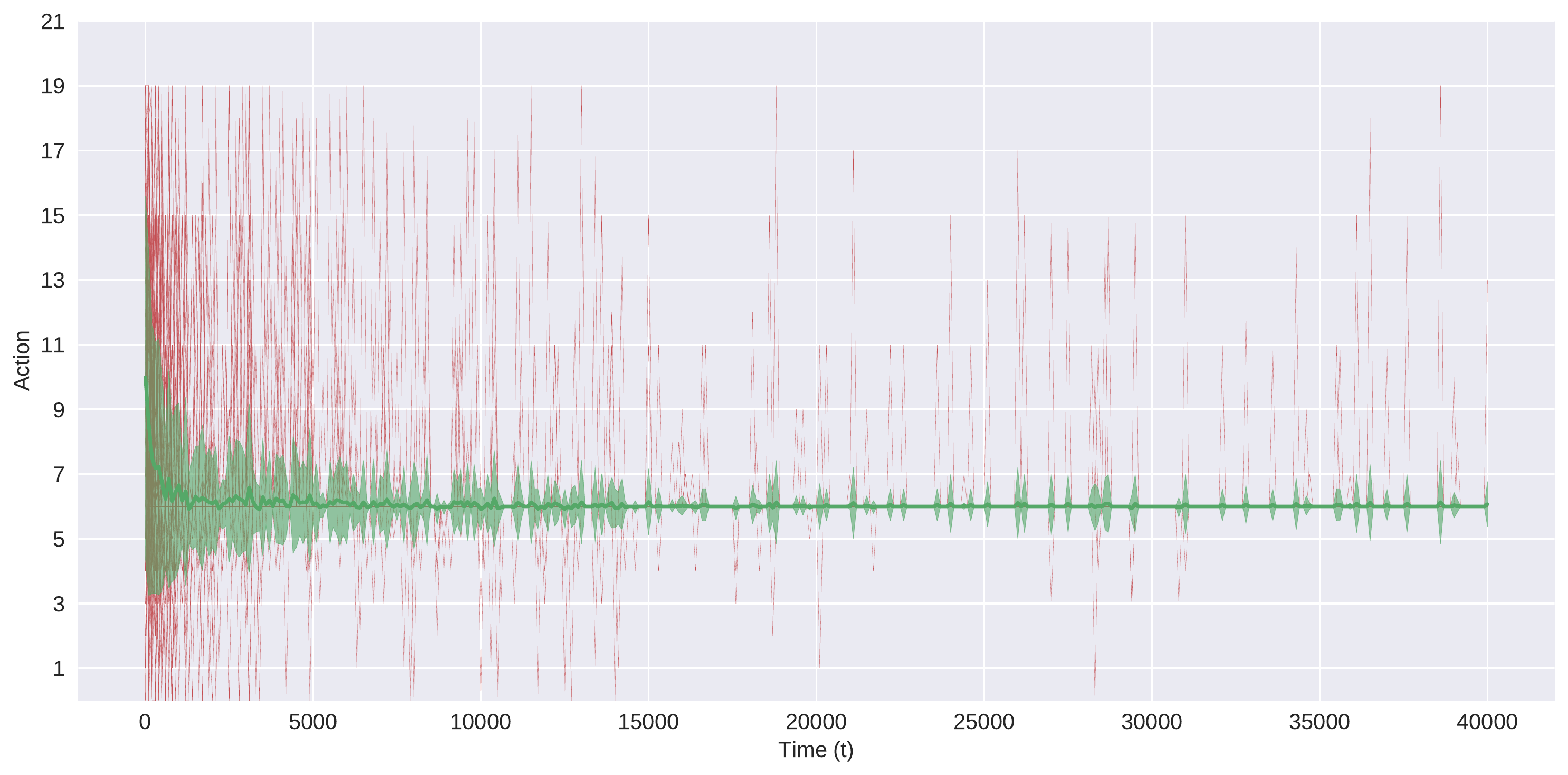}
		\subcaption{}
	\end{subfigure}\\
	\centering
	\begin{subfigure}[b]{0.75\textwidth}
		\centering
		\includegraphics[width=\textwidth]{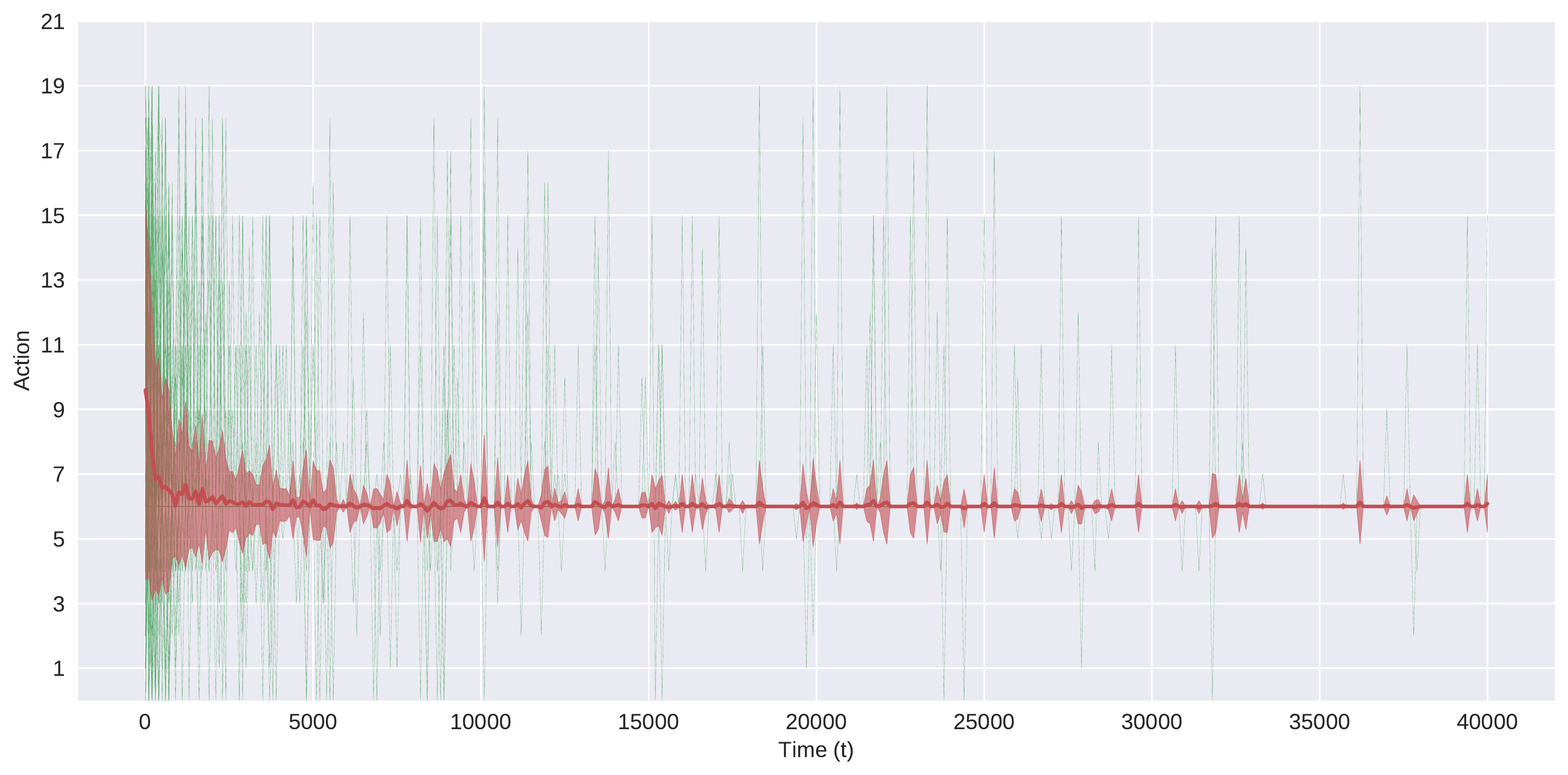}
		\subcaption{}
	\end{subfigure}
	\caption{\label{S1iss} {\bf Condorcet Scenario - ISS} Off-color lines are individual runs, thick lines are mean over runs, shading is standard deviation. (a) algorithm specified regret, (b) $\I_t$ selections, and (c) $\J_t$ selections.  }
\end{figure}

\begin{figure}[H]
	\centering
	\begin{subfigure}[b]{0.75\textwidth}
		\centering
		\includegraphics[width=\textwidth]{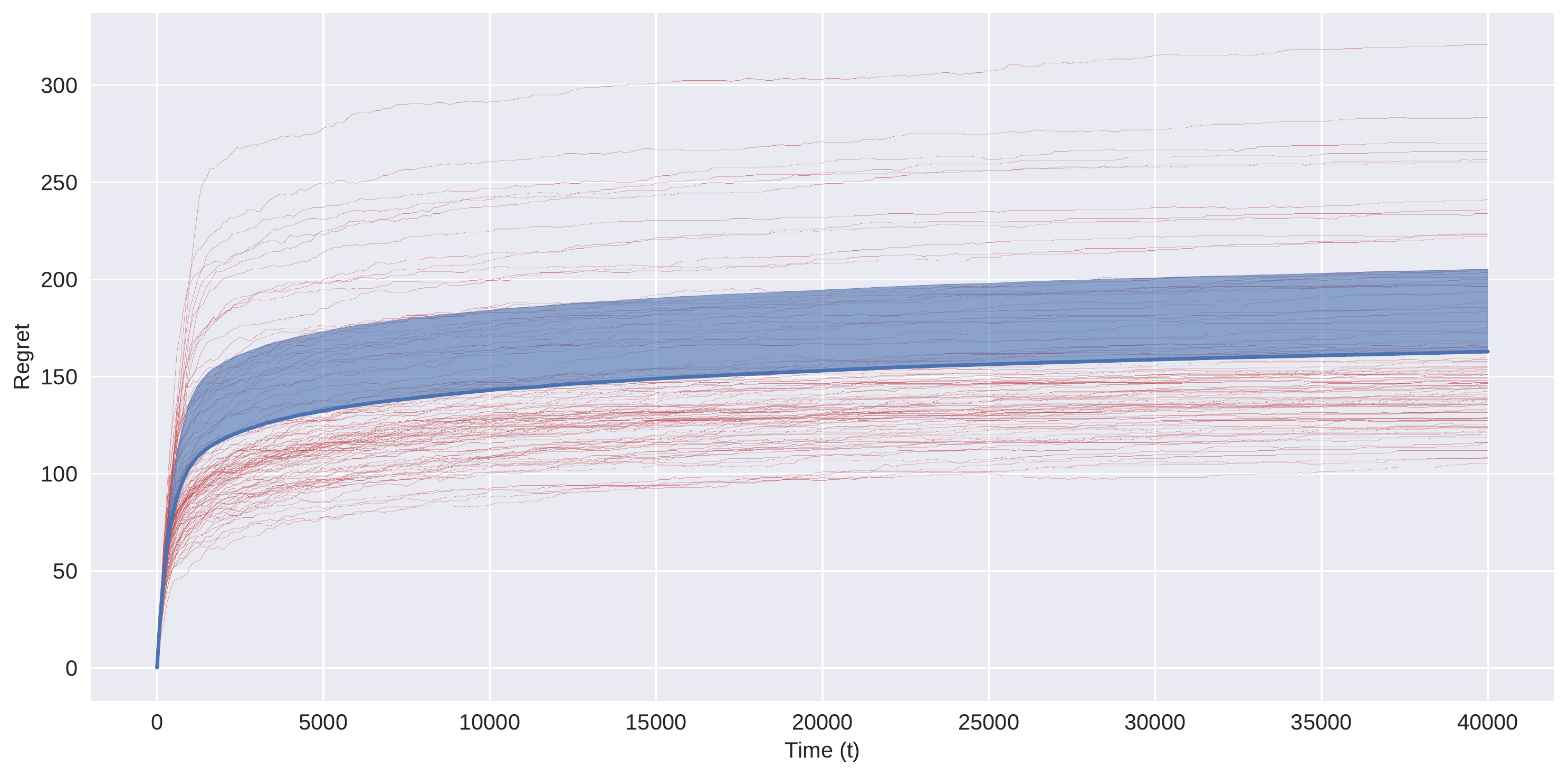}
		\subcaption{}
	\end{subfigure}\\
	\begin{subfigure}[b]{0.75\textwidth}
		\centering
		\includegraphics[width=\textwidth]{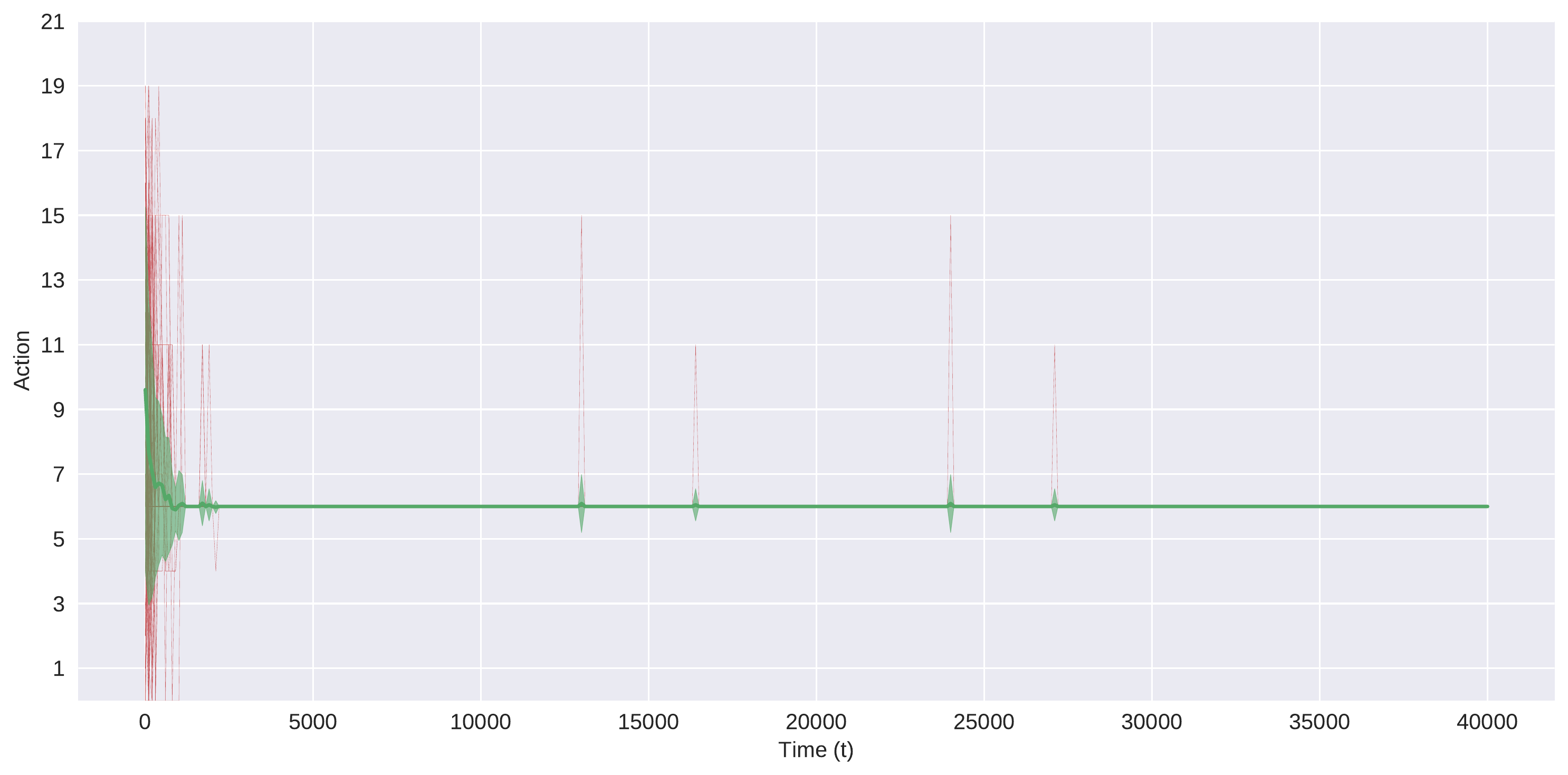}
		\subcaption{}
	\end{subfigure}\\
	\centering
	\begin{subfigure}[b]{0.75\textwidth}
		\centering
		\includegraphics[width=\textwidth]{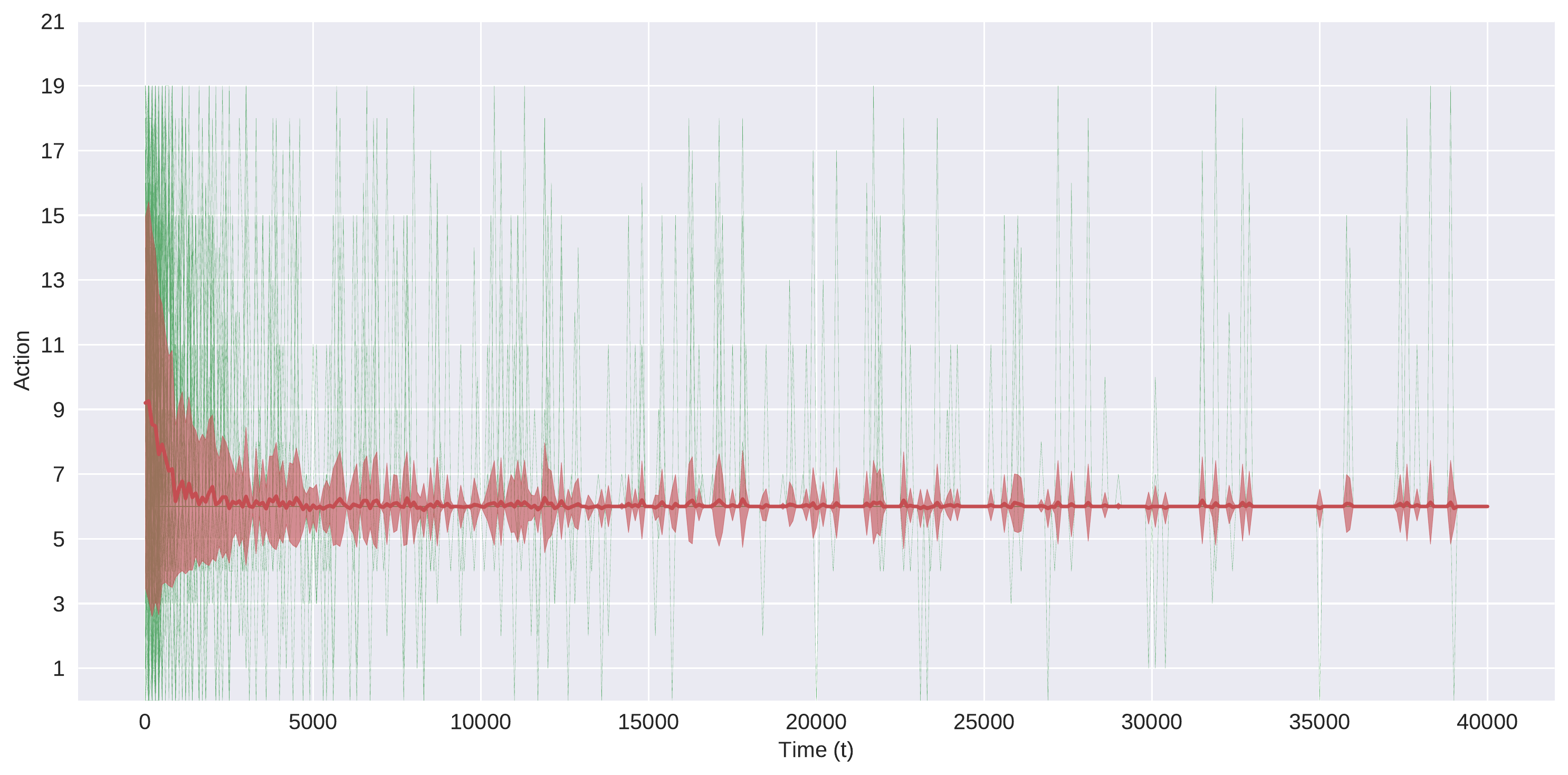}
		\subcaption{}
	\end{subfigure}
	\caption{\label{S1dts} {\bf Condorcet Scenario - DTS} Off-color lines are individual runs, thick lines are mean over runs, shading is standard deviation. (a) algorithm specified regret, (b) $\I_t$ selections, and (c) $\J_t$ selections.  }
\end{figure}

\begin{figure}[H]
	\centering
	\begin{subfigure}[b]{0.75\textwidth}
		\centering
		\includegraphics[width=\textwidth]{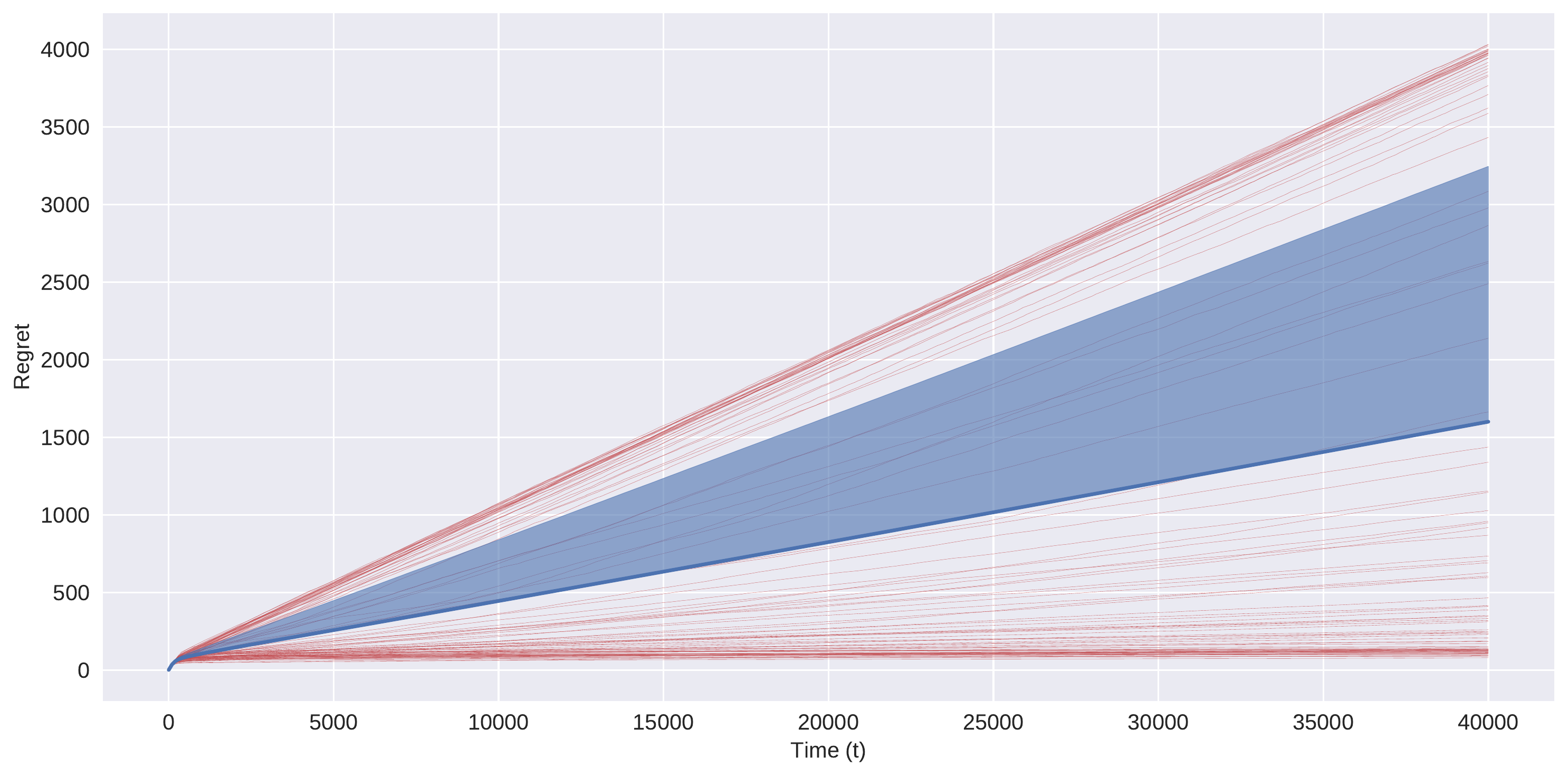}
		\subcaption{}
	\end{subfigure}\\
	\begin{subfigure}[b]{0.75\textwidth}
		\centering
		\includegraphics[width=\textwidth]{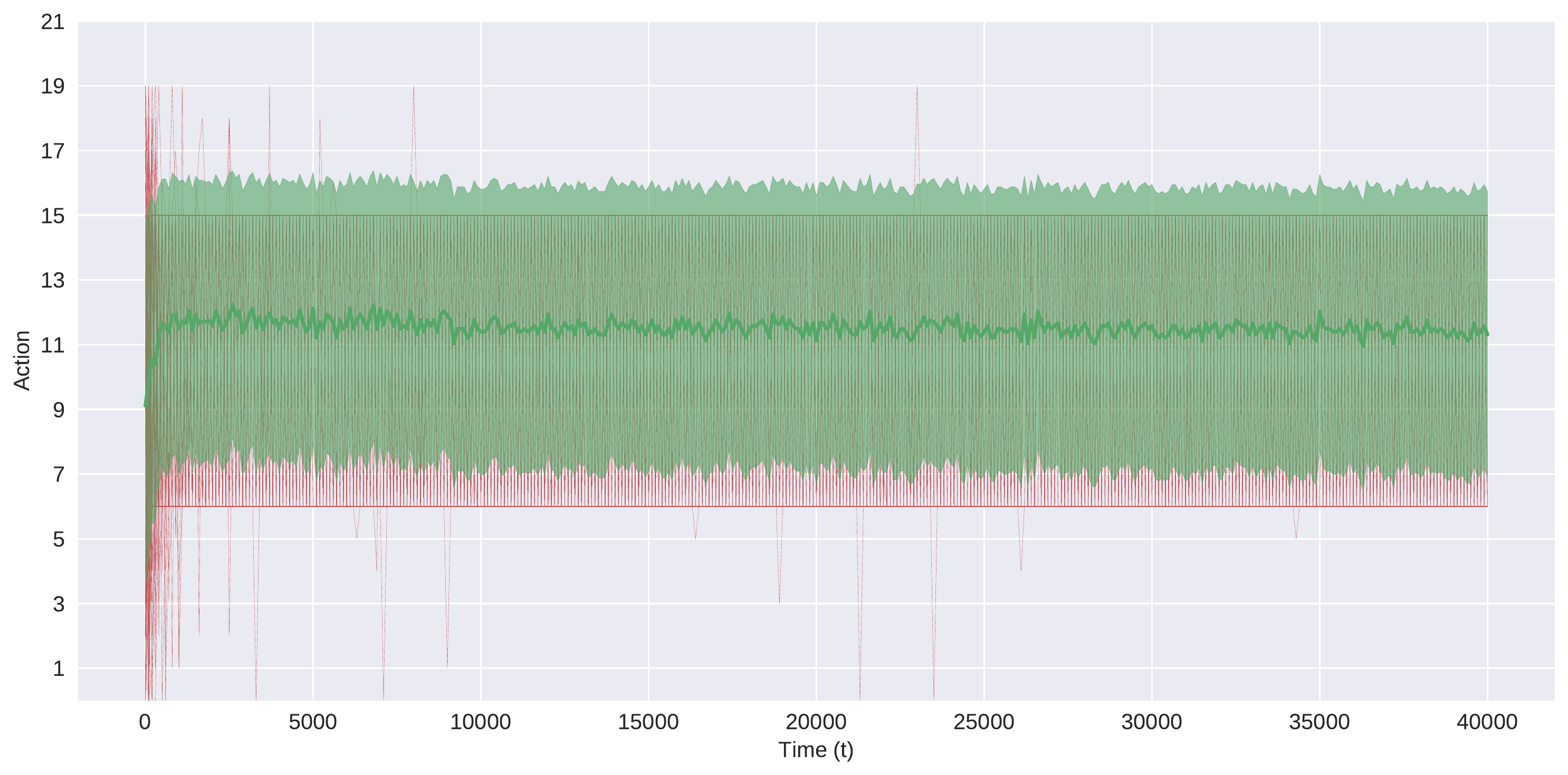}
		\subcaption{}
	\end{subfigure}\\
	\centering
	\begin{subfigure}[b]{0.75\textwidth}
		\centering
		\includegraphics[width=\textwidth]{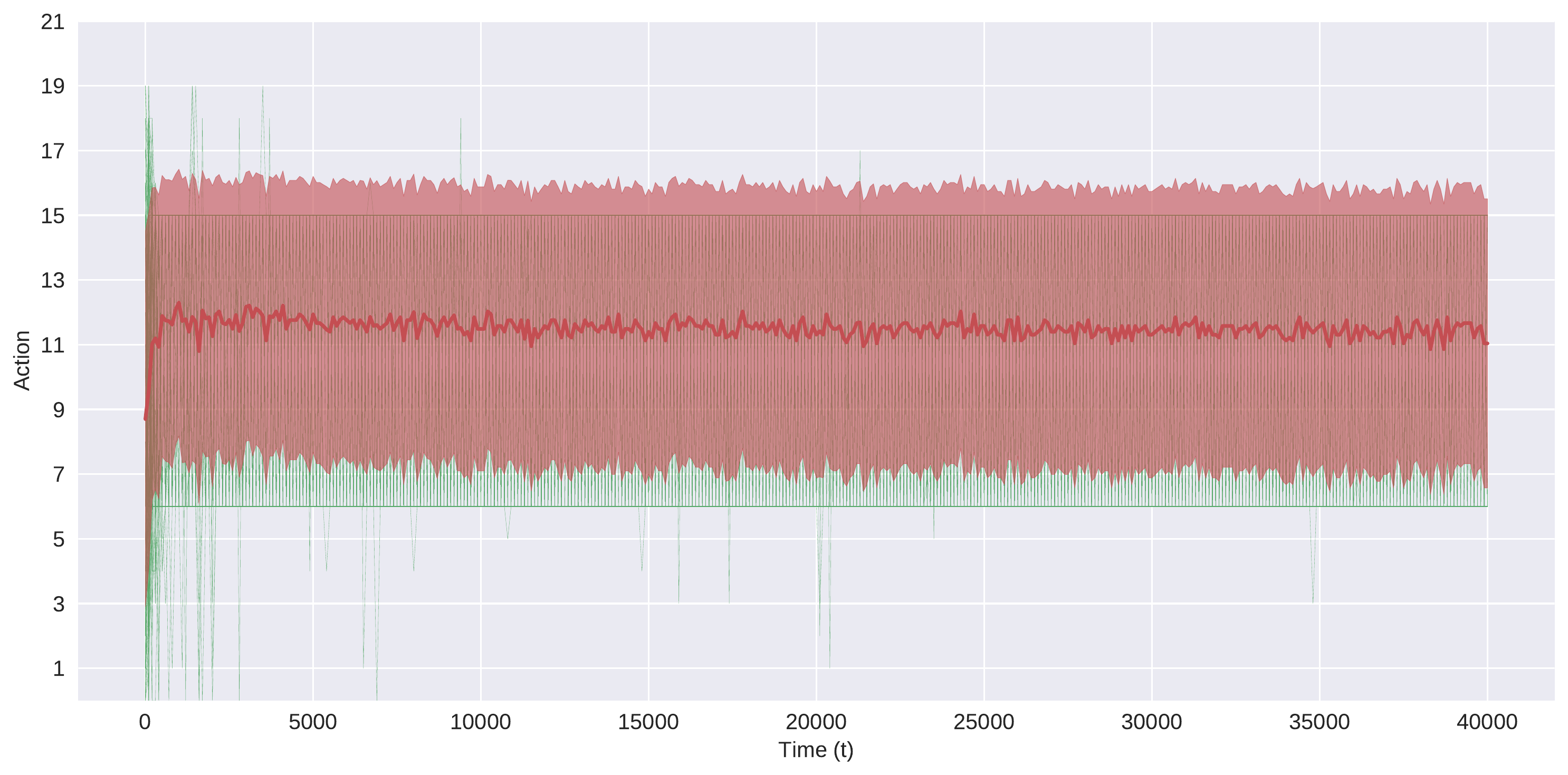}
		\subcaption{}
	\end{subfigure}
	\caption{\label{S2tsmm} {\bf Borda Scenario - Thompson Sampling (Maximin)} Off-color lines are individual runs, thick lines are mean over runs, shading is standard deviation. (a) algorithm specified regret, (b) $\I_t$ selections, and (c) $\J_t$ selections.  }
\end{figure}

\begin{figure}[H]
	\centering
	\begin{subfigure}[b]{0.75\textwidth}
		\centering
		\includegraphics[width=\textwidth]{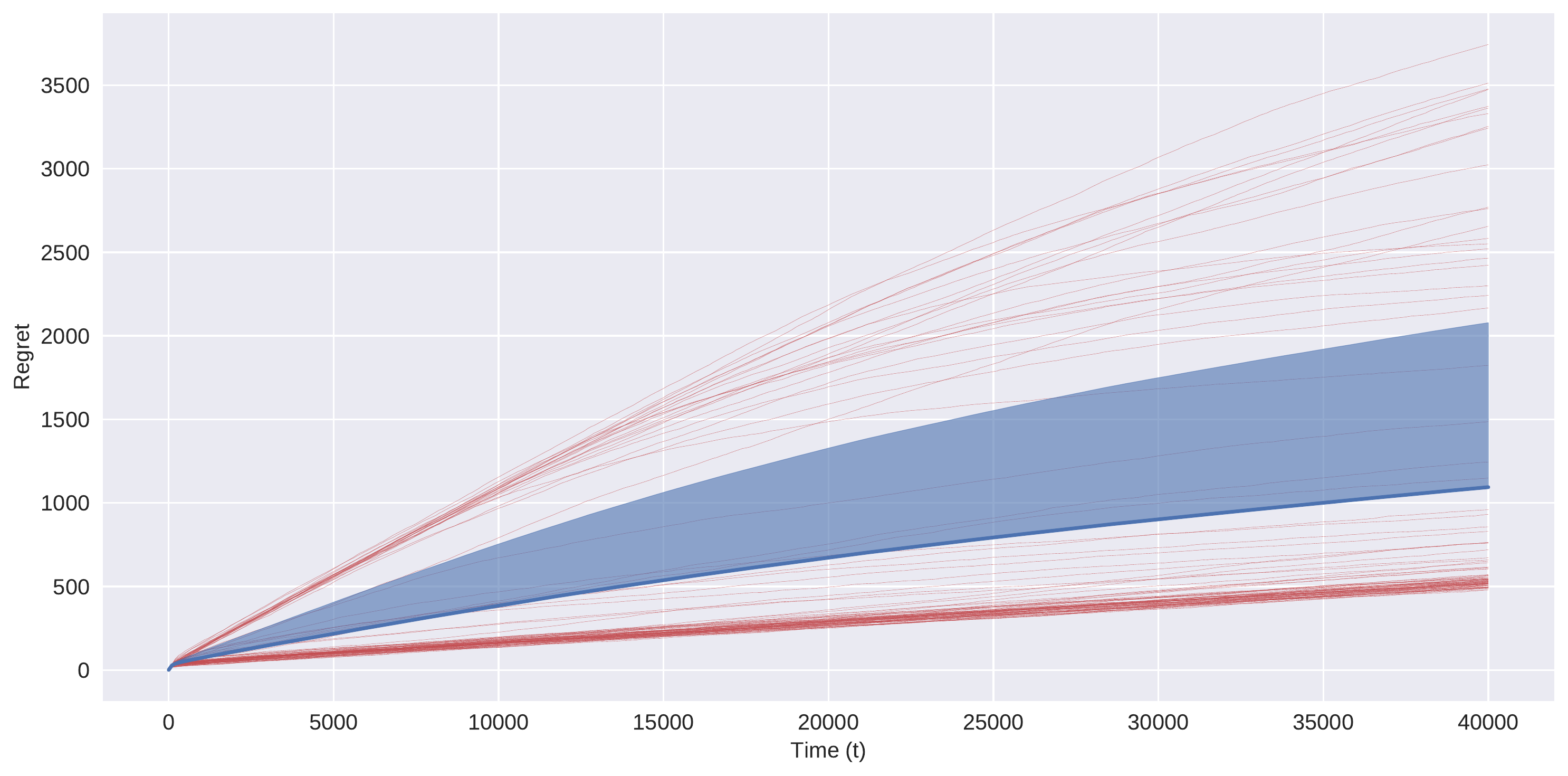}
		\subcaption{}
	\end{subfigure}\\
	\begin{subfigure}[b]{0.75\textwidth}
		\centering
		\includegraphics[width=\textwidth]{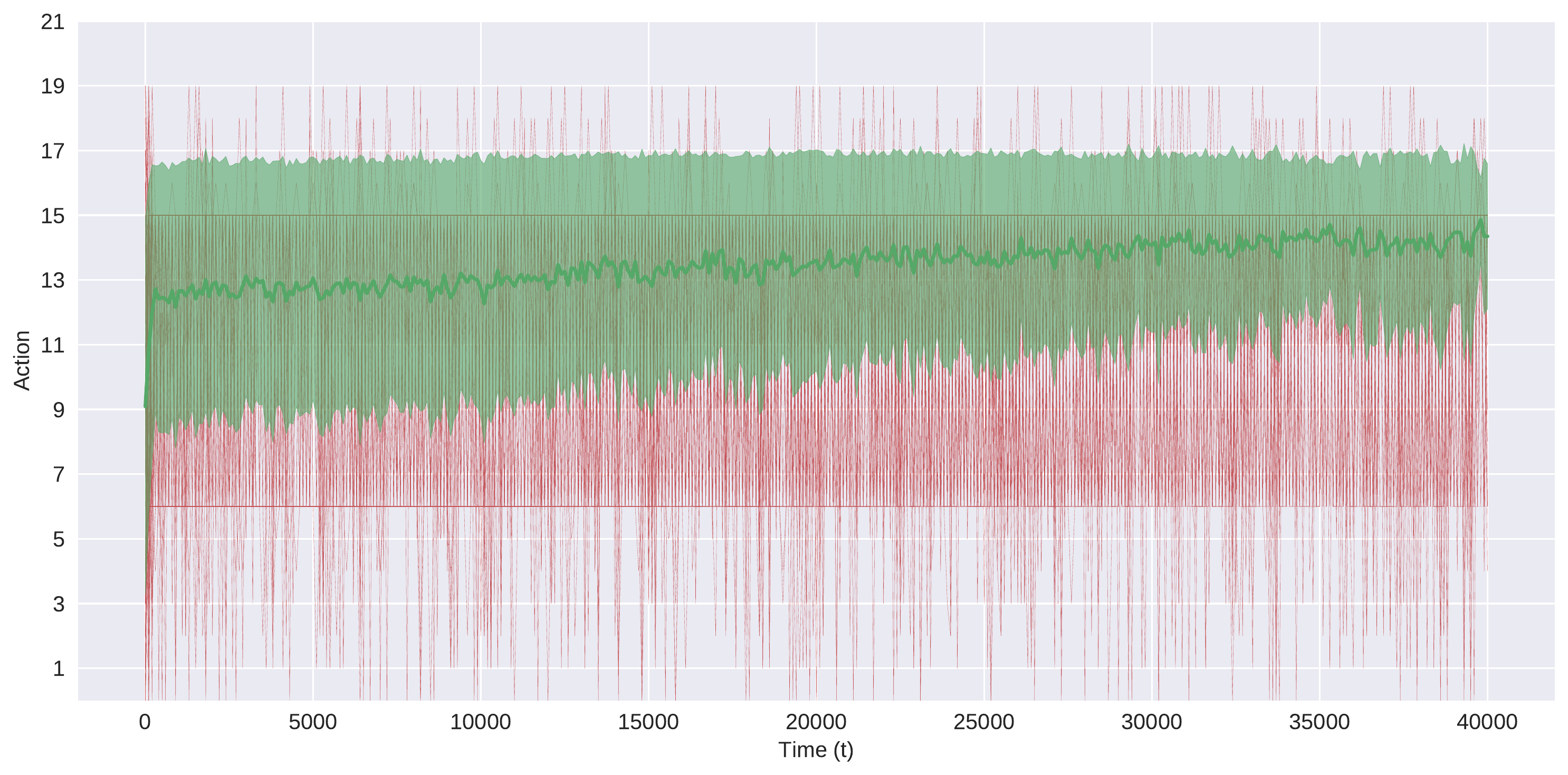}
		\subcaption{}
	\end{subfigure}\\
	\centering
	\begin{subfigure}[b]{0.75\textwidth}
		\centering
		\includegraphics[width=\textwidth]{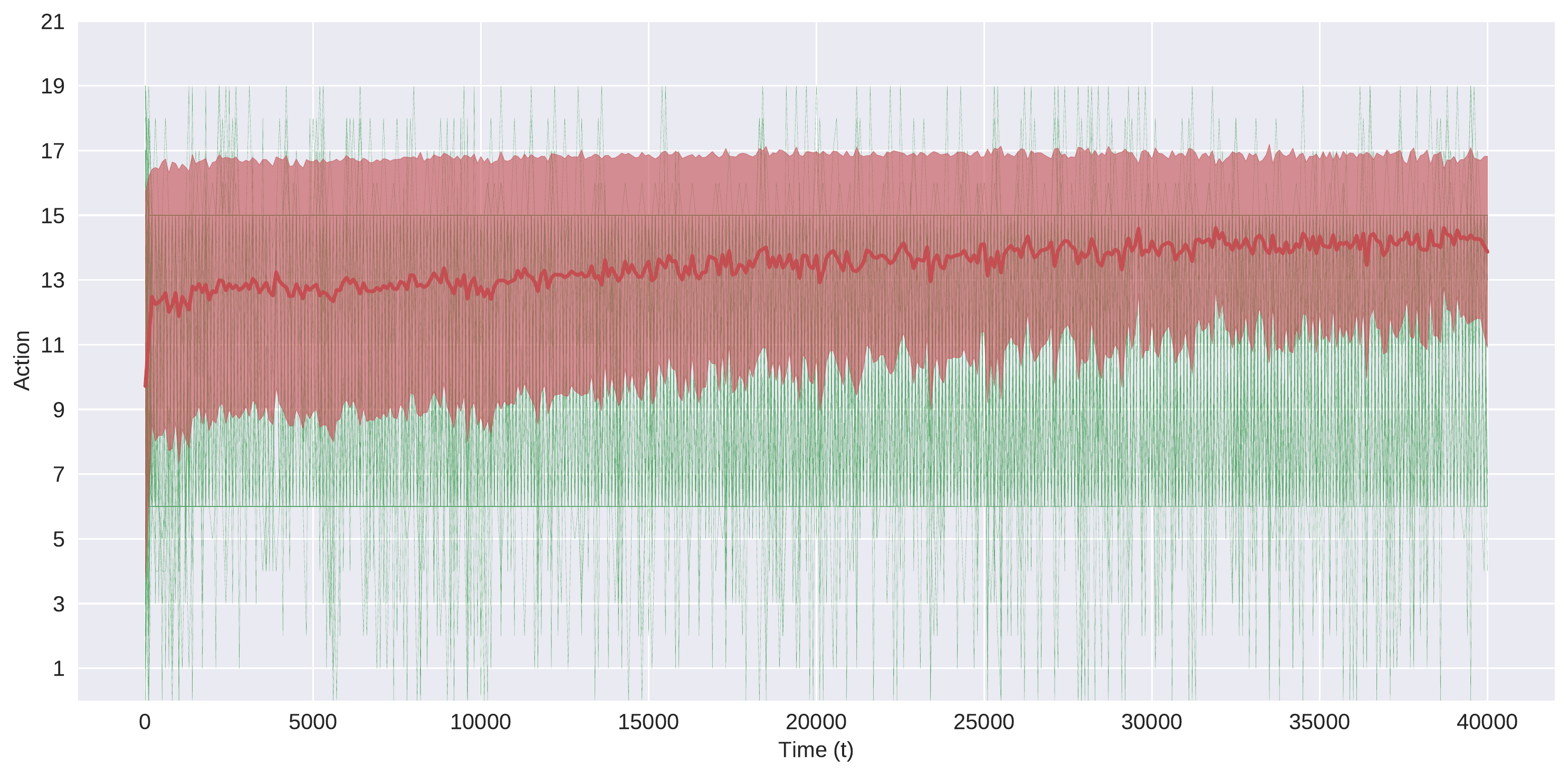}
		\subcaption{}
	\end{subfigure}
	\caption{\label{S2tsbr} {\bf Borda Scenario - Thompson Sampling (Borda)} Off-color lines are individual runs, thick lines are mean over runs, shading is standard deviation. (a) algorithm specified regret, (b) $\I_t$ selections, and (c) $\J_t$ selections.  }
\end{figure}

\begin{figure}[H]
	\centering
	\begin{subfigure}[b]{0.5\textwidth}
		\centering
		\includegraphics[width=\textwidth]{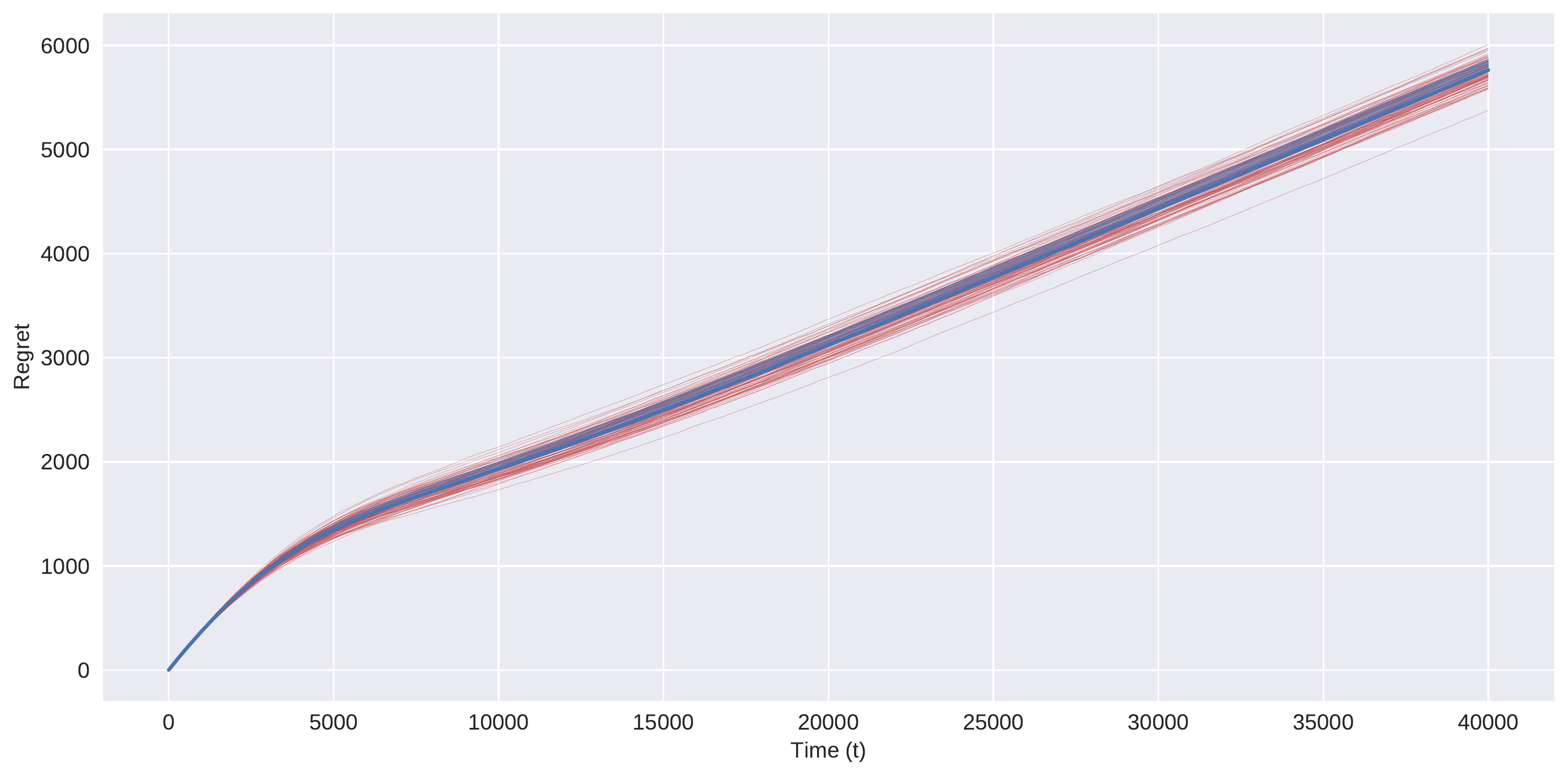}
		\subcaption{}
	\end{subfigure}\\
	\begin{subfigure}[b]{0.5\textwidth}
		\centering
		\includegraphics[width=\textwidth]{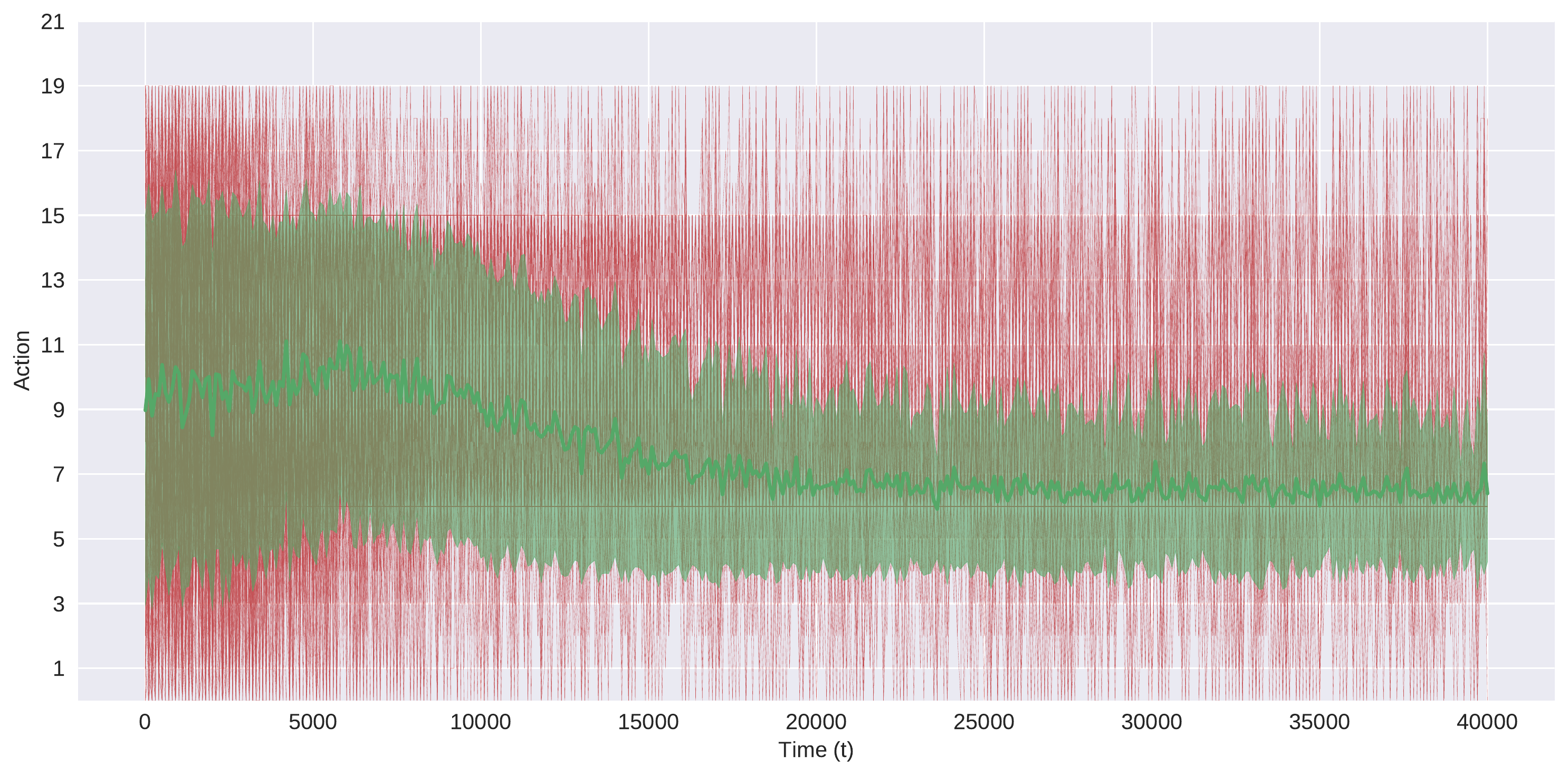}
		\subcaption{}
	\end{subfigure}\\
	\centering
	\begin{subfigure}[b]{0.5\textwidth}
		\centering
		\includegraphics[width=\textwidth]{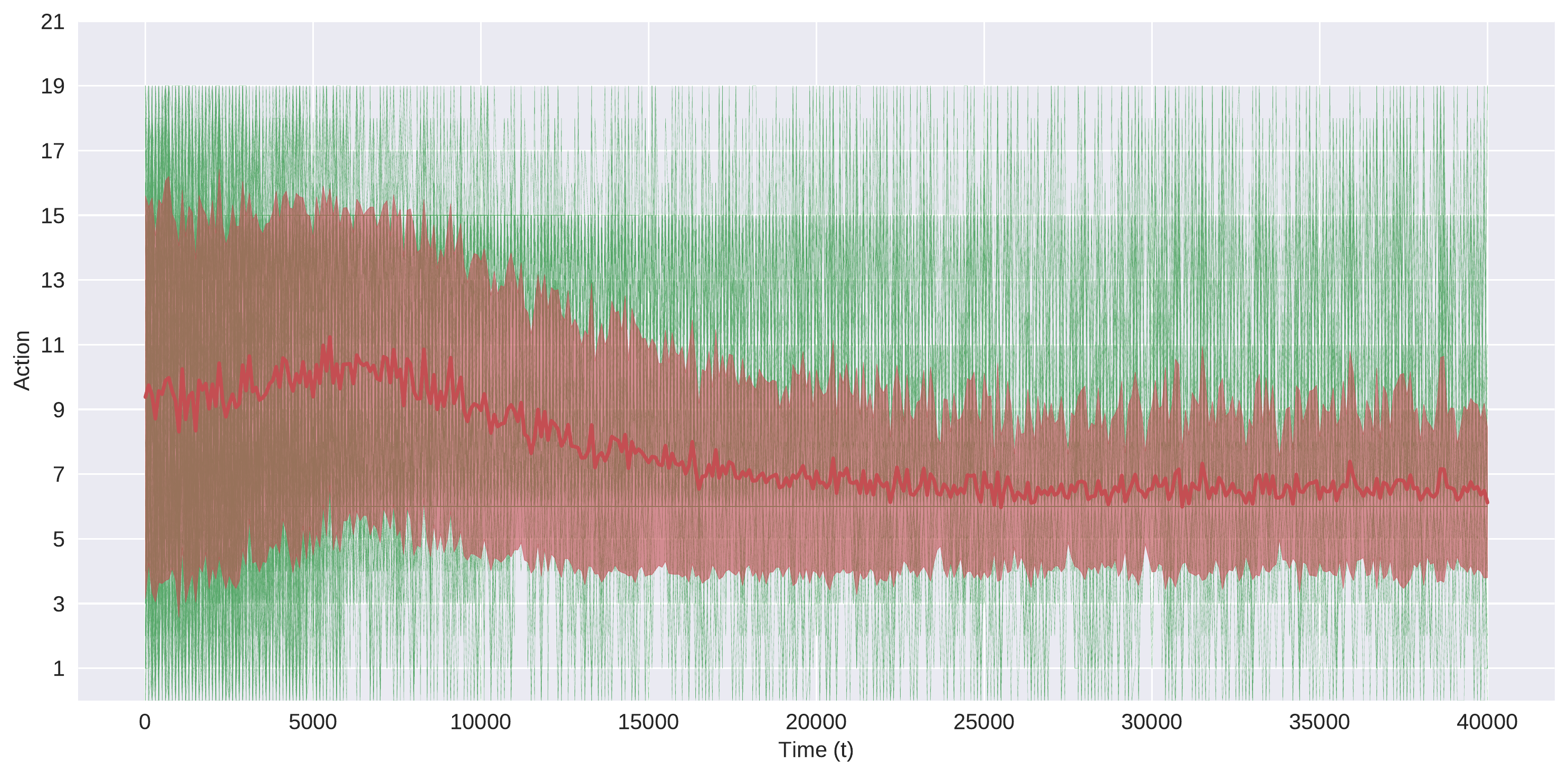}
		\subcaption{}
	\end{subfigure}\\
	\begin{subfigure}[b]{0.5\textwidth}
		\centering
		\includegraphics[width=\textwidth]{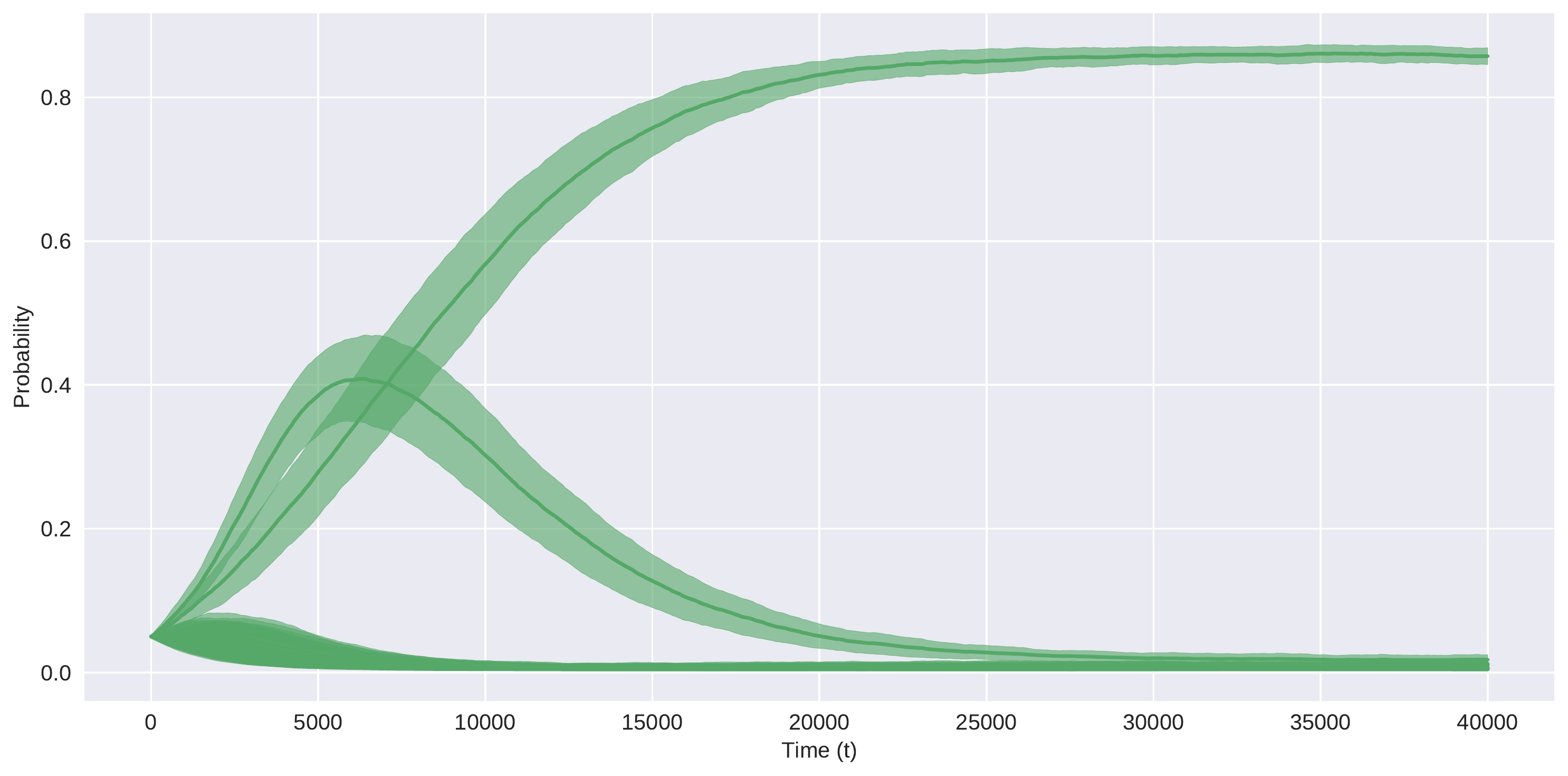}
		\subcaption{}
	\end{subfigure}\\
	\begin{subfigure}[b]{0.5\textwidth}
		\centering
		\includegraphics[width=\textwidth]{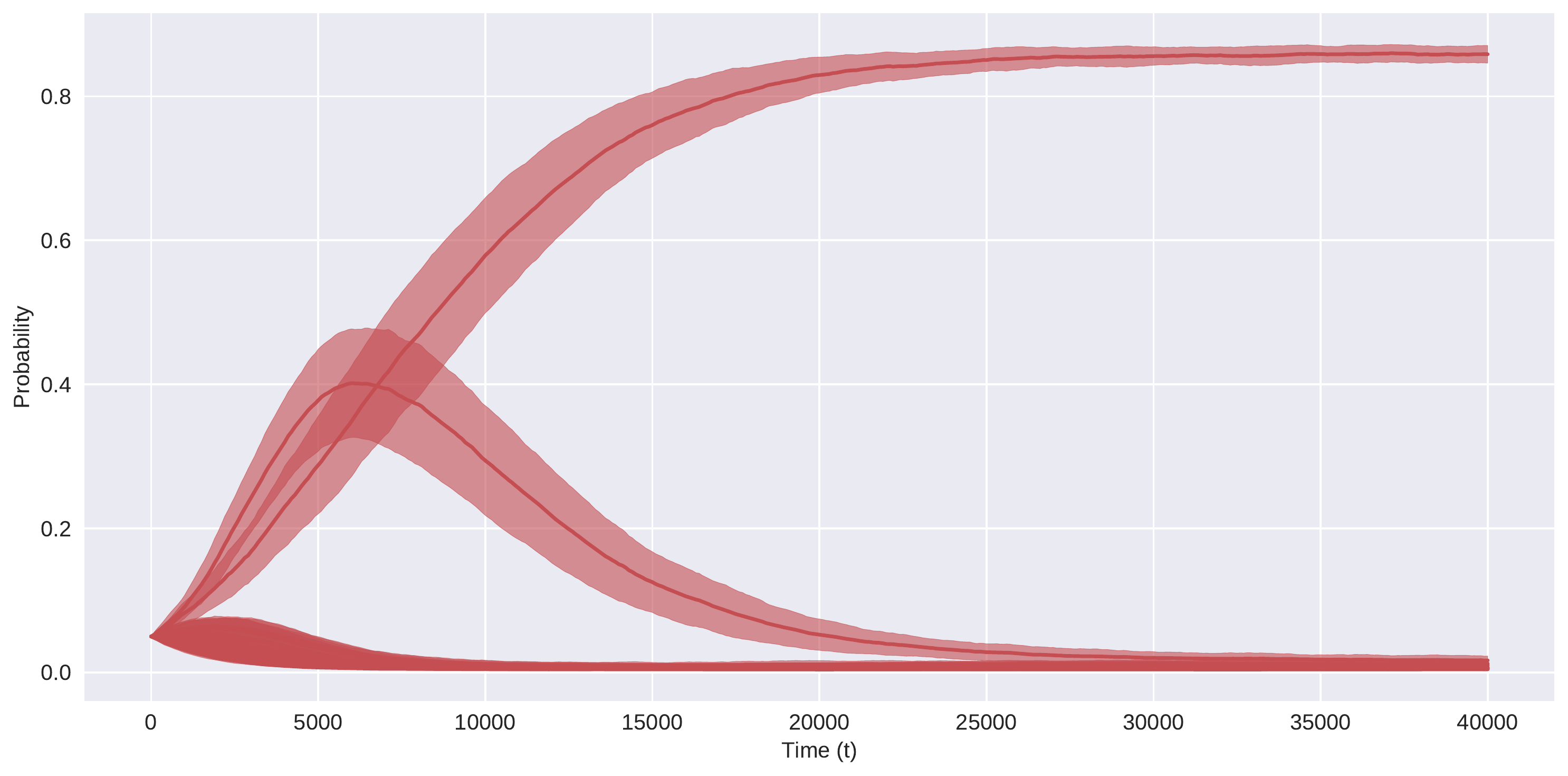}
		\subcaption{}
	\end{subfigure}
	\caption{\label{S2exp3p} {\bf Borda Scenario - SparringExp3.P} Off-color lines are individual runs, thick lines are mean over runs, shading is standard deviation. (a) algorithm specified regret, (b) $\I_t$ selections, (c) $\J_t$ selections, (d) $\p_t$ strategy, and (e) $\q_t$ strategy.  }
\end{figure}

\begin{figure}[H]
	\centering
	\begin{subfigure}[b]{0.65\textwidth}
		\centering
		\includegraphics[width=\textwidth]{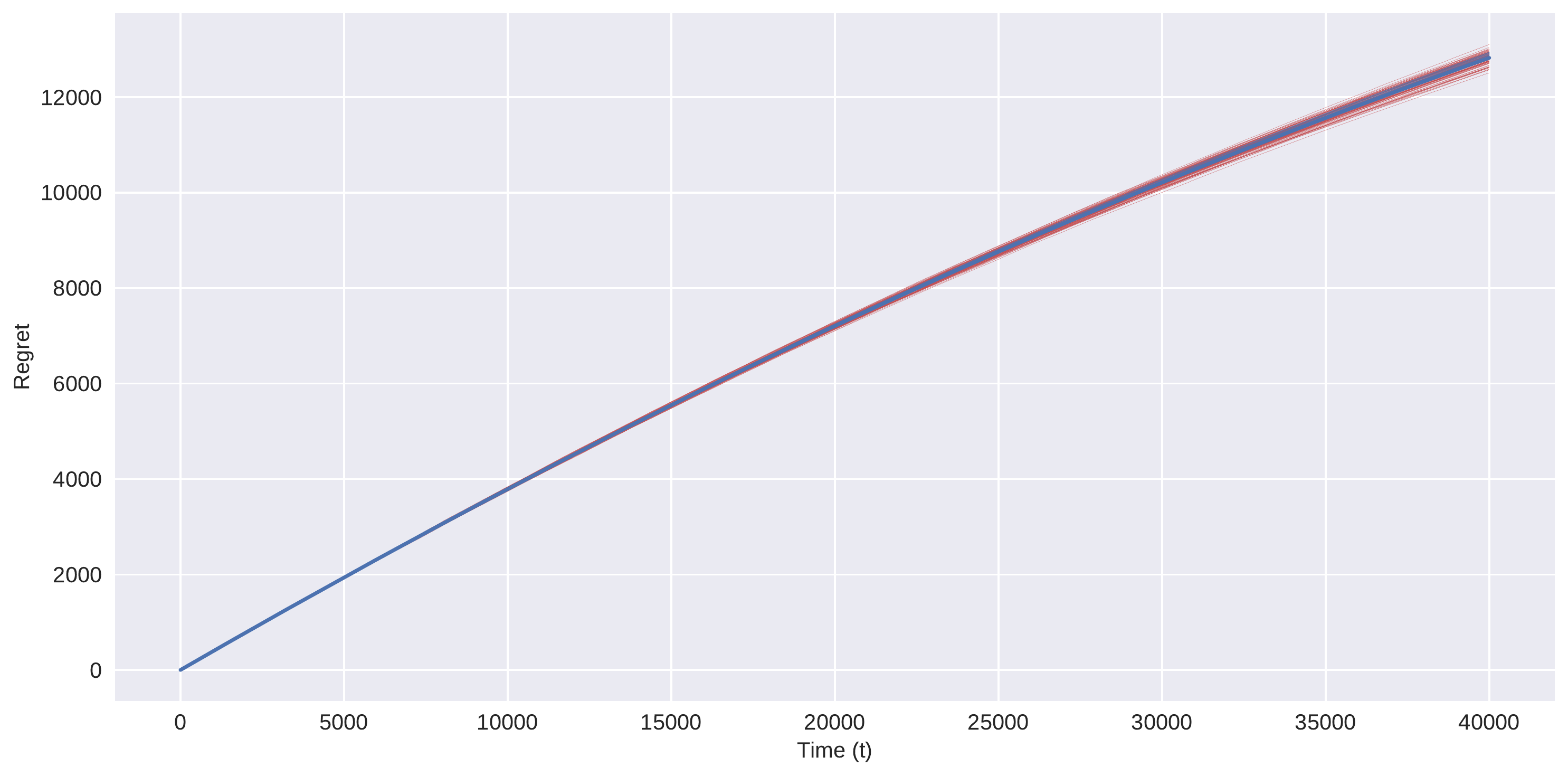}
		\subcaption{}
	\end{subfigure}\\
	\begin{subfigure}[b]{0.65\textwidth}
		\centering
		\includegraphics[width=\textwidth]{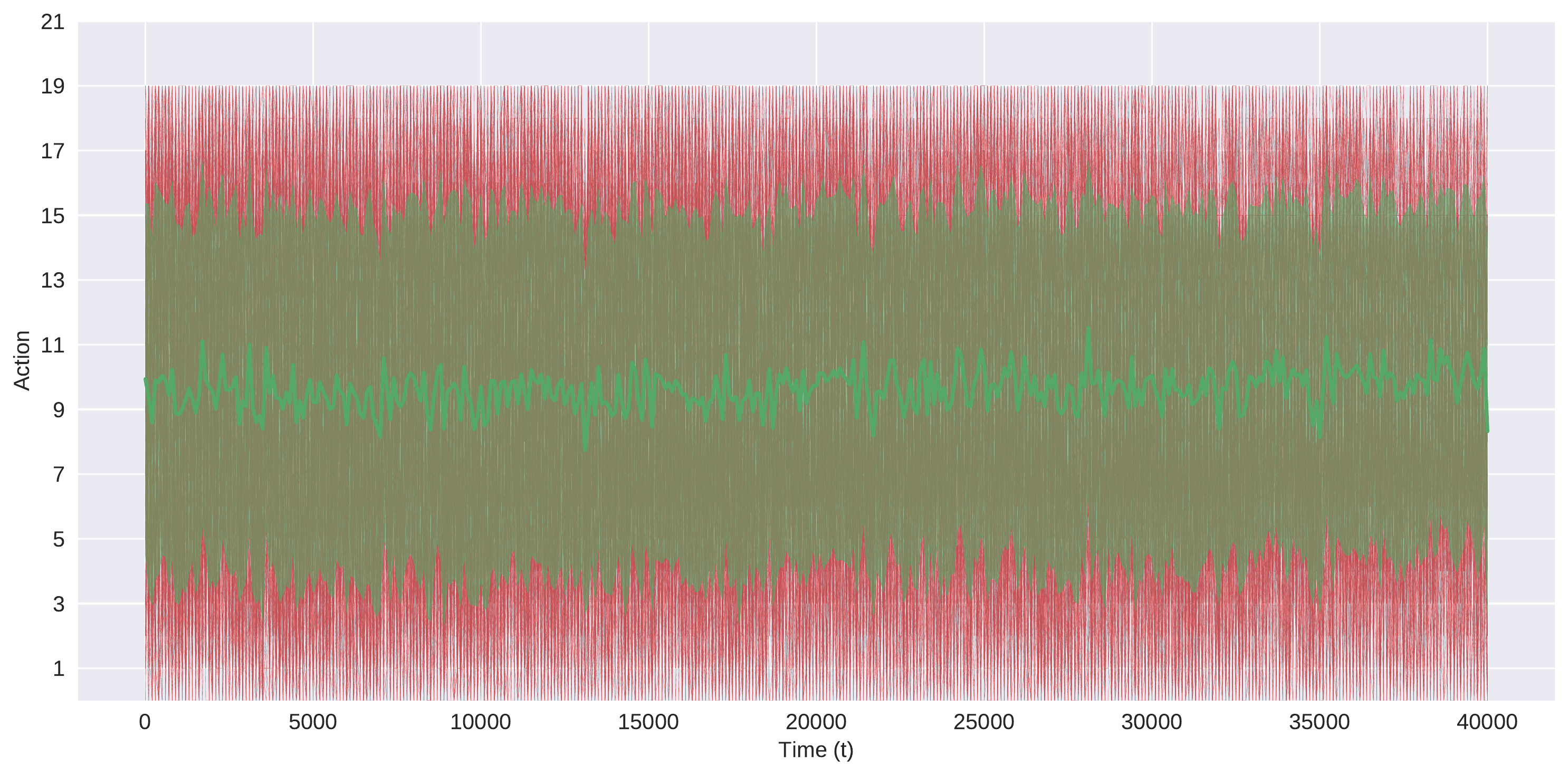}
		\subcaption{}
	\end{subfigure}\\
	\centering
	\begin{subfigure}[b]{0.65\textwidth}
		\centering
		\includegraphics[width=\textwidth]{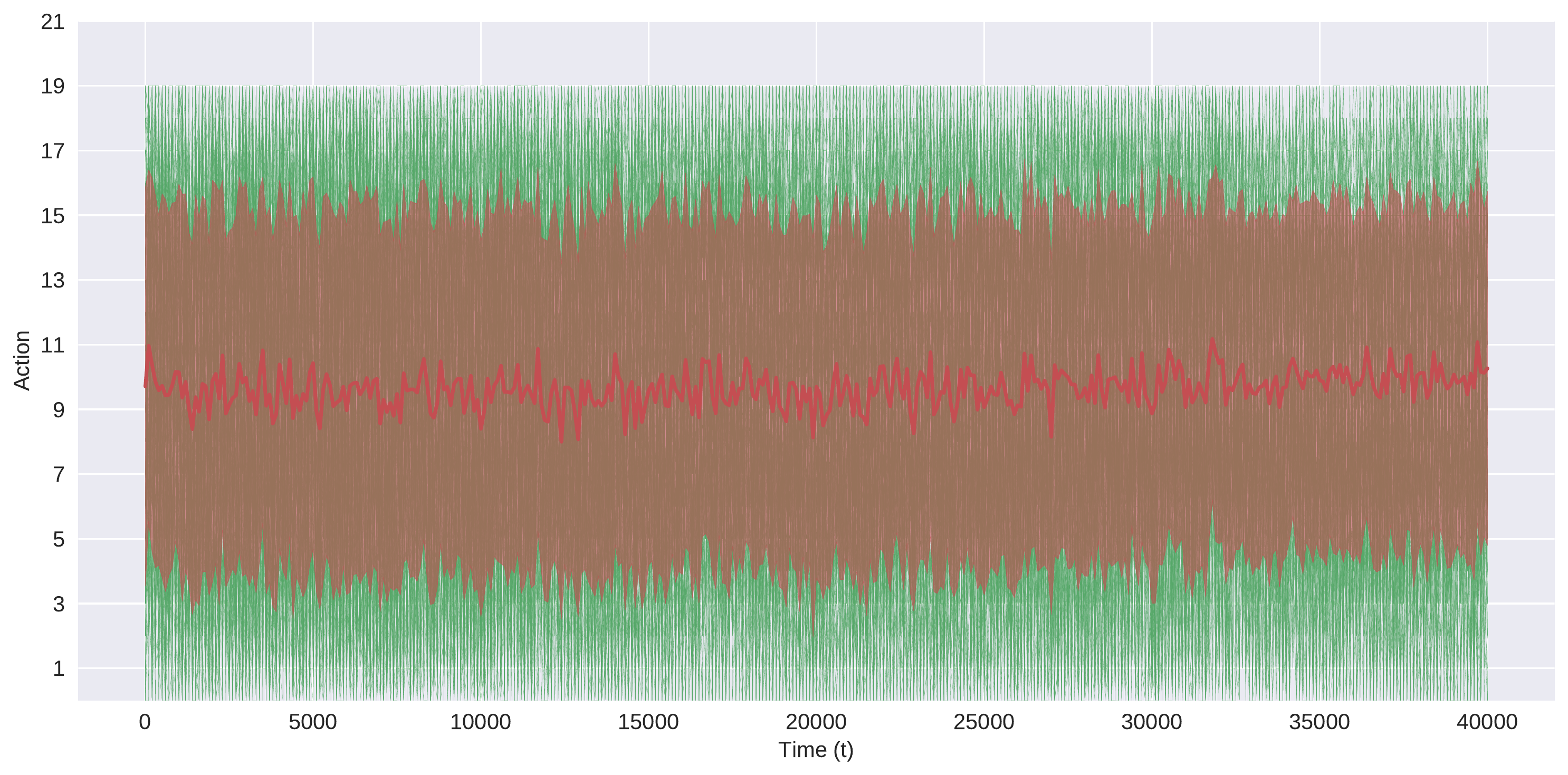}
		\subcaption{}
	\end{subfigure}\\
	\begin{subfigure}[b]{0.65\textwidth}
		\centering
		\includegraphics[width=\textwidth]{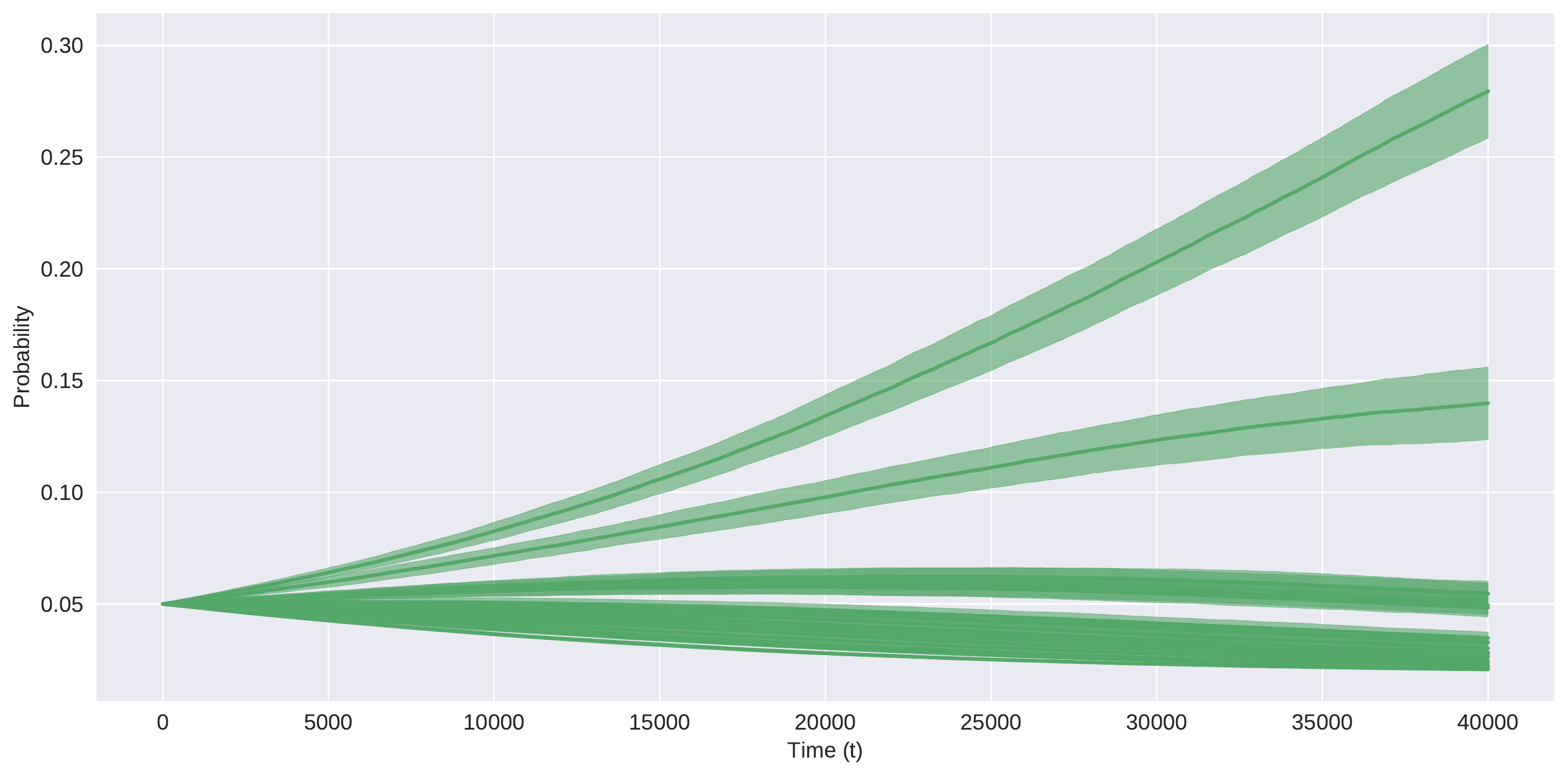}
		\subcaption{}
	\end{subfigure}
	\caption{\label{S2pm} {\bf Borda Scenario - Partial Monitoring} Off-color lines are individual runs, thick lines are mean over runs, shading is standard deviation. (a) algorithm specified regret, (b) $\I_t$ selections, (c) $\J_t$ selections, and (d) $\p_t$ strategy.  }
\end{figure}

\begin{figure}[H]
	\centering
	\begin{subfigure}[b]{0.75\textwidth}
		\centering
		\includegraphics[width=\textwidth]{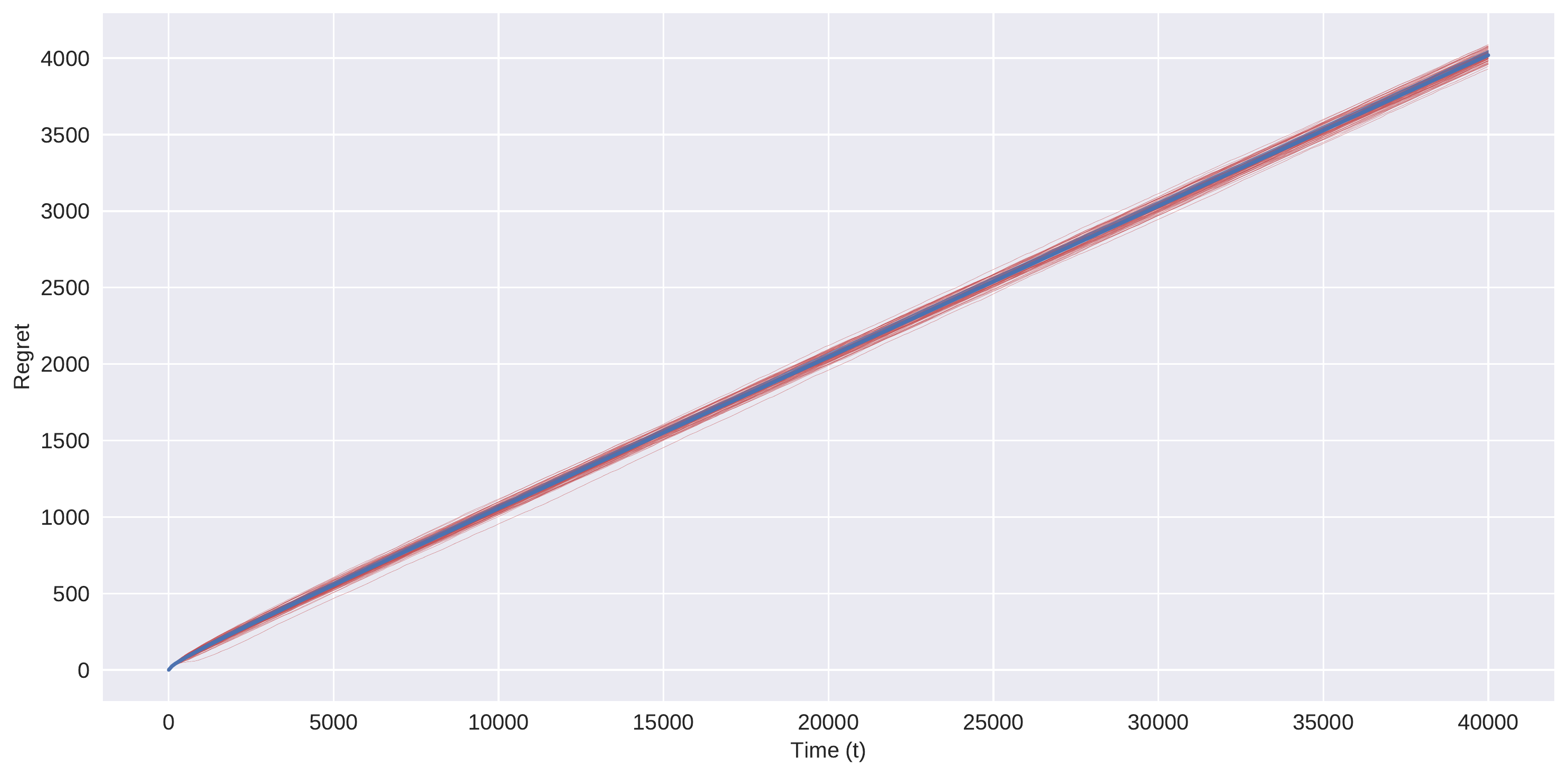}
		\subcaption{}
	\end{subfigure}\\
	\begin{subfigure}[b]{0.75\textwidth}
		\centering
		\includegraphics[width=\textwidth]{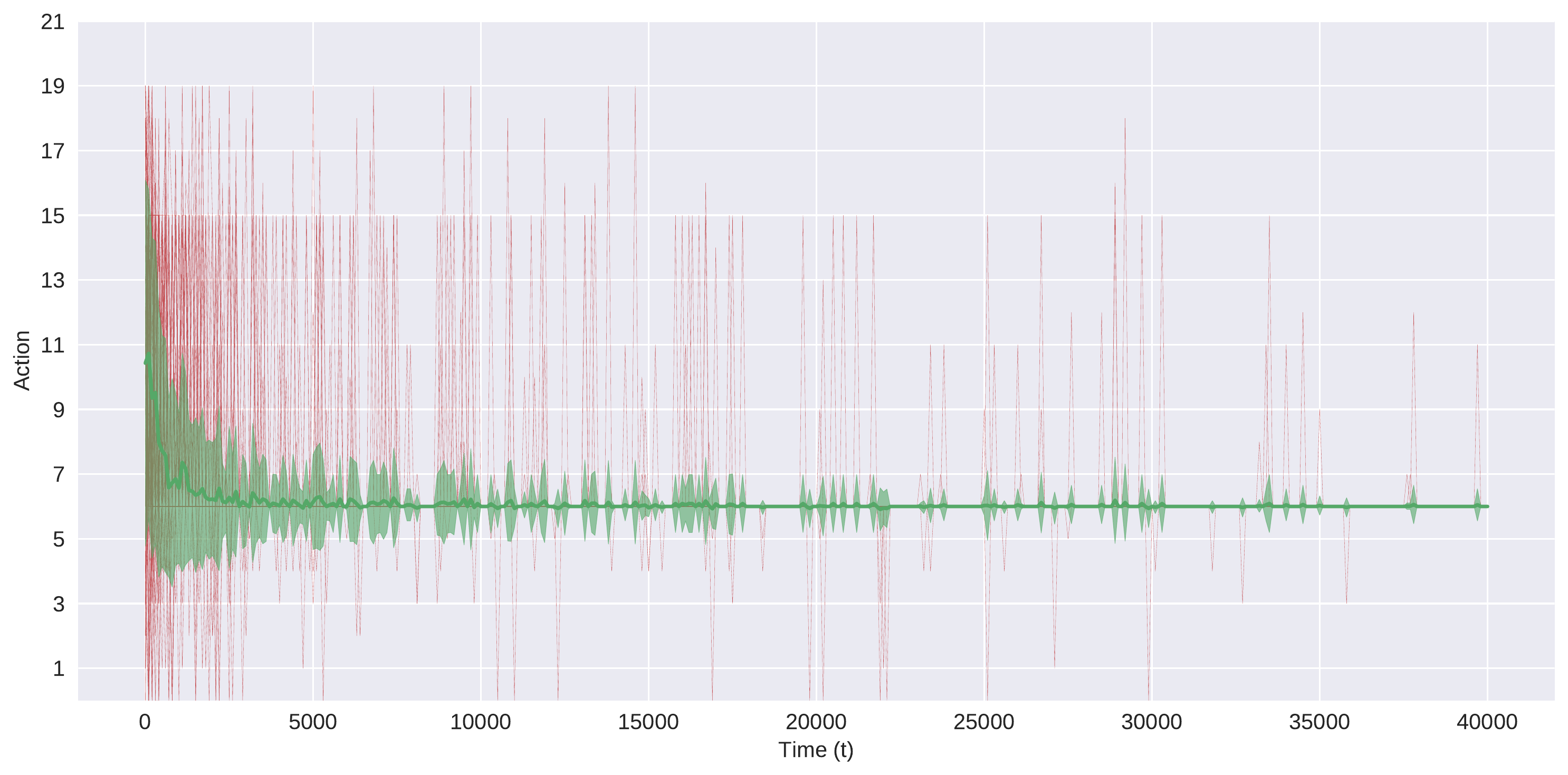}
		\subcaption{}
	\end{subfigure}\\
	\centering
	\begin{subfigure}[b]{0.75\textwidth}
		\centering
		\includegraphics[width=\textwidth]{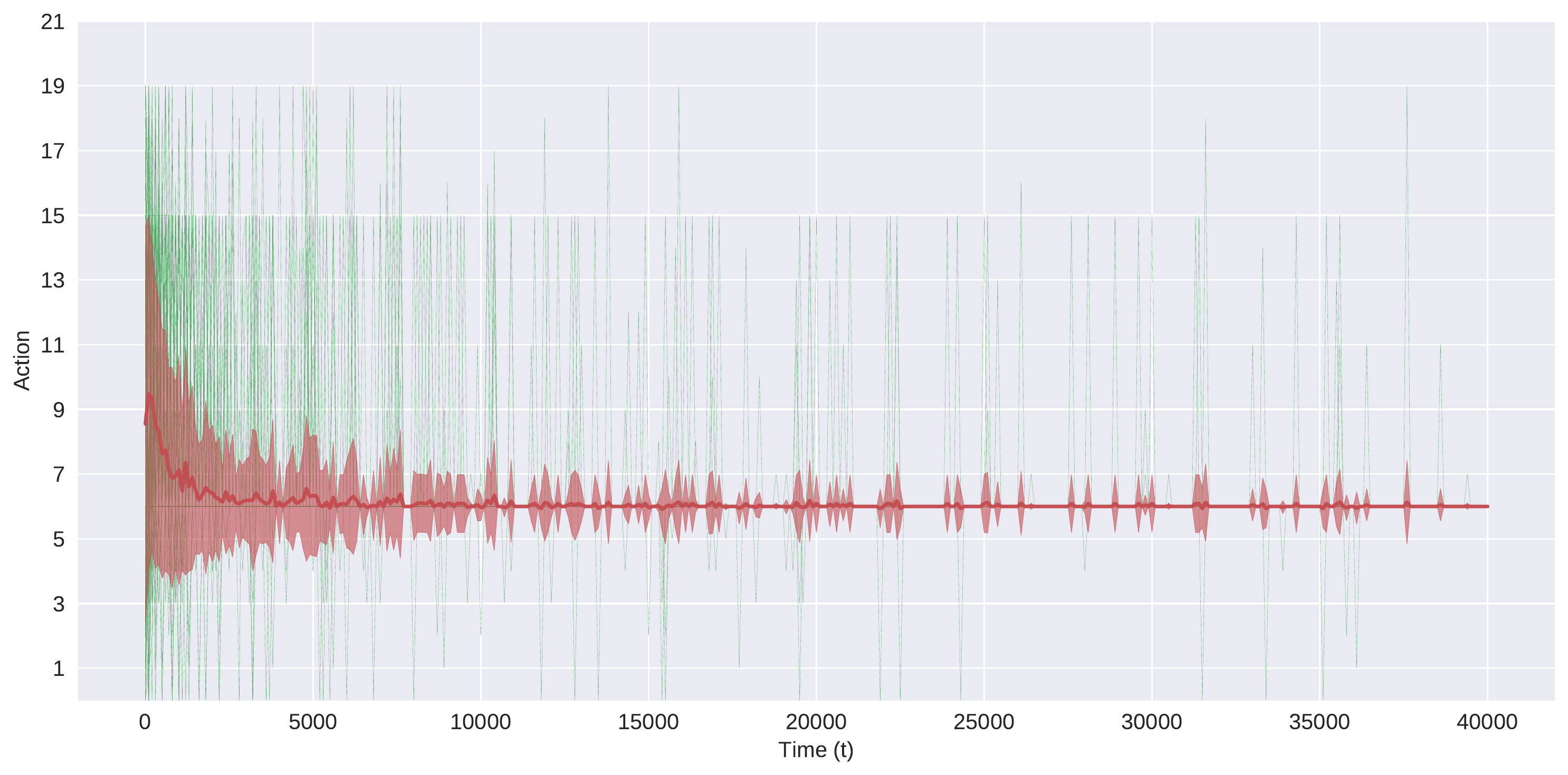}
		\subcaption{}
	\end{subfigure}
	\caption{\label{S2iss} {\bf Borda Scenario - ISS} Off-color lines are individual runs, thick lines are mean over runs, shading is standard deviation. (a) algorithm specified regret, (b) $\I_t$ selections, and (c) $\J_t$ selections.  }
\end{figure}

\begin{figure}[H]
	\centering
	\begin{subfigure}[b]{0.75\textwidth}
		\centering
		\includegraphics[width=\textwidth]{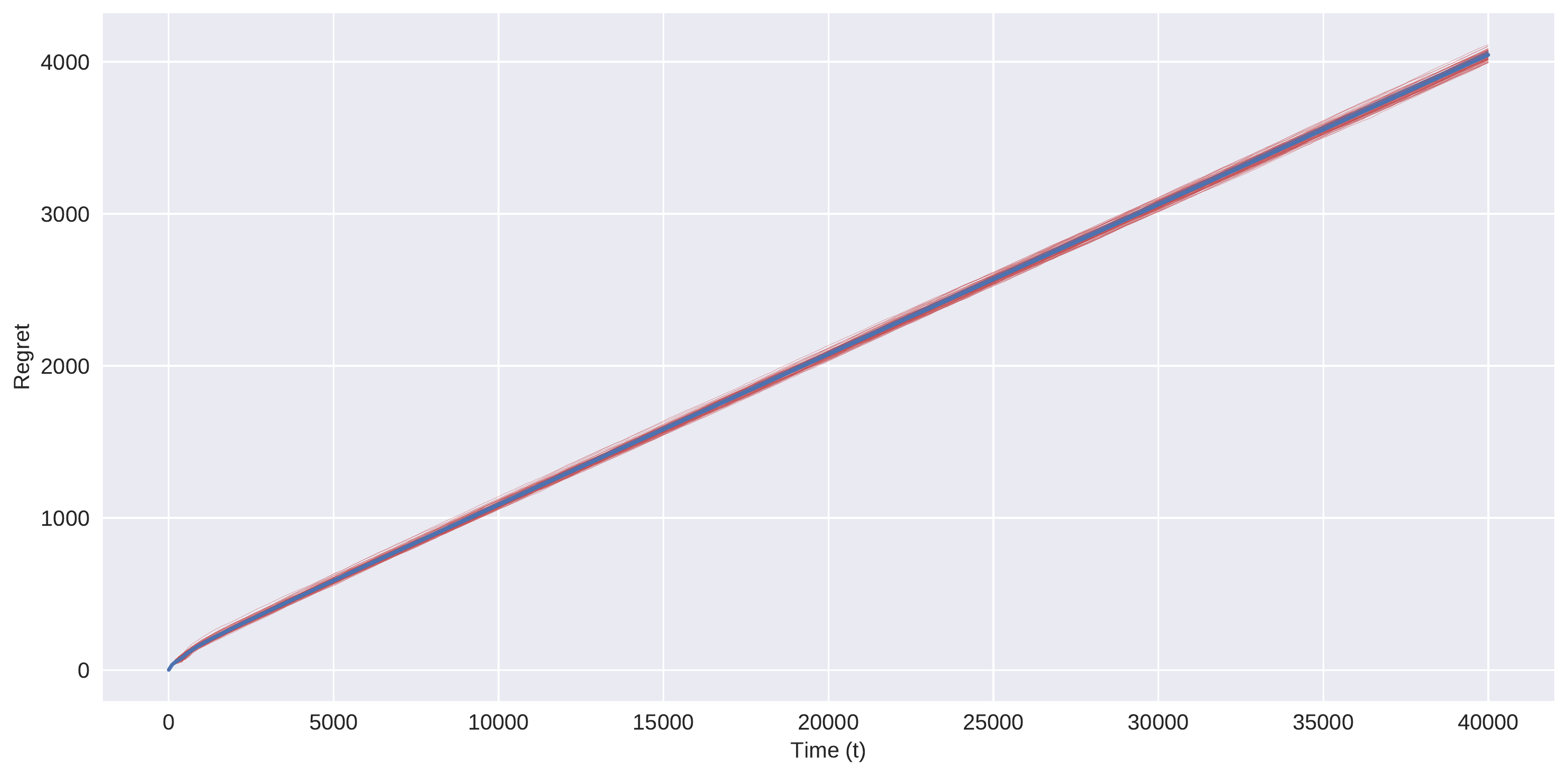}
		\subcaption{}
	\end{subfigure}\\
	\begin{subfigure}[b]{0.75\textwidth}
		\centering
		\includegraphics[width=\textwidth]{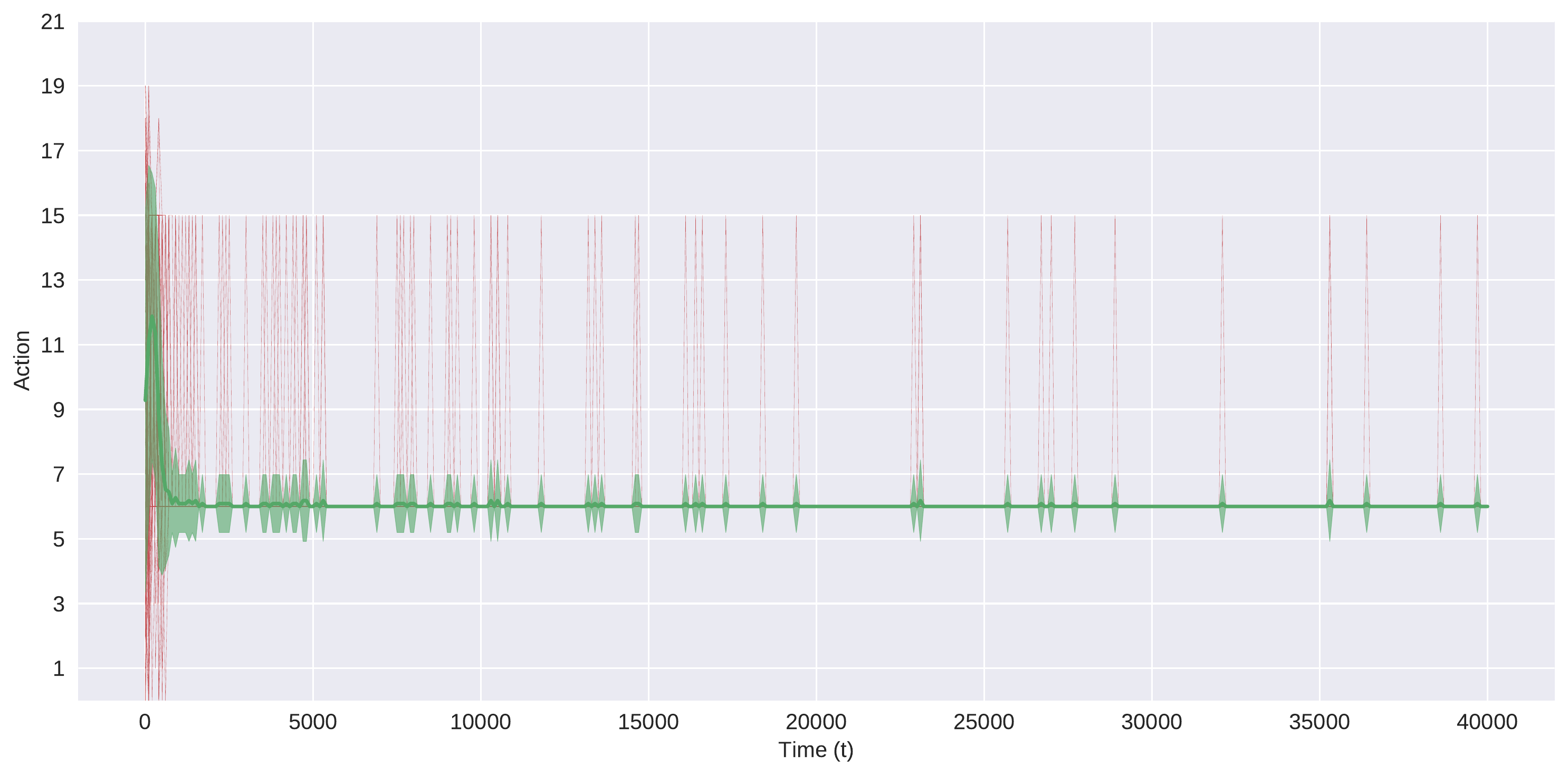}
		\subcaption{}
	\end{subfigure}\\
	\centering
	\begin{subfigure}[b]{0.75\textwidth}
		\centering
		\includegraphics[width=\textwidth]{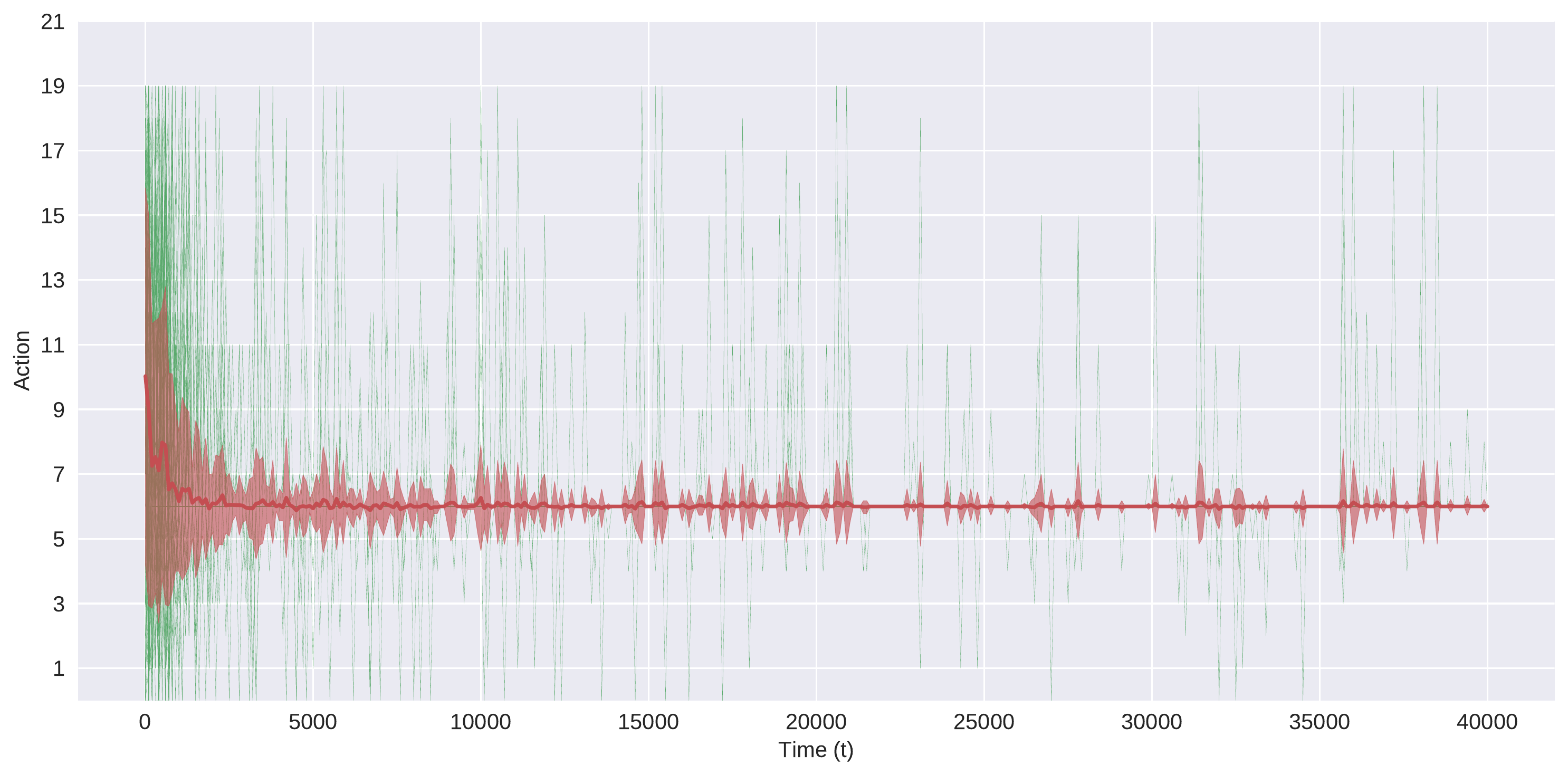}
		\subcaption{}
	\end{subfigure}
	\caption{\label{S2dts} {\bf Borda Scenario - DTS} Off-color lines are individual runs, thick lines are mean over runs, shading is standard deviation. (a) algorithm specified regret, (b) $\I_t$ selections, and (c) $\J_t$ selections.  }
\end{figure}

In this section, we provide additional plots that detail the behavior of the algorithms for the different experimental scenarios. For all figures:
\begin{itemize}
	\item (a) shows a detailed plot of the regret over the runs for the scenario, with off-color lines showing individual runs, thick line showing the mean over runs, and shaded area showing the standard deviation over runs (plotted above the mean)
	\item (b) shows the $\I_t$ action selections over the runs for the scenario, with off-color lines showing individual runs, thick line showing the mean over runs, and shaded area showing the standard deviation over runs (plotted above and below the mean)
	\item (c) shows the $\J_t$ action selections over the runs for the scenario, with off-color lines showing individual runs, thick line showing the mean over runs, and shaded area showing the standard deviation over runs (plotted above and below the mean)
	\item (d - if applicable) shows the $\p_t$ strategy over the runs for the scenario, with thick line showing the mean over runs, and shaded area showing the standard deviation over runs (plotted above and below the mean)
	\item (e - if applicable) shows the $\q_t$ strategy over the runs for the scenario, with thick line showing the mean over runs, and shaded area showing the standard deviation over runs (plotted above and below the mean)
\end{itemize}
$\hfill$\\\\
For the Condorcet scenario, see Figs. \ref{S1tsmm} - \ref{S1dts}). For the Borda scenario, see Figs. \ref{S2tsmm} - \ref{S2dts}.

\newpage

\newpage

\newpage

\newpage

\newpage

\newpage

\newpage

\newpage

\newpage

\newpage

\newpage

\newpage

\end{document}